\documentclass[letterpaper]{article} % DO NOT CHANGE THIS
\usepackage{aaai23}  % DO NOT CHANGE THIS
\usepackage{times}  % DO NOT CHANGE THIS
\usepackage{helvet}  % DO NOT CHANGE THIS
\usepackage{courier}  % DO NOT CHANGE THIS
\usepackage[hyphens]{url}  % DO NOT CHANGE THIS
\usepackage{graphicx} % DO NOT CHANGE THIS
\usepackage{natbib}  % DO NOT CHANGE THIS AND DO NOT ADD ANY OPTIONS TO IT
\usepackage{caption} % DO NOT CHANGE THIS AND DO NOT ADD ANY OPTIONS TO IT
\usepackage{mathtools}
\usepackage{mathrsfs}
\usepackage{amsfonts}
\usepackage{subcaption}
\usepackage{algorithm}
\usepackage{algorithmic}

\usepackage{url}
\usepackage{xcolor}

\usepackage{amsthm}

\newtheorem*{definition*}{Definition}
\usepackage{newfloat}
\usepackage{listings}
\DeclareCaptionStyle{ruled}{labelfont=normalfont,labelsep=colon,strut=off} % DO NOT CHANGE THIS
\floatstyle{ruled}
\newfloat{listing}{tb}{lst}{}
\floatname{listing}{Listing}
\title{Causal Recurrent Variational Autoencoder for Medical Time Series Generation}
\author{
    Hongming Li\textsuperscript{\rm 1},
    Shujian Yu\textsuperscript{\rm 2}\thanks{Corresponding Author.}, 
    Jose Principe\textsuperscript{\rm 1}\\
}
\affiliations{
    \textsuperscript{\rm 1} University of Florida \\
    \textsuperscript{\rm 2} UiT - The Arctic University of Norway \\
    hongmingli@ufl.edu,
    yusj9011@gmail.com,
    principe@cnel.ufl.edu
}

\usepackage{bibentry}
\begin{document}

\maketitle

\begin{abstract}
We propose \emph{causal recurrent variational autoencoder} (CR-VAE), a novel generative model that is able to learn a Granger causal graph from a multivariate time series $\mathbf{x}$ and incorporates the underlying causal mechanism into its data generation process. 
Distinct to the classical recurrent VAEs, our CR-VAE uses a multi-head decoder, in which the $p$-th head is responsible for generating the $p$-th dimension of $\mathbf{x}$ (i.e., $\mathbf{x}^p$). By imposing a sparsity-inducing penalty on the weights (of the decoder) and encouraging specific sets of weights to be zero, our CR-VAE learns a sparse adjacency matrix that encodes causal relations between all pairs of variables. Thanks to this causal matrix, our decoder strictly obeys the underlying principles of Granger causality, thereby making the data generating process transparent. We develop a two-stage approach to train the overall objective.
Empirically, we evaluate the behavior of our model in synthetic data and two real-world human brain datasets involving, respectively, the electroencephalography (EEG) signals and the functional magnetic resonance imaging (fMRI) data. Our model consistently outperforms state-of-the-art time series generative models both qualitatively and quantitatively. Moreover, it also discovers a faithful causal graph with similar or improved accuracy over existing Granger causality-based causal inference methods. Code of CR-VAE is publicly available at \url{https://github.com/hongmingli1995/CR-VAE}.
\end{abstract}

\section{Introduction}

Multivariate time series data are ubiquitous in numerous real-world applications. e.g., the electroencephalogram (EEG) signals~\cite{isaksson1981computer}, the climate records~\cite{runge2019detecting}, and the stellar light curves in astronomy~\cite{huijse2012information}. Traditional machine learning tasks on time series data include anomaly detection, segmentation, forecasting, classification, etc. Among them, the time series forecasting or prediction, which uses past or historical observations to predict future values, is perhaps the most popular one. 

In recent years, the design of generative models for time series data emerged as a challenge. One reason is that most of existing machine learning models, especially deep neural networks, are data-hungry, which means that a sufficient number of (labeled) samples are required during training before their practical deployment. Unfortunately, in some sensitive applications, especially those involving medical and healthcare domains, collecting and exchanging real data from patients requires a long administrative process or is even prohibited. This in turn may inhibit research progress on model comparison and reproducibility.  

% ~\cite{esteban2017real}

%collecting real data is usually expensive or even prohibitive. 

% and other types of physiological ; the stock prices in financial engineering~\cite{ozbayoglu2020deep}, 

%There are several problems around time series (e.g., anomaly detection, segmentation, forecasting, and classification), 

%which in turn motivate different ways to analyze these data and to study their characteristics (e.g., time domain analysis with respect to frequency domain analysis; linear models with respect to nonlinear models). 

%Usually, explicitly modeling the joint distribution $p(\mathbf{x}_{1:T})$ is hard. Autoregressive models factor the joint distribution into a product of conditional density or dynamics distribution $p(\mathbf{x}_{1:T})$. 

A good generative model for time series is expected to model both the joint distribution $p(\mathbf{x}_{1:T})$ and the transition dynamics $p(\mathbf{x}_t |\mathbf{x}_{1:t-1})$ for any $t$. 
Although most of popular predictive models, such as autoregressive integrated moving average (ARIMA), kernel adaptive filters (KAF)~\cite{liu2008kernel} and deep state-space models (SSMs)~\cite{rangapuram2018deep}, provide different ways to capture $p(\mathbf{x}_t |\mathbf{x}_{1:t-1})$ or $p(\mathbf{x}_t |\mathbf{x}_{t-\tau:t-1})$ in the window of length $\tau$, they are \emph{deterministic} mappers, rather than \emph{generative}. In other words, these models are incapable of inferring unobserved latent factors (such as trend and seasonality) from observational data, and generating new time series values by sampling from a tractable latent distribution.

%and how to sample from a tractable latent distribution to generate realistic new time series.   

%Popular time series predictive models, such as autoregressive integrated moving average (ARIMA), (kernel) least mean square (KLMS) filter~\cite{liu2008kernel} and recurrent neural networks~\cite{hewamalage2021recurrent}, are able to capture $p(\mathbf{x}_t |\mathbf{x}_{1:t-1})$ or $p(\mathbf{x}_t |\mathbf{x}_{t-\tau-1:t-1})$ ($\tau$ is the window length). However, they are fundamentally deterministic, rather than generative which requires modeling $p(x_{1:t} |z)$, in which $z$ denotes unobserved latent factors (such as trend or seasonality).

%Substantial efforts have been made on time series predictive models in the last decades. Notable examples are the autoregressive integrated moving average (ARIMA) in statistics and econometrics, the (kernel) least mean square (KLMS) filter~\cite{liu2008kernel} in signal processing, and the recurrent neural networks~\cite{hewamalage2021recurrent} in machine learning. In general, the predictive model is deterministic in nature, as it precisely models the conditional density or dynamics $p(x_t |x_{1:t-1})$, but cannot be used to generate new time series which requires modeling $p(x_{1:t} |z)$, in which $z$ denotes unobserved latent factors (such as trend or seasonality).

On the other hand, causal inference from time series data has also attracted increasing attention. Taking the functional magnetic resonance imaging (fMRI) data as an example, it is of paramount importance to identify causal influences between brain activated regions~\cite{deshpande2009multivariate}. This causal graph may also provide insights into brain network-based psychiatric disorder diagnosis~\cite{wang2020large}.

Given the urgent need for a reliable time series generative model and the modern trend of causal inference, one question arises naturally: can we develop a new generative model for time series such that it can also be used for causal discovery? 
In this paper, we give an explicit answer to this question. To this end, we develop \emph{causal recurrent variational autoencoder} (CR-VAE), which, to the best of our knowledge, is the first endeavor to integrate the concept of Granger causality within a recurrent VAE framework. Specifically, given a $M$-variate time series $\mathbf{x}=(\mathbf{x}^1,\mathbf{x}^2,\cdots,\mathbf{x}^M)$, our CR-VAE consists of an encoder and a multi-head decoder, in which the $p$-th head is responsible for generating the $p$-th dimension of $\mathbf{x}$ (i.e., $\mathbf{x}^p$). We impose a sparsity penalty on the weight matrix that connects input and hidden state (in the decoder), thereby encouraging the model to learn a sparse matrix $A \in \mathbb{R}^{M \times M}$ to encode the Granger causality between pairwise dimensions of $\mathbf{x}$. Such design also makes the generation process compatible with the underlying principles of Granger causality (i.e., causes appear prior to effects). Additionally, we also propose an error-compensation module to take into account the instantaneous influence $\varepsilon_t$ excluding the past of one process.

% We assume it to be additive, i.e., $\mathbf{x}_t = f(x_{1:t-1}) + \varepsilon_t$, and implement a VAE to estimate $\varepsilon_t$.

% such that it can naturally models both the time series transition dynamics $p(x_t |x_{1:t-1})$ and the generative mapping $p(x_{1:t} |z)$

We conduct extensive experiments on synthetic sequences and real-world medical time series. In terms of time series generation, we evaluate the closeness between real data distribution and synthetic data distribution both qualitatively and quantitatively. In terms of causal discovery, we compare our discovered causal graph with state-of-the-art (SOTA) approaches that also aim to identify the Granger causality. Our model achieves competitive performance in both tasks.

\section{Background Knowledge}
The proposed work lies at the intersection of multiple strands of research, combining themes from autoregressive models for temporal dynamics, Granger causality for causal discovery, and VAE-based time series models.
\subsection{Time Series Generative Models}
A deep generative model $g_\theta$ is trained to map samples from a simple and tractable distribution $p(\mathbf{z})$ to a more complicated distribution $p(g_\theta (\mathbf{z}))$, which is similar to the true distribution $p(\mathbf{x})$. For time series data, 
one can simply generate synthetic time series under a Generative Adversarial Network (GAN)~\cite{goodfellow2014generative} framework, by making use recurrent neural networks in both the generator and the discriminator~\cite{mogren2016c,esteban2017real,takahashi2019modeling}. However, these GAN-based approaches only model the joint distribution $p(\mathbf{x}_{1:T})$, but fails to take the transaction dynamics $p(\mathbf{x}_t |\mathbf{x}_{1:t-1})$ into account. 
TimeGAN~\cite{yoon2019time} addresses this issue by estimating and training this conditional density in an internal latent space.

Apart from a few early efforts (e.g., \cite{fabius2014variational}), the VAE-based time series generator is less investigated. Some of them, such as Z-forcing~\cite{alias2017z}), even encode the future information in the autoregressive structure, thereby violating the underlying principles of Granger causality~\cite{Granger1969} that cause happens prior to its effect. The recently developed TimeVAE~\cite{desai2021timevae} uses convolutional neural networks in both encoder and decoder, and adds a few parallel blocks in the decoder where each block accounts for a specific temporal property such as trend and seasonality. However, the building blocks introduce a set of new hyper-parameters which are hard to determine in practice. 

In this work, we also develop a new VAE-based time series generative model. Compared to the above mentioned approaches, our distinct properties include: 1) the ability to discover Granger causality, which makes the model itself more transparent than other baselines; 2) the ability to explicitly model conditional density $p(\mathbf{x}_t |\mathbf{x}_{1:t-1})$; and 3) a rigorous guarantee on the generation process to obey the underlying principles of Granger causality.

\subsection{Causal Discovery of Time Series}

Substantial efforts have been made on the causal discovery of a $M$-variate time series $\mathbf{x}=(\mathbf{x}^1,\mathbf{x}^2,\cdots,\mathbf{x}^M)$, where the goal is to discover, from the observational data, the causal relations between different dimensions of data in different time instants, e.g., if $\mathbf{x}^p$ causes $\mathbf{x}^q$ in time $t$ with a lag $\tau$?

Different types of causal graphs can be considered for time series~\cite{assaad2022survey}. Here, we consider recovery of a Granger causal graph, which separates past observations and present values of each variable and aims to discover all possible causations from past to present. Formally, the Graph causal graph is defined as:

\begin{definition*}[Granger Causal Graph]
Let $\mathbf{x}=(\mathbf{x}^1,\mathbf{x}^2,\cdots,\mathbf{x}^M)$ be a $M$-dimensional time series of length $T$,
where, for time instant $t$, each $\mathbf{x}_t$ is a vector $\mathbf{x}_t=[\mathbf{x}_t^1,\mathbf{x}_t^2,\cdots,\mathbf{x}_t^M]$ in which $\mathbf{x}_t^p$ represents a measurement of the $p$-th time series at time $t$. 
Let $G=(V,E)$ the associated Granger causal graph with $V$ representing the set of nodes and $E$ the set of edges. The set $V$ consists of the set of $M$ dependent time series $\mathbf{x}^1,\mathbf{x}^2,\cdots,\mathbf{x}^M$. There is an edge connects node $\mathbf{x}^p$ to $\mathbf{x}^q$ if: 1) for $p\neq q$, the past values of $\mathbf{x}^p$ (denoted $\mathbf{x}_{t-}^p$) provide unique, statistically significant information about the prediction of $\mathbf{x}_t^q$; and 2) for $p=q$, $\mathbf{x}_{t-}^q$ causes $\mathbf{x}_t^q$ (i.e., self-cause). 
\end{definition*}

Note that, the Granger causal graph may have self-loops as the past observations of one time series always cause its own present value. Hence, it does not need to be acyclic.

The approaches on causal discovery of time series are diverse. Interested readers can refer to~\cite{assaad2022survey} for a comprehensive survey. In the following, we briefly introduce the basic idea of Granger causality~\cite{Granger1969} and its recent advances. 

Wiener was the first mathematician to introduce the notion of ``causation” in time series~\cite{Wiener.N1956}. According to Wiener, the time series or variable $\mathbf{x}$ causes another variable $\mathbf{y}$ if, in a statistical sense, the prediction of $\mathbf{y}$ is improved by incorporating information about $\mathbf{x}$. However, Wiener’s idea was not fully developed until $1969$ by Granger~\cite{Granger1969}, who defined the causality in the context of linear multivariate auto-regression (MVAR) by comparing the variances of the residual errors with and without considering $\mathbf{x}$ in the prediction of $\mathbf{y}$. Not surprisingly, the basic idea of Granger causality can be extended to non-linear scenario by the kernel trick~\cite{marinazzo2008kernel} or by fitting locally linear models in the reconstructed phase space~\cite{chen2004analyzing}. The deep neural networks have been leveraged recently to identify Granger causality. Neural Granger causality~\cite{tank2021neural} is the first method that learns the causal graph by introducing sparsity constraints on the weights of autoregressive networks. The Temporal Causal Discovery Framework (TCDF)~\cite{nauta2019causal} uses an attention mechanism within dilated depthwise convolutional networks to learn complex non-linear causal relations and, in special cases, hidden common causes.

Information-theoretic measures, such as directed information~\cite{massey1990causality} and transfer entropy (TE)~\cite{schreiber2000measuring}, provide an alternative model-free approach to quantify the directed information flow among stochastic processes. Specifically, TE is defined as the conditional mutual information $I(\mathbf{y}_t;\mathbf{x}_{t-} |\mathbf{y}_{t-})$. However, TE is incapable to quantify instantaneous causality~\cite{amblard2012relation} and notoriously hard to estimate, especially in high-dimensional space. Recently, \cite{de2019data} relies on the matrix-based R\'enyi's $\alpha$-order entropy~\cite{giraldo2014measures} to estimate TE and achieves compelling performances. 
Interestingly, TE is equivalent to Granger causality in MVAR for Gaussian variables~\cite{barnett2009granger}. Essentially, both definitions can be regarded as comparing the model with and without considering the intervening variable $\mathbf{y}$~\cite{chen2021wiener}.

We provide in the supplementary material a table of other related works with additional details. Note, however, that none of the mentioned causal inference approaches can be used for time series generation. 

% We also summarize the difference between Granger causal graph and other popular causal graph such as the window causal graph. 

\section{Causal Recurrent Variational Autoencoder}

\subsection{Problem Formulation and Objectives}
Our high-level objective is to learn a distribution $\hat{p}(\mathbf{x}_{1:T})$ that matches well the true joint distribution $p(\mathbf{x}_{1:T})$. From a generative model perspective, this is achieved by sampling from a simple and tractable distribution $p(\mathbf{z})$ and then map to a more complicated distribution $\hat{p}(\mathbf{x}_{1:T})$. Usually, it is difficult to model $p(\mathbf{x}_{1:T})$ depending on its dimension $M$, length $T$ and possibly non-stationary nature.
To this end, we can apply the autoregressive decomposition $p(\mathbf{x}_{1:T}) = \prod^{T}_{t=1}p(\mathbf{x}_{t} \mid \mathbf{x}_{1:t-1})$ to infer the sequence iteratively~\cite{west2006bayesian}. The objective reduces to learn a conditional density $\hat{p}(\mathbf{x}_{t} \mid \mathbf{x}_{1:t-1})$ that equals to the true density $p(\mathbf{x}_{t} \mid \mathbf{x}_{1:t-1})$. Hence, our first objective is:
\begin{equation}
\min_{\hat{p}} \mathcal{D}(p(\mathbf{x}_{t} \mid \mathbf{x}_{1:t-1})\vert \vert \hat{p}(\mathbf{x}_{t} \mid \mathbf{x}_{1:t-1})),
\label{eq.ob1}
\end{equation}
for any $t$, where $\mathcal{D}$ is the divergence between distributions. 

%Meanwhile, We also wish to infer synthetic data using  $\mathcal{Z}$, so the first objective can be written as:

Our second objective is straightforward. Suppose $\mathbf{x}^1,\mathbf{x}^2,\cdots,\mathbf{x}^M$ are intrinsically correlated by a Granger causal graph $G=(V,E)$, we can characterize $G$ by its (unweighted) adjacency matrix $A$, whose $(u,v)$-th entry is defined as:
\begin{equation}
  A_{u,v} =
    \begin{cases}
      1 & \text{$\mathbf{x}_{t-}^v$ causes $\mathbf{x}_t^u$; i.e., edge $(u,v)\in E$}\\
      0 & \text{otherwise}.
    \end{cases}       
\end{equation}

Now, let $\text{PA}(\mathbf{x}^p)$ denote the set of parents (or causes) of $\mathbf{x}^p$ in $G$ (i.e., the non-zero elements in the $p$-th row of $A$), motivated by the additive noise model with nonlinear functions~\cite{hoyer2008nonlinear,chu2008search}, we can represent $\mathbf{x}_t^p$ as follows:
\begin{equation}
\mathbf{x}_t^p=f_p \left(\mathbf{x}_{t-}^p,\text{PA}(\mathbf{x}^p)_{t-} \right)+ \varepsilon_t^p,
\label{ob.2}
\end{equation}
in which $\mathbf{x}_{t-}^p$ denotes past observations of $\mathbf{x}_t^p$, $\text{PA}(\mathbf{x}^p)_{t-}$ denotes past observations of cause variables of $\mathbf{x}^p$, $\varepsilon_t^p$ are jointly independent over $p$ and $t$ and, for each $p$, \emph{i.i.d.}, in $t$.

Therefore, our second objective is to learn $f_p$ for each $\mathbf{x}^p$ and infer the matrix $A$.

\subsection{Methodology}
\begin{figure*}[t]
\centering
\noindent\begin{subfigure}[b]{0.28\textwidth}
    \centering 
    \includegraphics[width=1.0\textwidth]{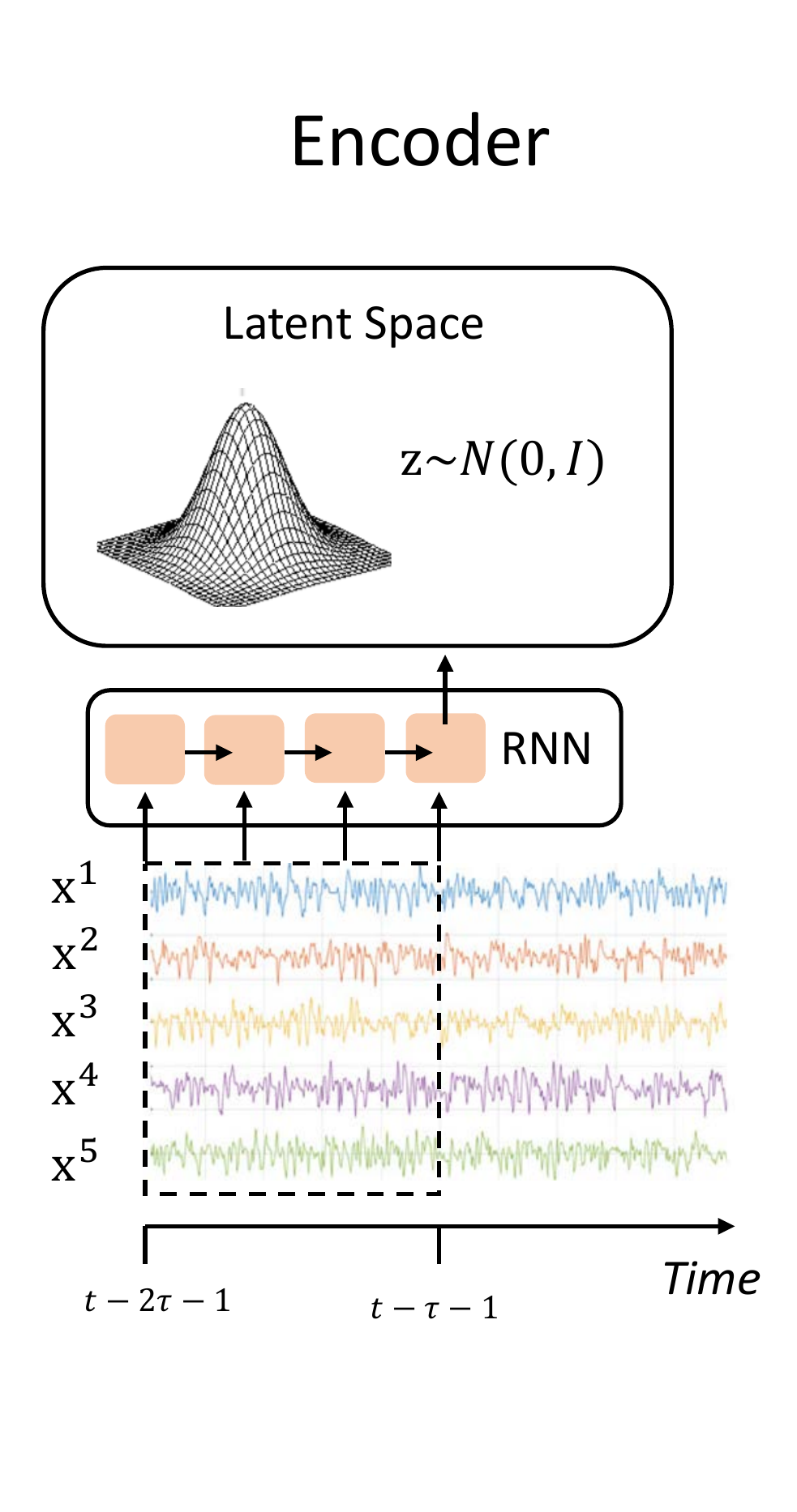}

\caption{The CR-VAE encoder approximates the intractable posterior $p(\mathbf{z}|\mathbf{x})$. We extract a time segment $\mathbf{x}_{t-2\tau-1:t}$ as our training data, and and use the first half clip $\mathbf{x}_{t-2\tau-1:t-\tau-1}$ as the input to encoder.}
\end{subfigure}%
\hspace{1.1cm}
\noindent\begin{subfigure}[b]{0.30\textwidth}
    \centering
    \includegraphics[width=1.0\textwidth]{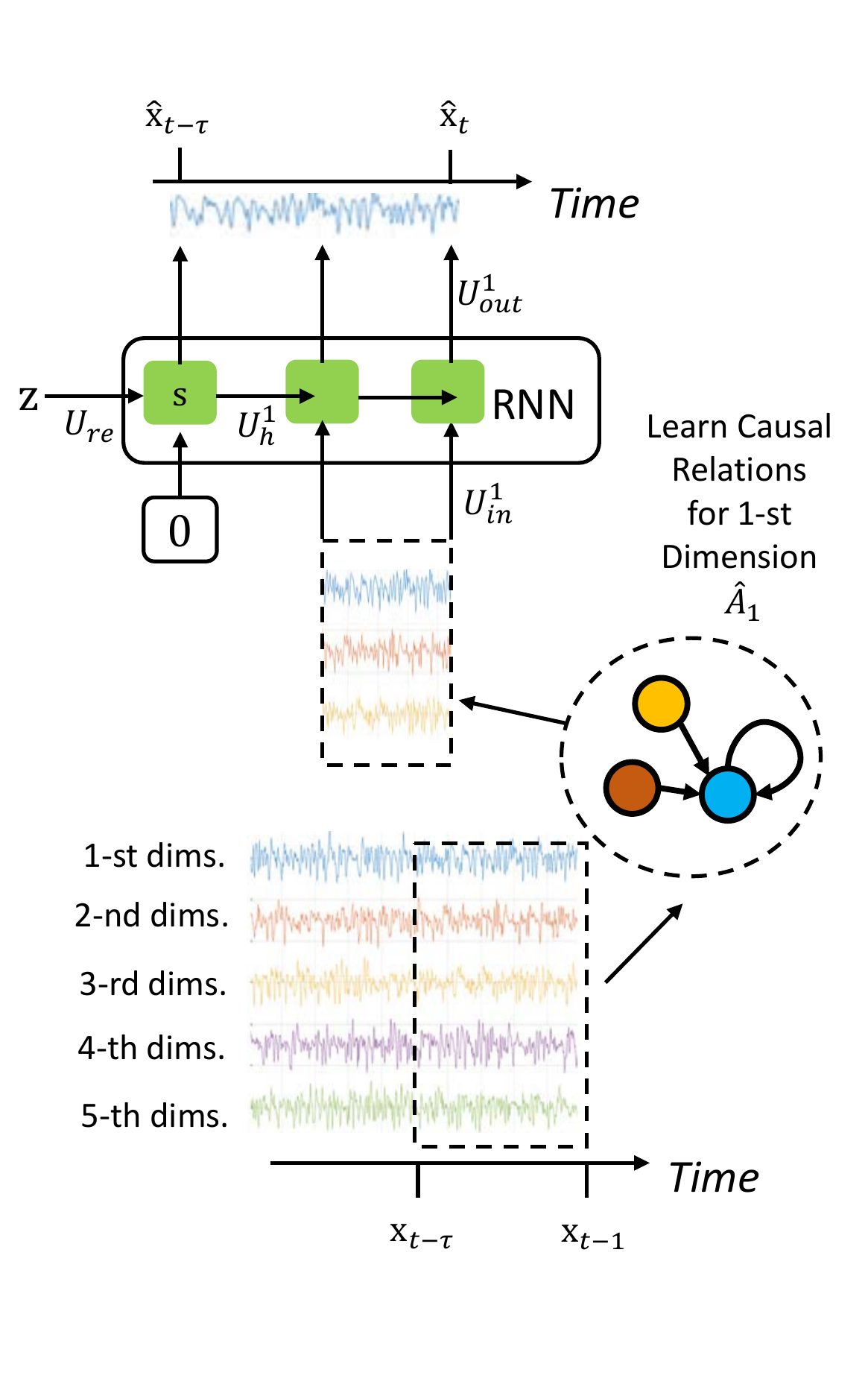}

\caption{The first head of the decoder, which predicts the 1-st dimension of $\mathbf{x}$. The second half of the time clip is used as the decoder inputs. RNN inputs are determined by the estimated causal relations $\hat{A}_1$ (the first row of $\hat{A}$).}
\end{subfigure}%
\hspace{1.1cm}
\noindent\begin{subfigure}[b]{0.28\textwidth}
    \centering
    \includegraphics[width=1.0\textwidth]{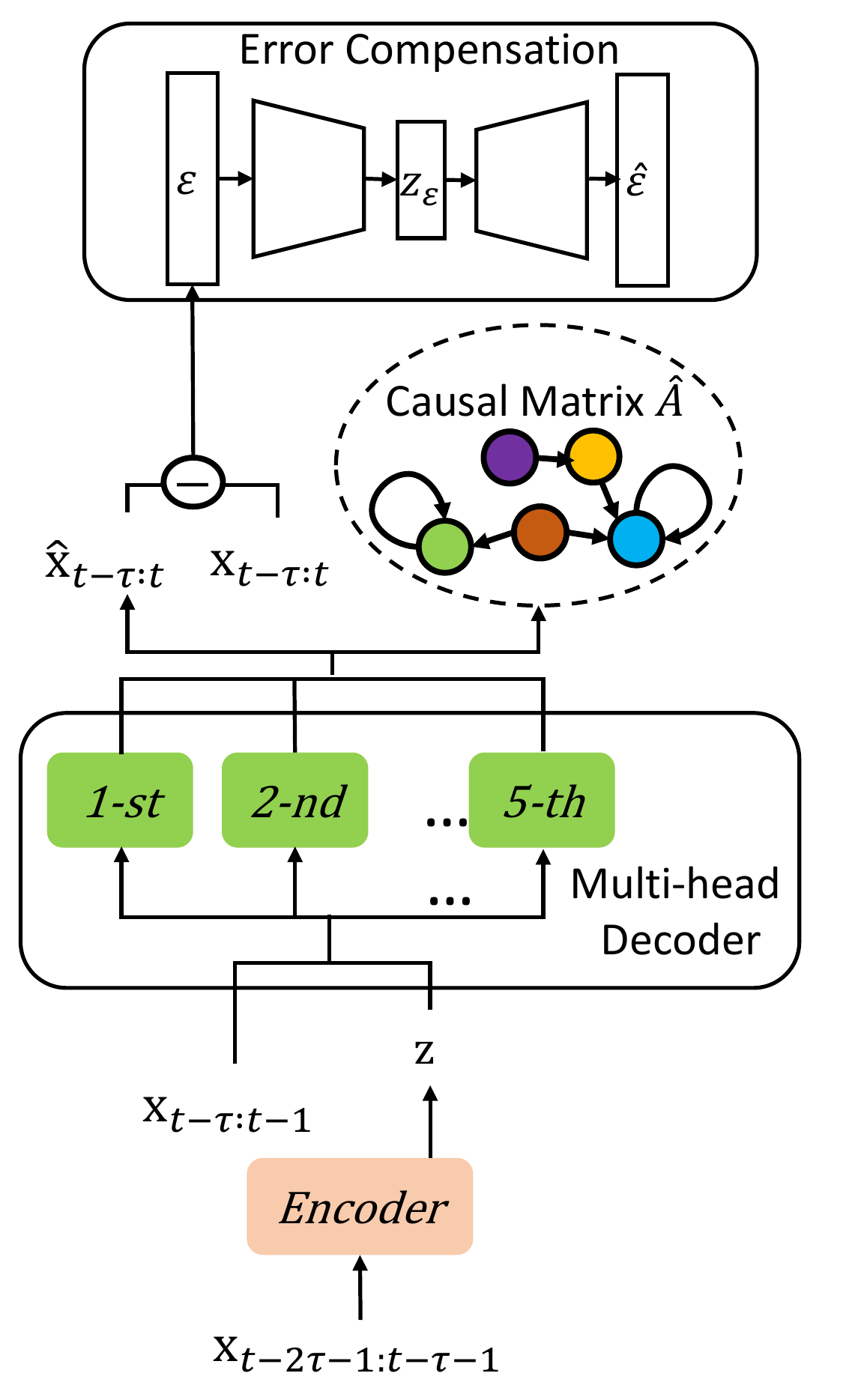}
\caption{The pipeline of CR-VAE. The multi-head decoder predicts $5$ variables separately in training; The compensation network approximates $\varepsilon_{t}$ in Eq. (\ref{ob.2}). The adjacency matrix $\hat{A}$ of Granger causal graph $\hat{A}$ can be obtained by stacking, i.e., $\hat{A} = \{\hat{A}_1; \hat{A}_2;\hat{A}_3;\hat{A}_4;\hat{A}_5\}$.}
\end{subfigure}%
\caption{Architecture of \emph{Causal Recurrent Variational Autoencoder} (CR-VAE). We use a $5$-variate time series as an example.}
\label{fig.struc}
\end{figure*}

Both encoder and decoder of our causal recurrent variational autoencoder (CR-VAE) consist of recurrent neural networks such that the hidden state is calculated based on the previous state and the observation at the current time instant. A CR-VAE model with time lag $\tau$ can be written as:
\begin{equation}
\hat{\mathbf{x}}_{t-\tau:t} = D_{\theta}(\mathbf{x}_{t-\tau : t-1}, E_{\phi}(\mathbf{x}_{t-2\tau-1 : t-\tau-1})) + \varepsilon_t,
\label{eq.feedforward}
\end{equation}
where $D_{\theta}, E_{\phi}$  represent decoder and encoder, which are parameterized by $\theta$ and $\phi$, respectively; $\varepsilon_t$ is the additive innovation term that has no specific distributional assumption. 

Our CR-VAE takes as input the segment $\mathbf{x}_{t-2\tau-1 : t-\tau-1}$, and aims to predict or reconstruct the segment $\mathbf{x}_{t-\tau:t}$ step-wisely. In this way, we obey the principle of Granger causality by preventing encoding future information before decoding. By contrast, other popular recurrent VAEs, such as VRAE \cite{fabius2014variational}, SRNN \cite{fraccaro2016sequential}, Z-forcing \cite{alias2017z}, use the same time segment in both encoder and decoder, thereby encoding future information in the recurrent structure.
Our training model is in fact motivated by T-forcing~\cite{williams1989learning} and other predictive autoregressive models (e.g., \cite{litterman1986forecasting, bengio2015scheduled}) that use the real past observations to predict the current value in the sequence.

Another distinction between CR-VAE and other popular recurrent VAEs \cite{fabius2014variational, NIPS2015_b618c321,alias2017z} is that our decoder has multiple heads, in which the $p$-th head is used to approximate $f_p \left(\mathbf{x}_{t-}^p,\text{PA}(\mathbf{x}_t^p)_{t-} \right)$ in Eq.~(\ref{ob.2}). Then, the full vector $\hat{\mathbf{x}_t}$ is constructed by stacking the output of all $M$ heads. In short, the term $D_{\theta}(\mathbf{x}_{t-\tau : t-1}, E_{\phi}(\mathbf{x}_{t-2\tau-1 : t-\tau-1}))$ learns to estimate a collection of $\{f_p(\cdot) | p =1,2 ... M\}$.

Fig.~\ref{fig.struc}(a) shows the structure of our encoder. 
Let $h$ be hidden states in our encoder, our encoder can be written as:
\begin{equation}
\begin{aligned}
    & \mathbf{h}_{t} = \text{tanh}(W_{in}\mathbf{x}_{t} + W_{h}\mathbf{h}_{t-1} + b),\\
    & \mu = W_{\mu}\mathbf{h}_{t-\tau-1} + b_{\mu},\\
    & \log(\sigma) = W_{\sigma}\mathbf{h}_{t-\tau-1} + b_{\sigma},
\end{aligned}
\label{eq.encoder}
\end{equation}
where $\{W_{in}, W_{h}, W_{\mu}, W_{\sigma} \} \subseteq \theta$. $W_{in}$ and $W_{h}$ are the weight matrix for inputs and hidden states, respectively; $W_{\mu}$ and $W_{\sigma}$ are the weights to compute mean and standard deviation of the learned Gaussian distribution, respectively; $b$ denotes the bias. 

Fig.~\ref{fig.struc}(b) shows the structure of the $1$-st head of our decoder, in which we use a $5$-variate time series as an example. The collection of all heads explicitly models $p(\mathbf{x}_t|\mathbf{x}_{1:t-1})$. Let $\mathbf{s}$ be the hidden state in our decoder. The initial state of decoder is sampled from the Gaussian distribution parameterized by $\mu$ and $\sigma$. More formally, we have:
\begin{equation}
\begin{aligned}
    & \mathbf{s}_{t-\tau} = \text{tanh}((U_{re}(\mu + \sigma\mathbf{z}) + b_{re}),\\ 
    &\mathbf{z} \sim N(0,I),\\
    &\mathbf{s}^{p}_{t} = \text{tanh}(U^{p}_{in}\mathbf{x}_{t-1} + U^{p}_{h}\mathbf{s}^{p}_{t-1} + b^{p}),\\
    &\hat{A}_p = U^{p}_{in},\\
    & \hat{\mathbf{x}}^{p}_{t} = U^{p}_{out}\mathbf{s}^{p}_{t} + b^{p}_{out},
\end{aligned}
\label{eq.decoder}
\end{equation}
where $\{U^{p}_{in},U_{re}, U^{p}_{h}, U^{p}_{out} \}\subseteq \phi$. $U^{p}_{in}$ and $U^{p}_{h}$ denote weight matrix for inputs and hidden states of the $p$-th head in the decoder. Similarly, $U_{re}$ and $U^{p}_{out}$ denote weights for reparameterization and output layers; $\hat{A}_p$ is the $p$-th row of the estimated adjacency matrix $\hat{A}$ of the Granger causal graph, which includes all cause variables of the $p$-th variable. Note that, we use a single-layer vanilla RNN as an example for simplicity. In practice, we use gated recurrent units (GRUs)~\cite{cho2014properties} to improve modeling ability.

Fig.~\ref{fig.struc}(c) shows the pipeline of full model. An error compensation network is applied to model an additive innovation term $\varepsilon_t$ in Eq.~(\ref{eq.feedforward}), thus further improving sequence generation performance. We assume $\varepsilon_t$ is not predictable or inferable by the information of the past. To compensate for it, another recurrent VAE parameterized by $\{\psi, \omega\}$, is utilized to estimate the additive noise $\varepsilon_{t-\tau:t}$. Here, we use the same sequence as inputs for both encoder and decode, since it does not disentangle the obtained causal graph $\hat{A}$.

\subsubsection{CR-VAE Loss Function}
In order to estimate $A$ in the process of learning, we invoke a sparsification trick which is first shown in neural Granger causality (NCG)~\cite{tank2021neural} and has also been used in recent causal discovery literature~\cite{marcinkevivcs2021interpretable, liu2020ec}. The essential sparsification trick is simple. It assumes that the causal matrix $A$ is sparse and applies sparsity-inducing penalty to $\hat{A}$. It shares the theme with the traditional prediction-based Granger causality -- the causes help predict the effects. Therefore,  we train CR-VAE by minimizing the following penalized  loss function with the stochastic gradient descent (SGD) and proximal gradients:
\begin{equation}
\begin{aligned}
&\mathcal{L}(\theta,\phi) = \sum_{p = 1}^{M} \left [\mathbb{E}_{q_{\phi}(\mathbf{z}|\mathbf{x}_{t-2\tau-1 : t-\tau-1})} [{\rm log}p_{\theta}(\mathbf{x}^{p}_{t-\tau : t}|\mathbf{x}_{t-\tau : t-1},\mathbf{z})]\right ]\\
& - \mathcal{D}_{KL}(q_{\phi}(\mathbf{z}|\mathbf{x}_{t-2\tau-1 : t-\tau-1}) \vert \vert p(\mathbf{z})) + \lambda R(\hat{A}),
% (\mathbf{z}|\mathbf{x}^{m}_{t-2\tau-1 : t-\tau-1})} + \frac{1+log\sigma^2 -\mu^2-\sigma^2}{2}\\
% % &+R(\hat{G}) 
\end{aligned}
\label{eq.key}
\end{equation}
where $p(\mathbf{z})$ is a standard Normal distribution. The loss function includes three terms: (1) the mean squared error (MSE) loss pushes the model towards high fidelity to sample space; (2) the KL divergence term ensures that the latent space behaves as a Gaussian emission; and (3) a sparsity-inducing penalty term $R(\cdot)$ on $\hat{A}$ with a hyper-parameter $\lambda$. The first two terms correspond to our multi-head recurrent VAE. 

Meanwhile, the additional $\varepsilon$ compensation network of CR-VAE is trained by minimizing:
\begin{equation}
\begin{aligned}
&\mathcal{L}(\psi,\omega) =  \mathbb{E}_{q_{\omega}(\mathbf{z}_{\varepsilon}|\varepsilon_{t-\tau : t})} {\rm log}p_{\psi}(\varepsilon_{t-\tau : t}|\mathbf{z}_{\varepsilon})\\
 & - \mathcal{D}_{KL}(q_{\omega}(\mathbf{z}_{\varepsilon}|\varepsilon_{t-\tau : t}) \vert \vert p(\mathbf{z}_{\varepsilon})).\\
% &\left \| \varepsilon_{t-\tau:t} - \hat{\varepsilon}_{t-\tau:t} \right \|^2_2 + \frac{1+log\sigma_\varepsilon^2 -\mu_\varepsilon^2-\sigma_\varepsilon^2}{2}
\end{aligned}
\label{eq.compensation}
\end{equation}

This is a standard VAE objective function, and its update does not affect the result of $\hat{A}$.

\subsubsection{CR-VAE Learning and Optimization}
The ideal choice of $R(\cdot)$ is the $\ell_0$ norm which represents the number of non-zero elements, but the optimization of $\ell_0$ norm in neural network is still challenging. Hence we apply $\ell_1$ norm, and the Eq.~(\ref{eq.key}) becomes a typical lasso problem. Proximal gradient descent is the most popular method for non-convex lasso objective optimization. In practice, we use iterative shrinkage-thresholding algorithms (ISTA)~\cite{daubechies2004iterative,chambolle1998nonlinear} with fixed step size. The feature of thresholding leads to exact zero solutions in $U^p_{in}$.
More formally, we start to update weights $U^{p}_{in}$ iteratively from $U^{p}_{in}(0)$:

\begin{equation}
\begin{aligned}
U^{p}_{in}(i+1) = {\rm prox}_{\gamma}(U^{p}_{in}(i) - \gamma\nabla \mathcal{L}_{c}(U^{p}_{in}(i))),
\end{aligned}
\label{eq.prox}
\end{equation}
where ${\rm prox}_{\gamma}$ denotes the proximal operator with step size $\gamma$; $\mathcal{L}_{c}$ denotes the convex part of the loss function that is the first and second term in Eq.~(\ref{eq.key}). During training, two separate optimization methods are implemented: proximal gradient on the weights of input layers $U^{p}_{in}$, and stochastic gradient descent (SGD) on all other parameters. 
% Besides, We can also propose different criteria for early stopping based on prior knowledge, e.g., sparsity level of the causal matrix.

It performs well in causal discovery, but reduces the generation performance since we invoke $\ell_{1}$ norm as our sparsity penalty.
% Recall that $L_{0}$ is the optimal choice, and $L_1$ norm is simply a substitute to solve the optimization problem. The objective of sparsity is purely encouraging zero solutions, but $L_1$ norm also reduce the absolute value of $U^{p}_{in}$, thereby significantly limiting the solution space of $\mathcal{L}$. For example, let $I$ be a identity matrix and $\mathcal{L}(2I) < \mathcal{L}(I)$, but we have $L_{1}(I)<L_{1}(2I)$, $L_{0}(I)=L_{0}(2I)$. In this case of $L_1$, the algorithm tends to search solutions near $I$ instead of $2I$.
To solve this problem, we propose a two-stage training strategy inspired by network pruning \cite{liang2021pruning}, as summarized in Algorithm \ref{alg:algorithm_train}.  In line 1-8, we train the CR-VAE with both proximal gradient and SGD to obtain the sparse causal graph (Phase I); In line 9, we fix the zero elements; In line 10-13, we continue training CR-VAE with SGD to improve generation performance (Phase II).
% First, we train our model using Eq.~(\ref{eq.key}) with $\ell_1$ norm and stop when the criteria is met or convergence. 
% % We refer to the phase as structure learning stage. 
% Then we fix all zero weights in $U_{in}$, and train other weights using SGD.
% We refer to the phase as value learning stage, that 

%Algorithm \ref{alg:algorithm_train} summarises the training procedure of the proposed framework. 

Once the CR-VAE has been trained, we can obtain the estimated causal matrix by stacking $U^{p}_{in}$, and it can also be used for synthetic sequence generation. During sequence generation, we independently sample two sets of noise $\mathbf{z}$ and $\mathbf{z}_{\varepsilon}$, and then feed them to the decoders to iteratively generate a time series of arbitrary length.

\begin{algorithm}[tb]
\caption{Training pipeline of CR-VAE}
\label{alg:algorithm_train}
\textbf{Require}: The multivariate time sequence $\{\mathbf{x}_t\}^{T}_{t = 1}$ with $M$ dimensions; model lag $\tau$; step size $\gamma$ for ISTA; initialize $\{\theta, \phi, \psi, \omega\}$ \\
\textbf{Output}: Estimated adjacency matrix $\hat{A}$ of Granger causal graph, the trained $\{\theta, \phi, \psi, \omega\}$.
\begin{algorithmic}[1] %[1] enables line numbers

\WHILE{not stop criteria or converge}
\STATE Sample a batch of $\mathbf{x}_{t-2\tau-1 : t}$ from $\{\mathbf{x}_t\}^{T}_{t = 1}$.
\STATE Compute the gradients of $\mathcal{L}_{c}$, i.e, convex terms in Eq.~(\ref{eq.key}).
\STATE Update $\theta, \phi$ except $U_{in}$ using SGD.
\STATE Update $U_{in}$ using proximal gradient descent in Eq.~(\ref{eq.prox}).
\STATE Update $\psi, \omega$ by minimizing Eq.~(\ref{eq.compensation}).

\ENDWHILE
\STATE Stack $U_{in}$ to obtain the $M \times M$ estimated causal matrix $\hat{A}$.
\STATE Prune out all zeros edges in $U_{in}$ based on $\hat{A}$.
\WHILE{not or converge}
\STATE Compute the gradients of $\mathcal{L}_{c}$.
\STATE Update $\theta, \phi$ using SGD.
\STATE Update $\psi, \omega$ by minimizing Eq.`(\ref{eq.compensation}).

\ENDWHILE
\STATE \textbf{return} $\hat{G}$, trained $\{\theta, \phi, \psi, \omega\}$.
\end{algorithmic}
\end{algorithm}

\section{Experiments}
We first evaluate CR-VAE on a synthetic linear autoregressive process to illustrate the importance of each module. We then compare CR-VAE with several state-of-the-art (SOTA) approaches on four benchmark time series datasets to demonstrate its advantages in both causal discovery and synthetic time series generation. 

\subsection{Linear Autoregressive Process}

%In order to task with both causal discovery and generation, CR-VAE is characterized by (1)  unidirectional inputs and (2) compensation VRAE. To illustrate the importance of each contribution, we first experiment on sequences from linear multivariate autoregressive process. There are 10 dimensions in the process, and the lag (order) is 3. 

CR-VAE features a few special designs over the traditional RVAE, such as the multi-head decoder, the unidirectional inputs and the error-compensation module. To illustrate the importance of each component to the performance gain, we first test CR-VAE on a synthetic linear multivariate autoregressive process with $10$ dimensions and maximum lag of $3$. More formally:
\begin{equation}
\mathbf{x}_{t} =  a_1\mathbf{x}_{t-1} + a_2\mathbf{x}_{t-2} + a_3\mathbf{x}_{t-3} + \varepsilon_t,
\label{eq.linear}
\end{equation}
where $\varepsilon_t \sim N(0,I)$; the true causal matrix $G$ can be represented by all non-zeros elements of the $a_1+a_2+a_3$.

\subsubsection{Unidirectional Inputs: Don't Peep on the Future.}
The original VRAE and its recent variants \cite{alias2017z,fabius2014variational} use $\mathbf{x}_{t-\tau : t-1}$ as the input of both encoder and decoder.
This way, information of the entire sequence is encoded before decoding. Those approaches estimate $p(x_{t}| x_{1:T} )$, rather than $p(x_{t}| x_{1:t-1} )$, i.e., the future input values at time $t$ cannot be used in the conditional variable. This is called causal conditioning as proposed by Massey and Kramer~\cite{kramer1998causal}. From a causal discovery perspective, it violates the underlying principles of Granger causality by ``peeping on the future" and hence can never identify causality in the sense of Granger.

\begin{figure}[htp]
	\centering
		\includegraphics[scale=.4]{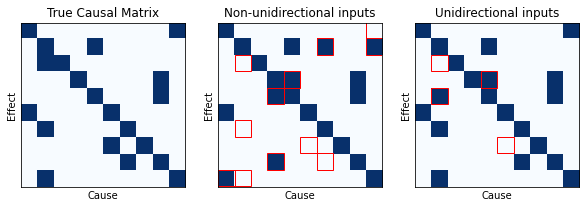}
	\caption{(a) shows the true adjacency matrix $A$ of Granger causal graph; (b) and (c) show the recovered adjacency matrix $\hat{A}$ by non-unidirectional CR-VAE and our CR-VAE. False causal relations are highlighted by red rectangles.}
	\label{MINE}
\label{fig:linear_GC}
\end{figure}

%In original VRAE or other popular non-unidirectional recurrent VAE approaches, $x_{t-\tau : t-1}$ is the input of both encoder and decoder. 
%  follow \cite{fabius2014variational} and
To support our argument, we take $\mathbf{x}_{t-\tau : t-1}$ as the input of encoder (rather than $\mathbf{x}_{t-2\tau-1 : t-\tau-1}$). We term this modification the non-unidirectional CR-VAE. As shown in Fig.~\ref{fig:linear_GC}, CR-VAE identifies majority of true causal relations, whereas its non-unidirectional baseline, whose encoder peeps on future values, fails to discover causal directions between most pairs of time series. 

\subsubsection{Error-Compensation Network.} 
\begin{figure}[htb]

% \begin{minipage}[b]{0.48\linewidth}
%   \centering
%   \centerline{\includegraphics[width=4cm]{LaTeX/pca_linear_without.png}}
% %  \vspace{2.0cm}
%   \centerline{PCA without compensation }\medskip
% \end{minipage}
% %
% \hfill
% \begin{minipage}[b]{0.48\linewidth}
%   \centering
%   \centerline{\includegraphics[width=4cm]{LaTeX/pca_linear_with.png}}
% %  \vspace{2.0cm}
%   \centerline{PCA with compensation}\medskip
% \end{minipage}

\begin{minipage}[b]{.48\linewidth}
  \centering
  \centerline{\includegraphics[width=4.0cm]{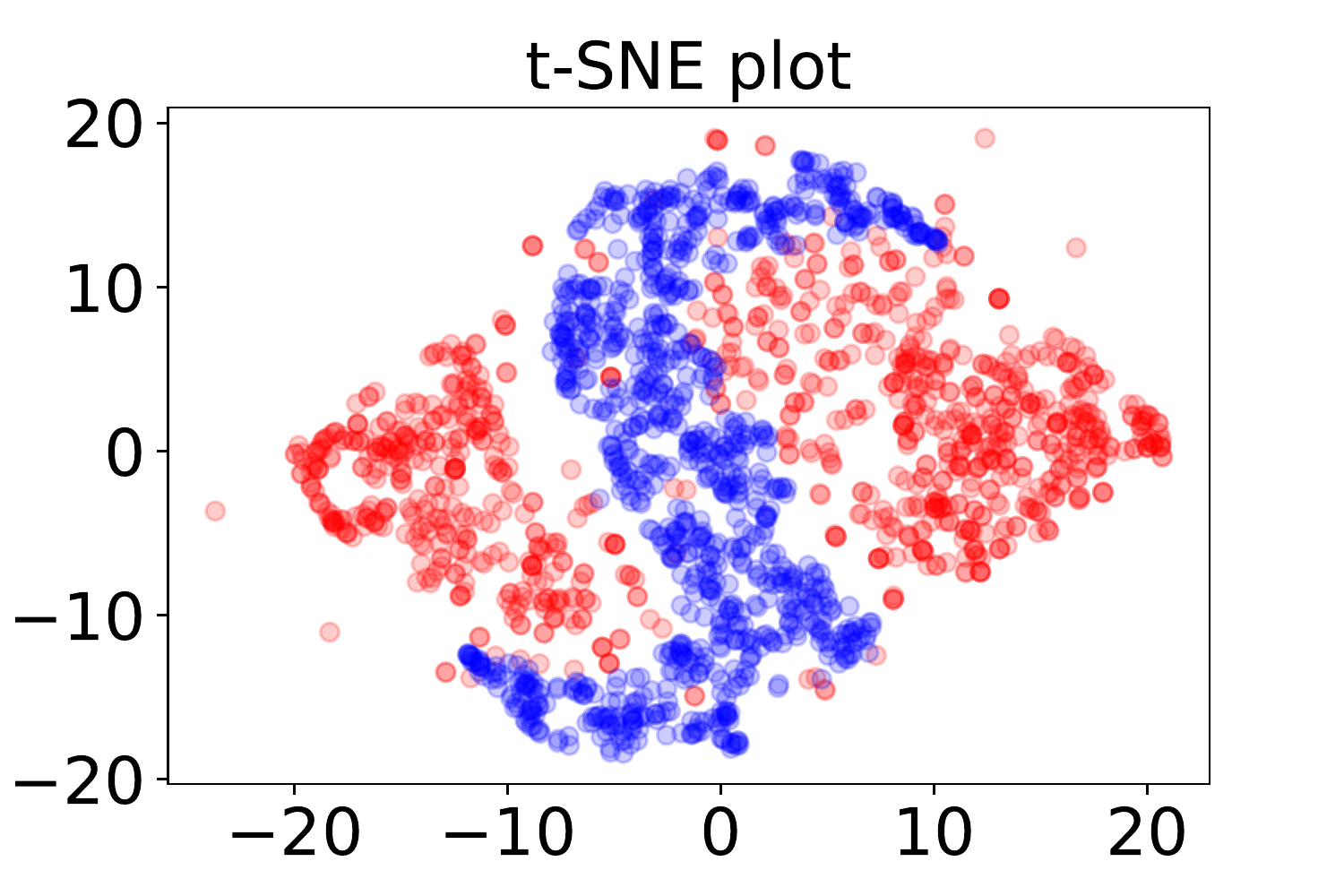}}
%  \vspace{1.5cm}
  \centerline{t-SNE without compensation}\medskip
\end{minipage}
\hfill
\begin{minipage}[b]{0.48\linewidth}
  \centering
  \centerline{\includegraphics[width=4.0cm]{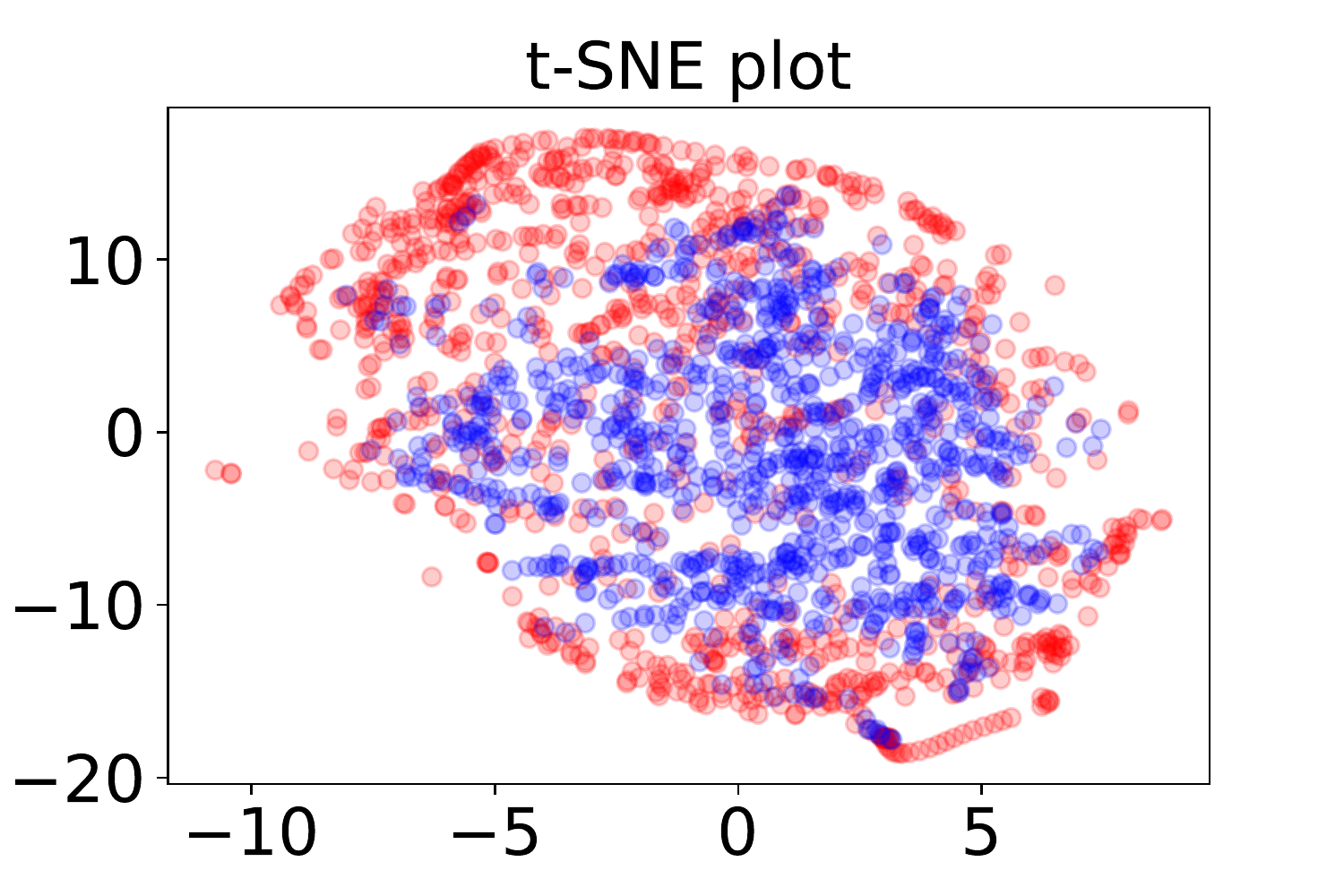}}
%  \vspace{1.5cm}
  \centerline{t-SNE with compensation}\medskip
\end{minipage}

\caption{t-SNE visualization on the illustrative linear system: red samples correspond to real time series, whereas blue samples correspond to synthetic time series.}
\label{fig:linear_compensation}
\end{figure}

We then validate the indispensability of error-compensation network. We compare the time series generation results of the original CR-VAE and its degraded version without error-compensation. We use t-SNE~\cite{van2008visualizing} to visualize the generated samples. 
A good generative model should lead to similar synthetic distribution to real data distribution. 
As shown in Fig.~\ref{fig:linear_compensation}, the error-compensation network leads to a significant performance gain. 
In fact, samples generated by CR-VAE without error-compensation converge quickly to values nearly zero. This makes sense for a linear AR process, because it can only diverge to $\infty$ or converge to nearly zero if we omit $\varepsilon_t$ in Eq.~\ref{eq.linear}. In our case, we tune values of $\{a_1,a_2,a_3\}$ to avoid divergence, and the true $\mathbf{x}_{t} =  a_1\mathbf{x}_{t-1} + a_2\mathbf{x}_{t-2} + a_3\mathbf{x}_{t-3}$ did converge. In other words, the degraded CR-VAE captures the dynamics $p(\mathbf{x}_{t} \mid \mathbf{x}_{1:t-1})$, but ignores $\varepsilon_t$.

% because the CR-VAE without error compensation just capture the dynamics $p(x_{t} \mid x_{1:t-1}, z)$, but ignore $\varepsilon_t$
% .\textcolor{red}{[I still feel this explanation is too weak!]} 
%In fact, samples generated by non-compensated model converge to values close to zero (not exactly zero). It is what will happen if we omit the $\varepsilon_t$ in Eq.~\ref{eq.linear}, when we create real samples of the linear system. 
%In summary, non-compensated CR-VAE can only capture the dynamics $p(x_{t} \mid x_{1:t-1}, z)$, but ignore $\varepsilon_t$. Therefore, It is crucial to introduce the compensation network for $\varepsilon_t$ estimation.

\subsection{Experiments on Different Data}
%We test the performance of CR-VAE across time-series datasets that are widely applied in the field of Granger causal discovery. The following four data are selected on the basis of our two goals: 
We then systematically evaluate the performances of CR-VAE in causal discovery and time series generation on four widely used time series data. 
\begin{itemize}
\item  \textbf{H\'enon maps:} We select $6$ coupled H\'enon chaotic maps \cite{kugiumtzis2013direct}, whose true causal relation is $x^{i-1} \rightarrow x^{i}$. We generate $2,048$ samples to constitute our training data. Equations can be found in supplementary material.
\item \textbf{Lorenz-96 model:} It is a nonlinear model formulated by Edward Lorenz in $1996$ to simulate climate dynamics~\cite{lorenz1996predictability}. The $p$-dimensional Lorenz-96 model is defined as: 
\begin{equation}
\frac{d \mathbf{x}_t^i}{dt} = (\mathbf{x}^{i+1}-\mathbf{x}^{i-2})\mathbf{x}^{i-1} - \mathbf{x}^i + F,
\end{equation}
where $\mathbf{x}^{-1}=\mathbf{x}^{p-1}$, $\mathbf{x}^{0}=\mathbf{x}^{p}$ and $\mathbf{x}^{p+1}=\mathbf{x}^{1}$. $F$ is the forcing constant which is set to be $10$. We take $p=10$ and generate $2,048$ samples as training data. 
\item \textbf{fMRI:} It is a benchmark for causal discovery, which consists of realistic simulations of blood-oxygen-level-dependent (BOLD) time series~\cite{smith2011network} generated using the dynamic causal modelling functional magnetic resonance imaging (fMRI) forward model\footnote{\url{https://www.fmrib.ox.ac.uk/datasets/netsim/}}. Here, we select simulation no. 3 of the original dataset. It has $10$ variables, and we randomly select $2,048$ observations. 
\item \textbf{EEG:} It is a dataset of real intracranial EEG recordings from a patient with drug-resistant epilepsy\footnote{\url{http://math.bu.edu/people/kolaczyk/datasets.html}} \cite{kramer2008emergent}. We select 12 EEG time series from 76 contacts since they are recorded at deeper brain structures than cortical level. Note, however, that there is no ground truth of causal relation in this dataset.
\end{itemize}

\subsubsection{Causal Discovery Evaluations}
We compare CR-VAE with $4$ popular Granger causal discovery approaches. They are: kernel Granger causality (KGC)~\cite{marinazzo2008kernel} that uses kernel trick to extend linear Granger causality to non-linear scenario; transfer entropy (TE)~\cite{schreiber2000measuring} estimated by the matrix-based R{\'e}nyi's $\alpha$-order entropy functional~\cite{giraldo2014measures}; Temporal Causal Discovery Framework (TCDF)~\cite{nauta2019causal} which integrates attention mechanism into a neural network; Neural Granger causality (NGC)~\cite{tank2021neural}, which is the first neural network-based approach for Granger causal discovery. 

KGC and TE rely on information-theoretic measures (on independence or conditional independence) and post-processing (e.g., hypothesis test), whereas TCDF, NGC and our CR-VAE are neural network-based approaches that obtain causal relations actively and automatically in a learning process. All methods are trained only on one sequence that is stochastically sampled based on lag. We use true lag for KGC and TE and set it to be $10$ for TCDF, NGC and CR-VAE. For each approach, we compare the estimated causal adjacency matrices with respect to the ground-truth and apply areas under receiver operating characteristic curves (AUROC) as a quantitative metric. For neural network-based approaches, we select the estimated causal matrices by searching smallest convex loss. Relevant hyper-parameters of all learnable models are tuned to minimize the loss function. Details can be found in supplementary material.

Table~\ref{table.1} summarizes the quantitative comparison results.
The neural network-based approaches outperform traditional KGC and TE by a considerable margin. This is because traditional approaches are incapable of detecting self-causes.
Our CR-VAE outperforms TCDF in all datasets and achieves similar performance to NGC. This can be expected. Note that, both CR-VAE and NGC apply $\ell_1$ sparsity penalty on network weights to discover causal relations.

\begin{figure}[htp]
	\centering
		\includegraphics[scale=.15]{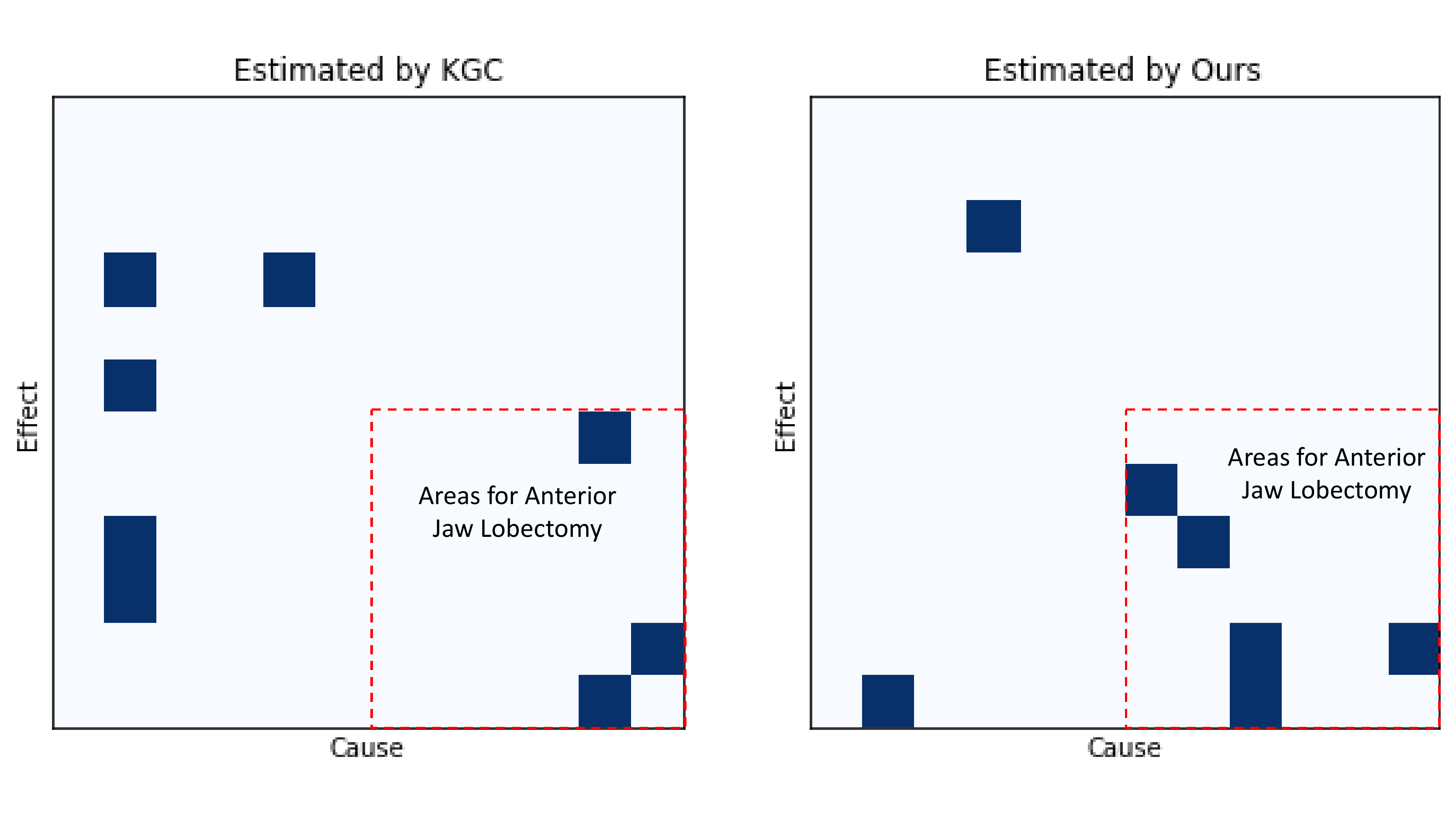}
	\caption{Estimated causal
adjacency matrix (without self-causal elements) of KGC and CR-VAE on EEG data. Our causal elements concentrate in the last six sequence.}
	\label{MINE}
\label{fig:EEG}
\end{figure}

Although the ground-truth causal relation of EEG data is not available, we compare the estimated causal matrices by our method and KGC in Fig.~\ref{fig:EEG}. We observed that most of causal relations in our estimation are concentrated on the last six sequences, whereas the causal elements found by KGC distribute more evenly. Our results make more sense because doctors often perform anterior jaw lobectomy for patients with epilepsy by resecting the last six contact areas \cite{stramaglia2014synergy,kramer2008emergent}. KGC fails to capture this.

%The neural network-based approaches outperform traditional KGC and TE by a considerable margin, because traditional methods mitigate self-causal relationships by estimating conditional risk or probability. 

%In contrast, our method can capture self-causal edges without introducing extra algorithms.

\begin{table}[htb]
\begin{tabular}{cccccc}
Methods & KGC & TE    & TCDF  & NGC   & Ours  \\ \hline
H\'enon   & 0.465   & 0.465 & \underline{0.911} & \textbf{0.960} & \textbf{0.960} \\
Lorenz  &0.631     & 0.408       & 0.871 & \textbf{0.980} & \underline{0.954} \\
fMRI    &0.379     &0.380       & 0.881 & \underline{0.950} & \textbf{0.957} \\
\end{tabular}
\caption{Comparison of causal discovery using AUROC on H\'enon, Lorenz-96 and fMRI. The best performance is in bold. The second-best performance is underlined.}
\label{table.1}
\end{table}

\subsubsection{Time Series Generation Evaluation} In time series generation, we compare CR-VAE with $3$ baselines: Time-series generative adversarial network (TimeGAN)~\cite{yoon2019time}
that takes transition dynamics into account under the framework of GAN; the popular variational RNN (VRNN) \cite{NIPS2015_b618c321}; and the variational recurrent autoencoder (VRAE)~\cite{kingma2013auto} which is the backbone of our approach. 

%For visualization, We implement t-SNE and PCA dimension reduction on both the original and synthetic samples after flattening to 2-dimensional vectors. The we observe how closely the distribution of generated samples resembles the original distribution in the 2-dimensional space. In the paper, we show the results of t-SNE, PCA plots can be found in the Supplementary Materials.

\begin{figure}[htb]

\vspace{-0.3cm}
\begin{minipage}[b]{0.24\linewidth}
  \centering
  \centerline{\includegraphics[width=2.4cm]{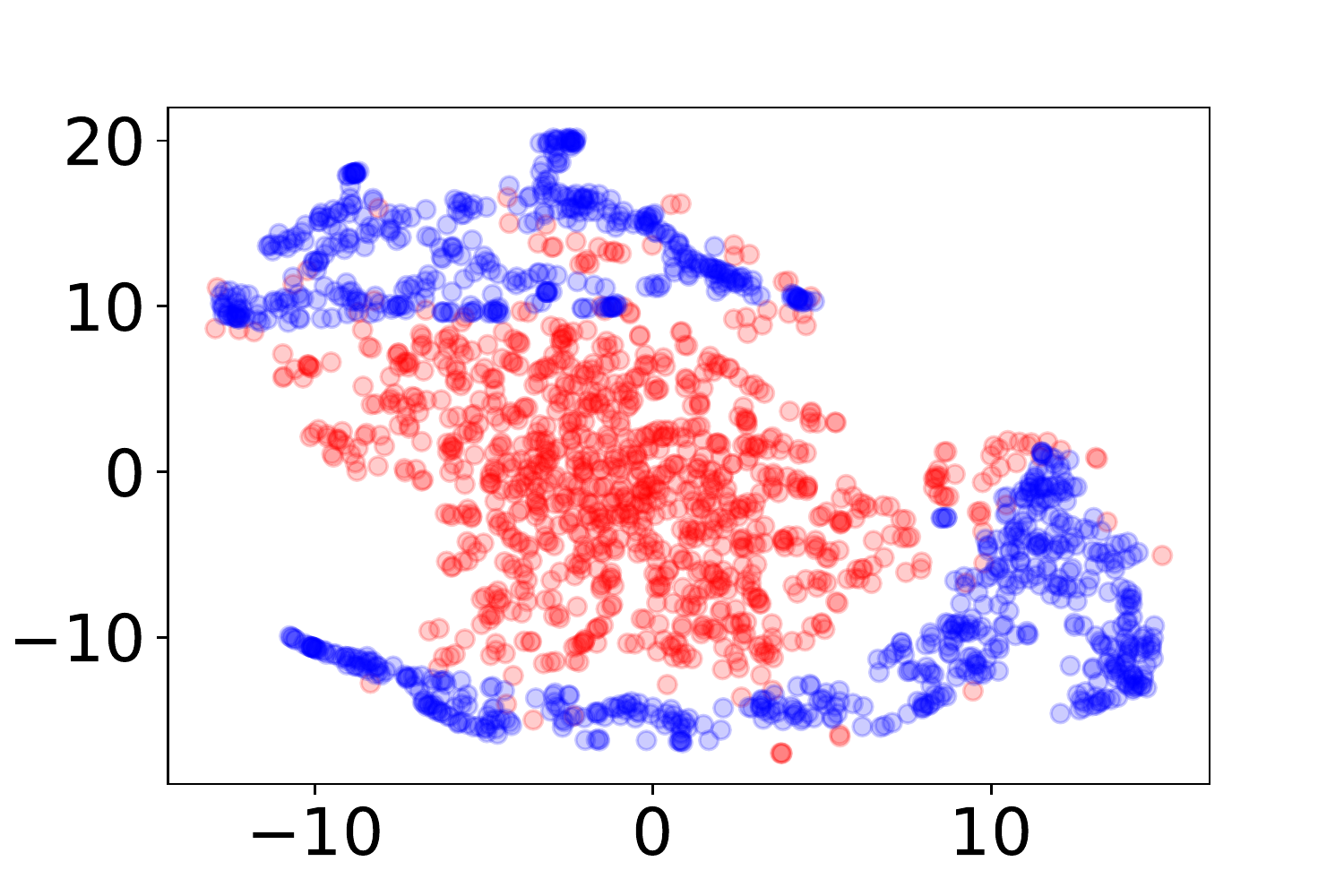}}
  %\hspace{-1.1cm}
  \centerline{}\medskip
   %\hspace{-1.1cm}
\end{minipage}
\hfill
\begin{minipage}[b]{.24\linewidth}
  \centering
  \centerline{\includegraphics[width=2.4cm]{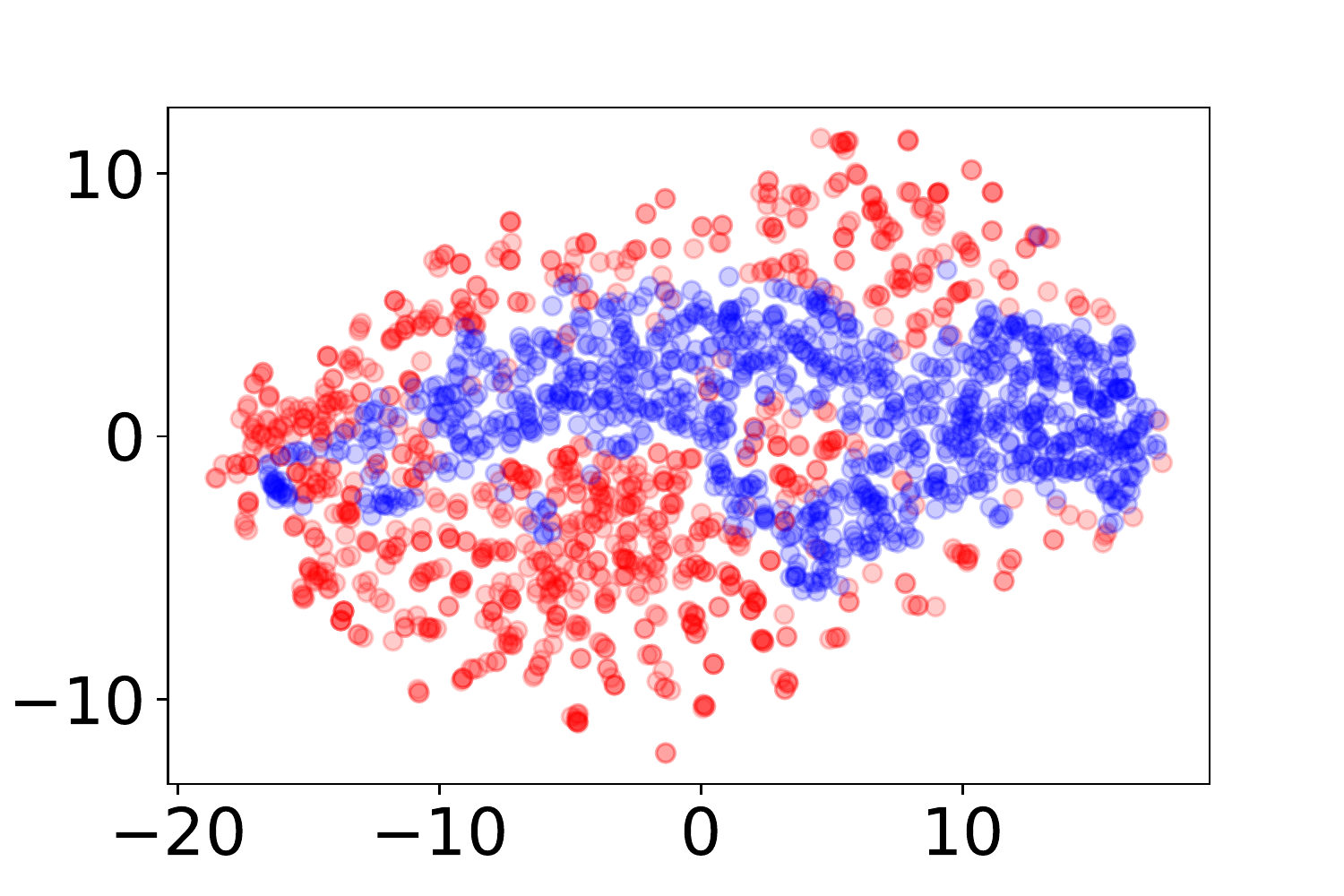}}
   %\hspace{-1.1cm}
  \centerline{}\medskip
\end{minipage}
\hfill
\begin{minipage}[b]{0.24\linewidth}
  \centering
  \centerline{\includegraphics[width=2.4cm]{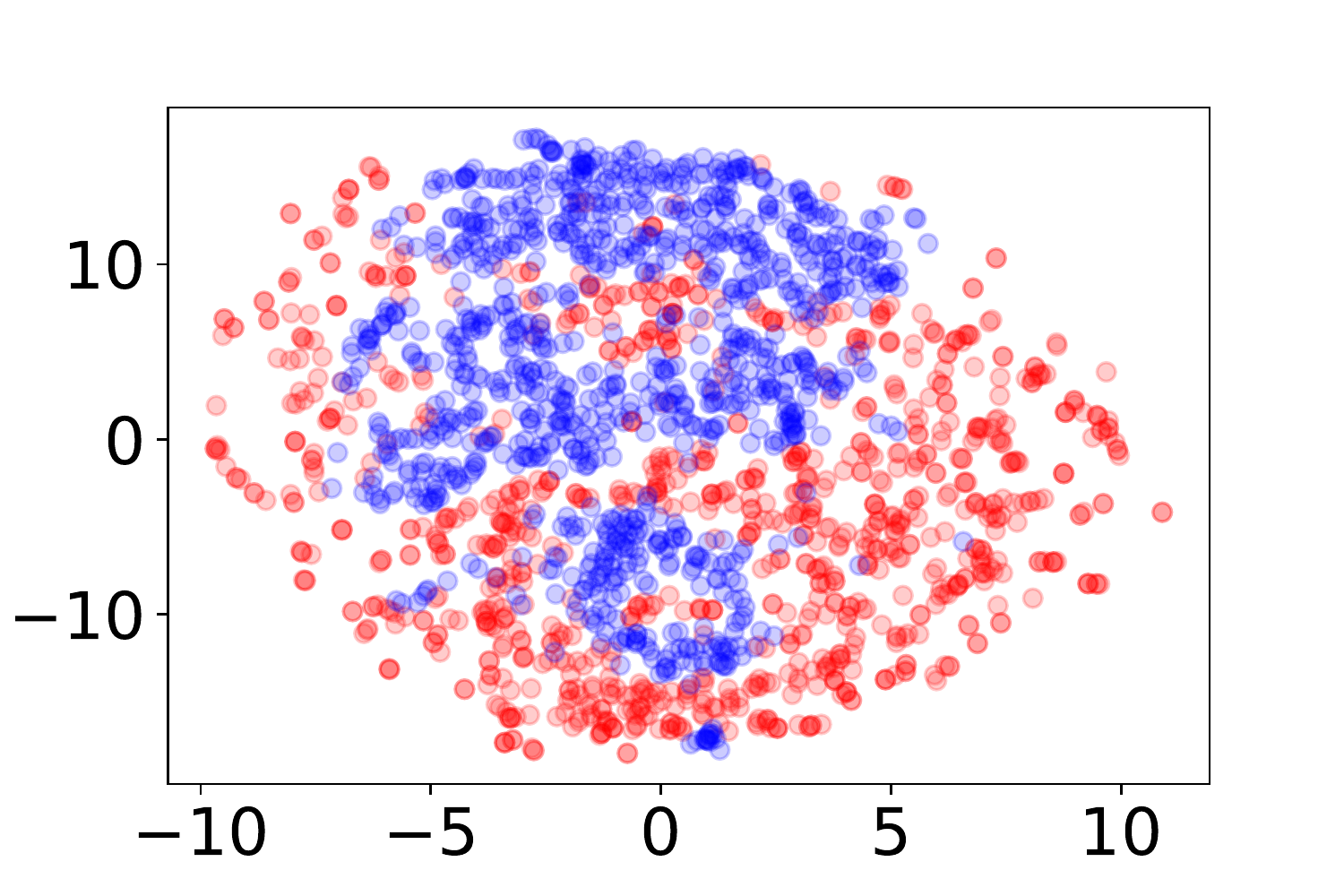}}
  %\hspace{-1.1cm}
  \centerline{}\medskip
\end{minipage}

\vspace{-0.5cm}

\begin{minipage}[b]{0.24\linewidth}
  \centering
  \centerline{\includegraphics[width=2.4cm]{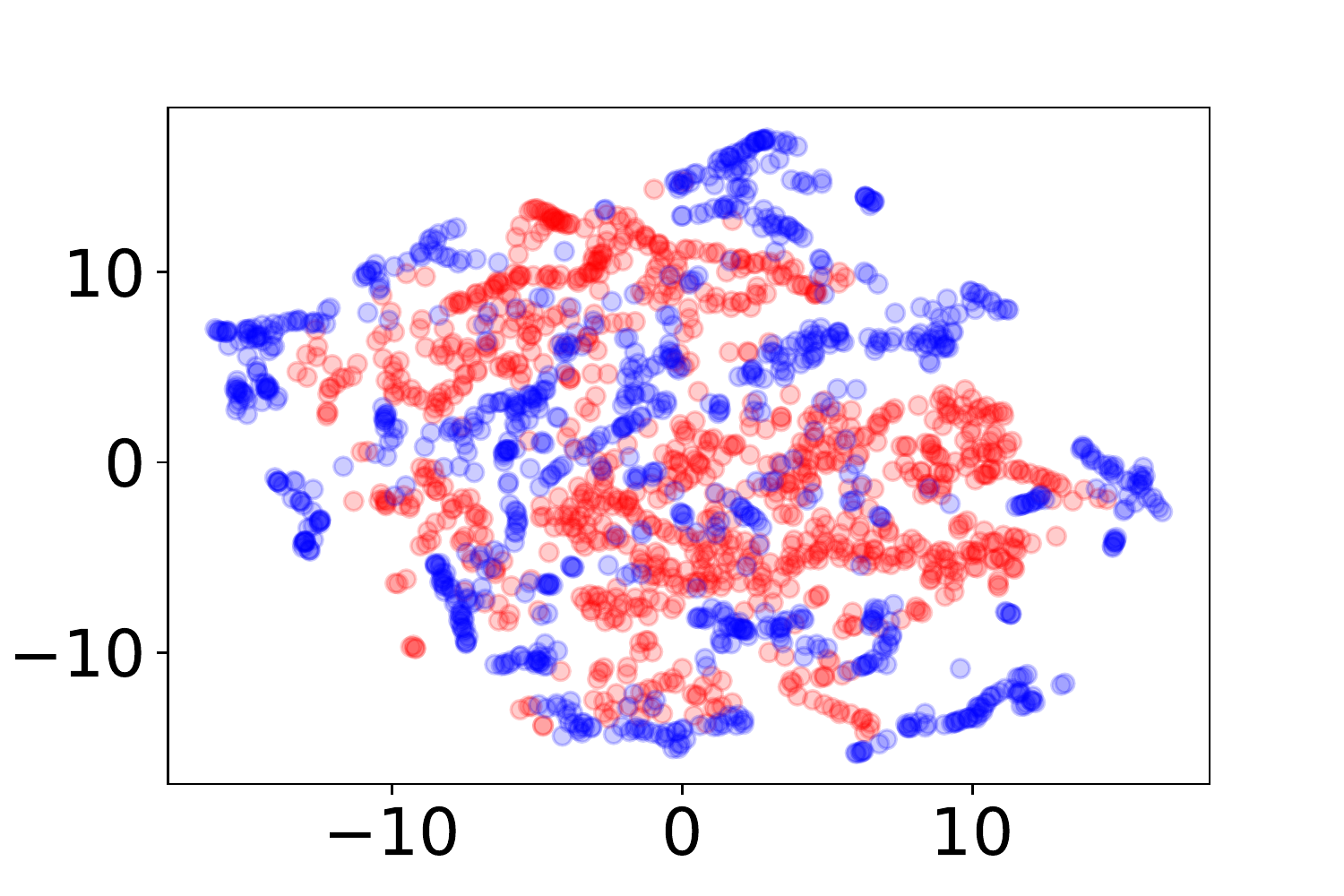}}
  % \hspace{0.cm}
  \centerline{}\medskip
\end{minipage}
\hfill
\begin{minipage}[b]{.24\linewidth}
  \centering
  \centerline{\includegraphics[width=2.4cm]{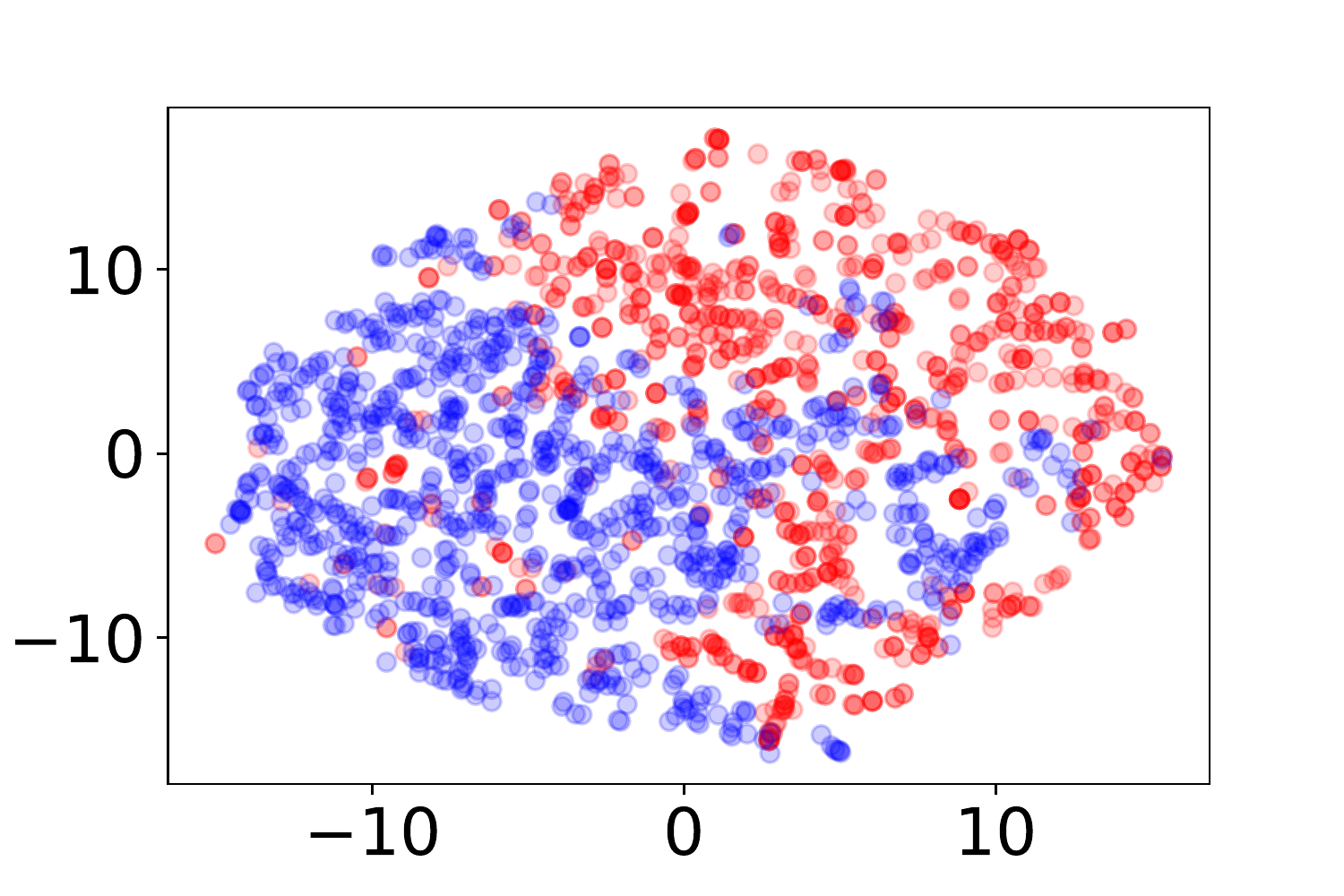}}
  % \hspace{0.cm}
  \centerline{}\medskip
\end{minipage}
\hfill
\begin{minipage}[b]{0.24\linewidth}
  \centering
  \centerline{\includegraphics[width=2.4cm]{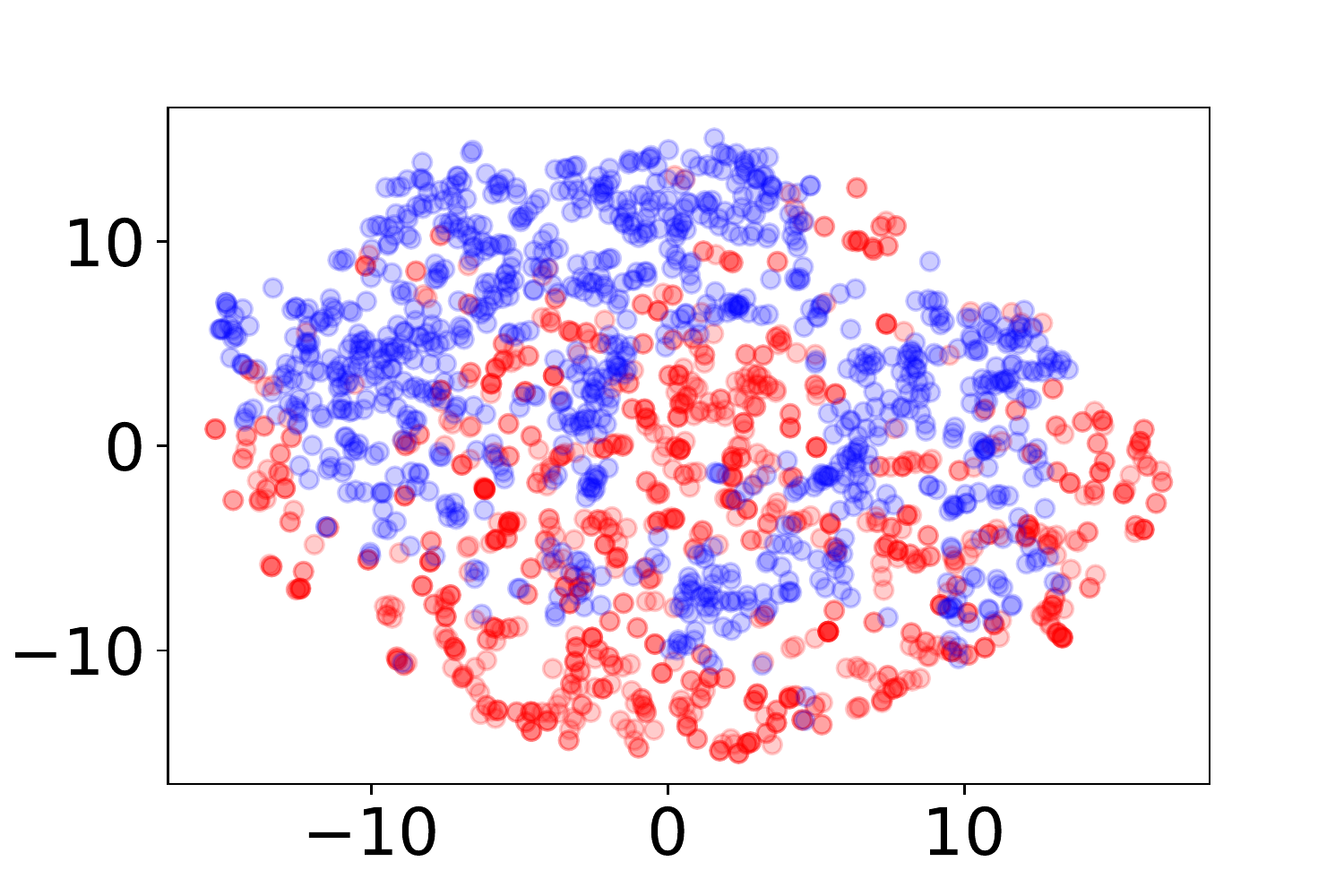}}
  % \hspace{0.cm}
  \centerline{}\medskip
\end{minipage}
\vspace{-0.5cm}

\begin{minipage}[b]{0.24\linewidth}
  \centering
  \centerline{\includegraphics[width=2.4cm]{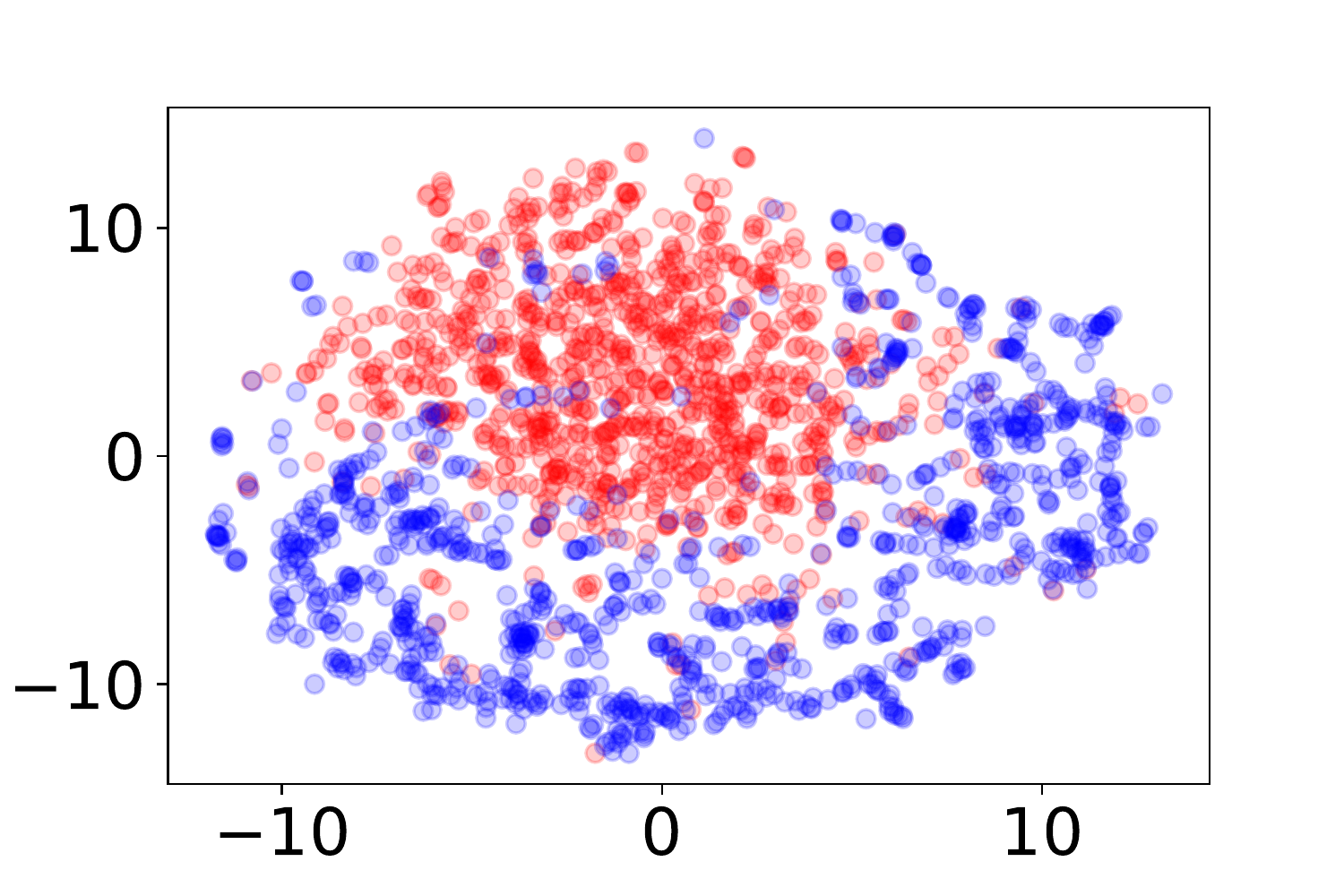}}
%  \vspace{1.5cm}
  \centerline{}\medskip
\end{minipage}
\hfill
\begin{minipage}[b]{.24\linewidth}
  \centering
  \centerline{\includegraphics[width=2.4cm]{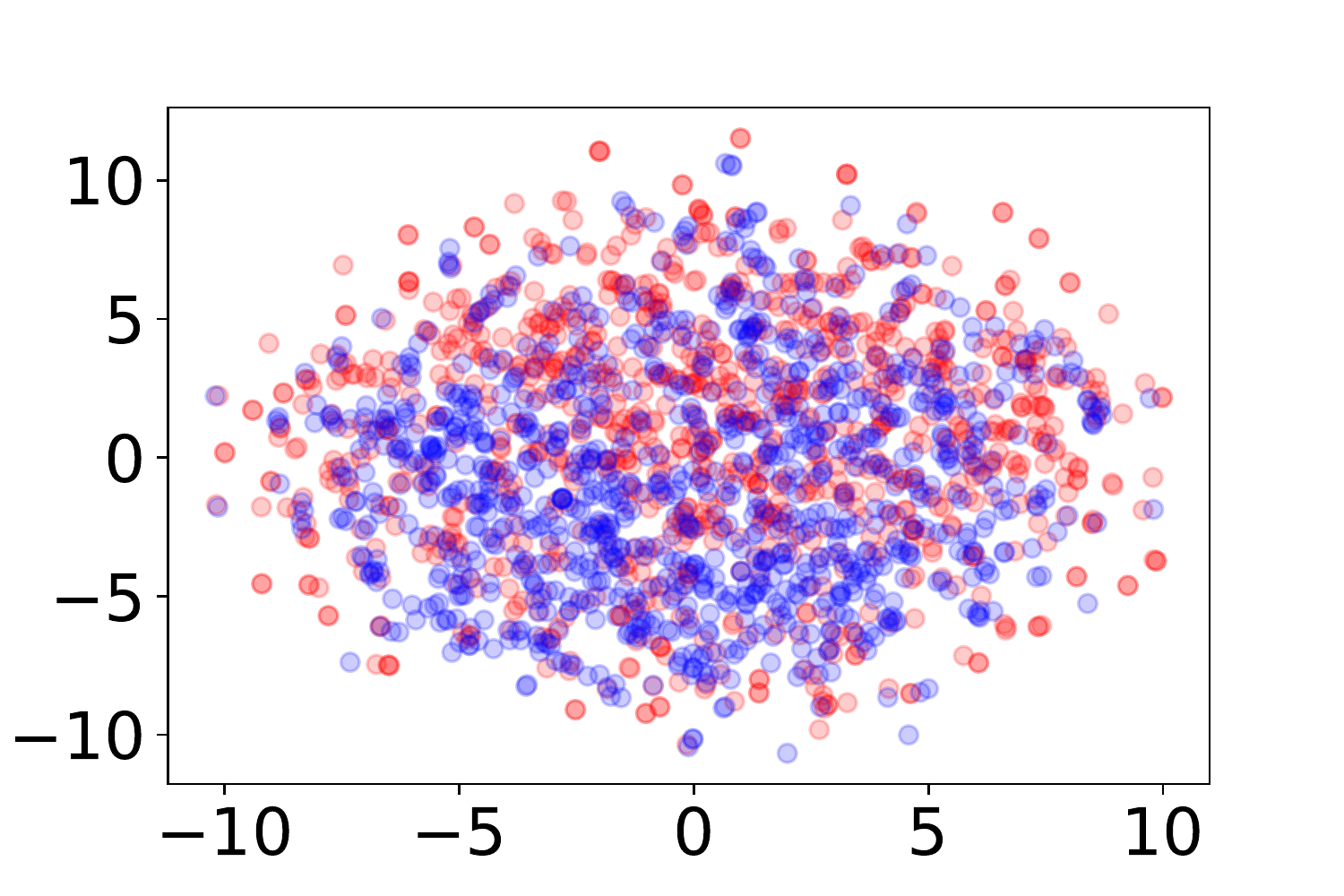}}
%  \vspace{1.5cm}
  \centerline{}\medskip
\end{minipage}
\hfill
\begin{minipage}[b]{0.24\linewidth}
  \centering
  \centerline{\includegraphics[width=2.4cm]{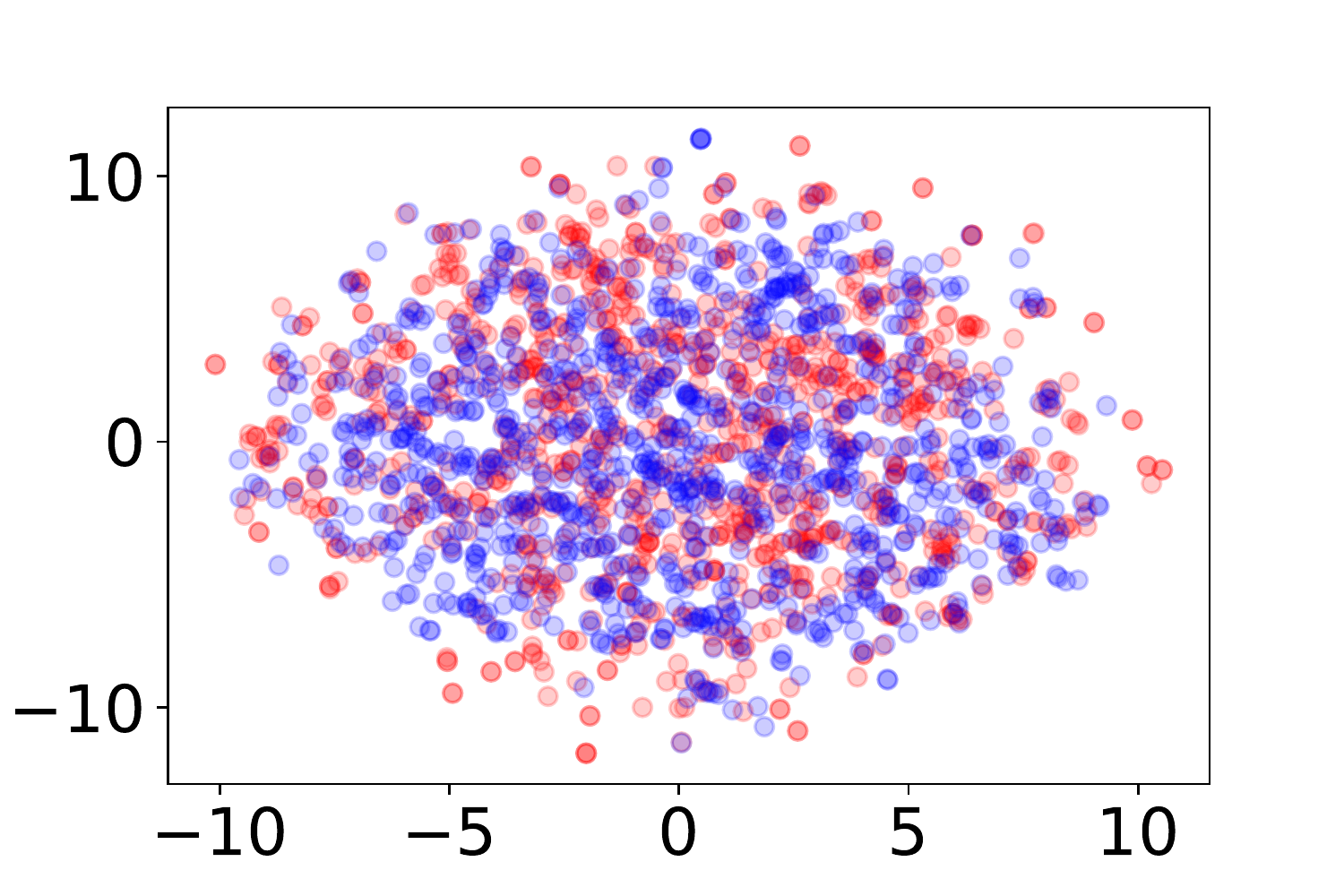}}
%  \vspace{1.5cm}
  \centerline{}\medskip
\end{minipage}

\vspace{-0.5cm}

\begin{minipage}[b]{0.24\linewidth}
  \centering
  \centerline{\includegraphics[width=2.4cm]{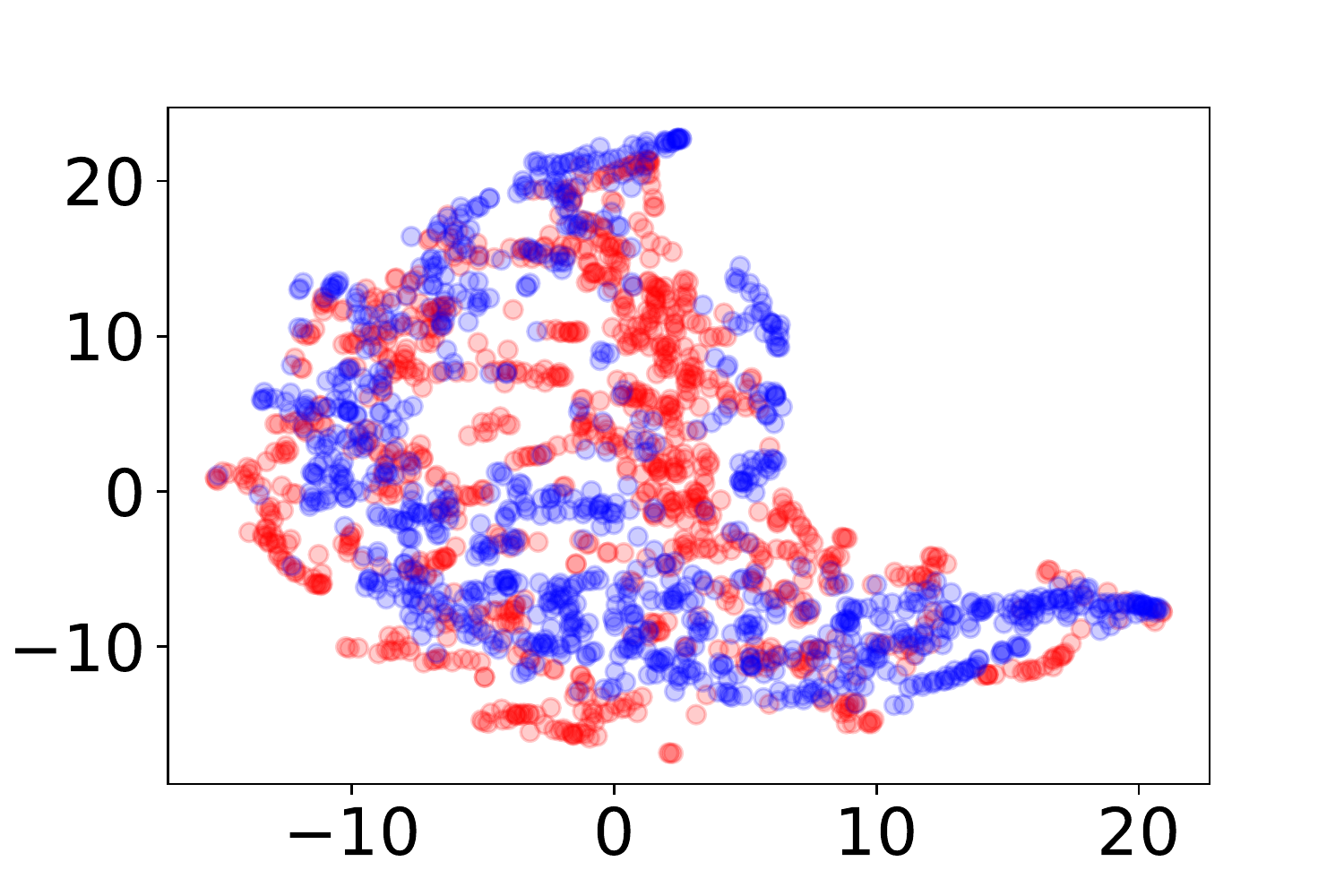}}
%  \vspace{1.5cm}
  \centerline{TimeGAN}\medskip
\end{minipage}
\hfill
\begin{minipage}[b]{.24\linewidth}
  \centering
  \centerline{\includegraphics[width=2.4cm]{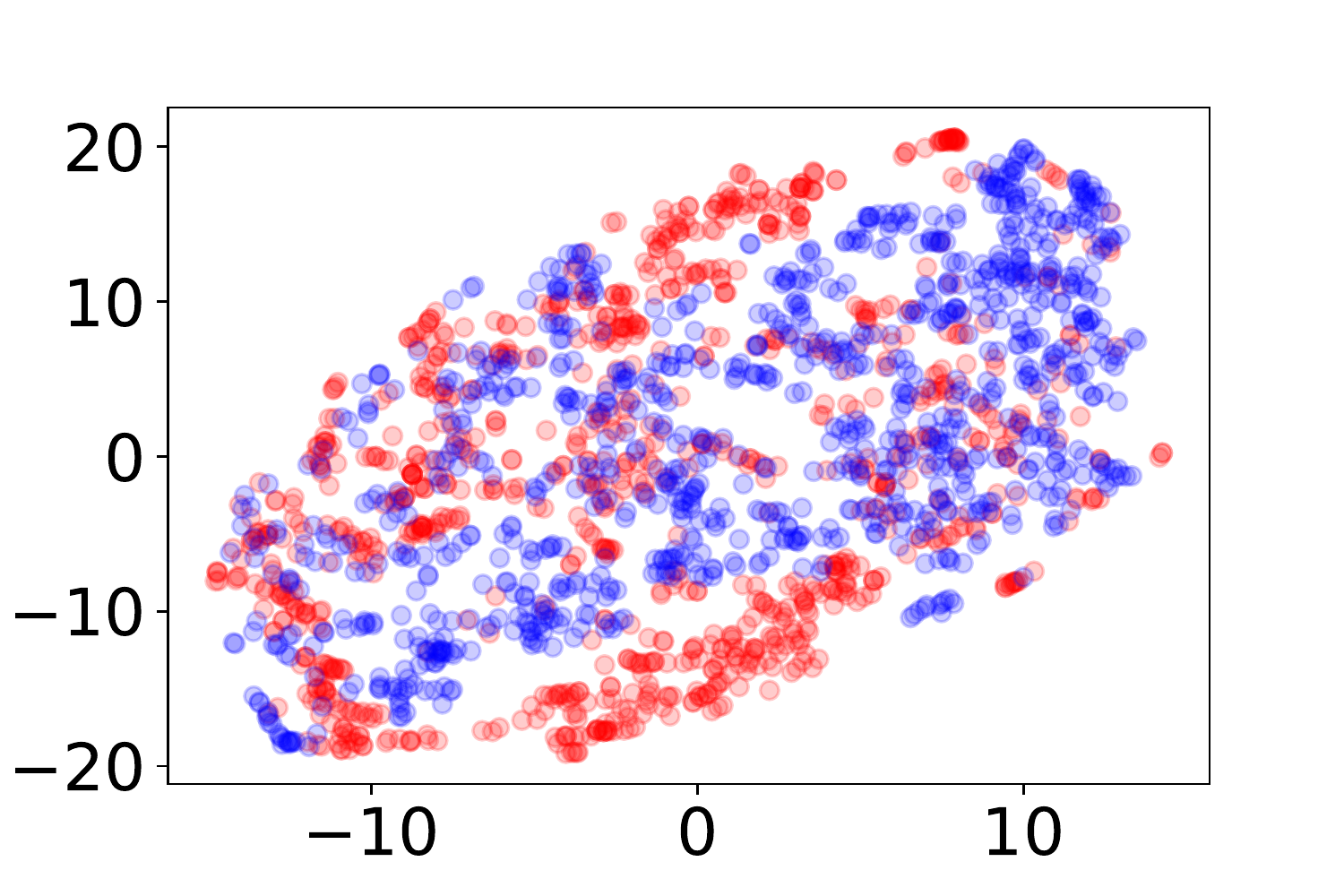}}
%  \vspace{1.5cm}
  \centerline{VRAE}\medskip
\end{minipage}
\hfill
\begin{minipage}[b]{0.24\linewidth}
  \centering
  \centerline{\includegraphics[width=2.4cm]{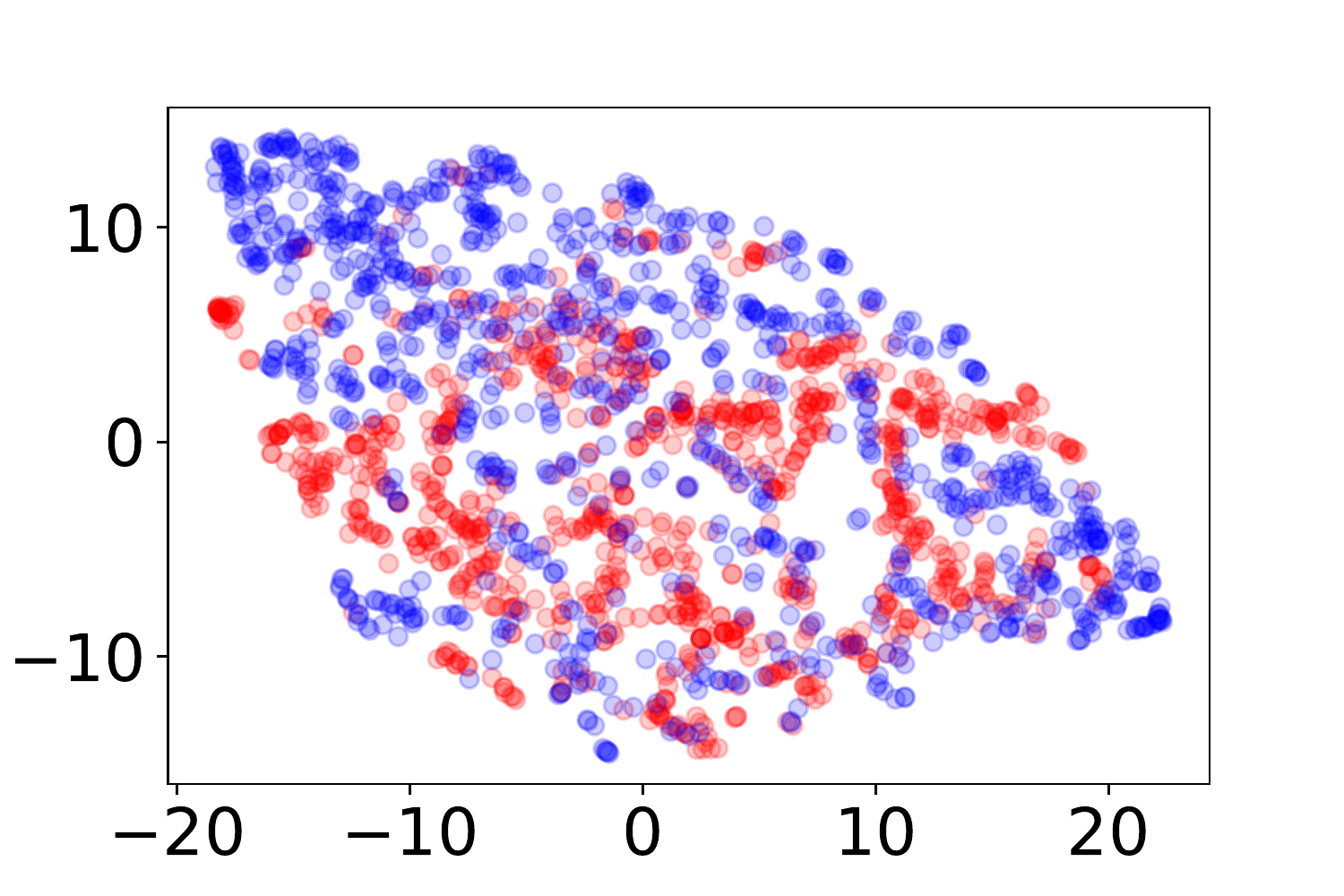}}
%  \vspace{1.5cm}
  \centerline{Ours}\medskip
\end{minipage}
\caption{t-SNE visualization on Henon (1st row), Lorenz 96  (2nd row), fMRI (3rd row) and EEG (4th row). Red samples correspond to real time series, whereas blue samples correspond to synthetic time series.}
\label{tsne}
\end{figure}

We first qualitatively evaluate the quality of generated time series by projecting both real and synthetic ones into a $2$-dimensional space with t-SNE. A good generative model is expected to encourage close distributions for real and synthetic data. As can be seen from Fig.~\ref{tsne}, CR-VAE demonstrates markedly better overlap with the original data than TimeGAN and performs slightly better than VRAE. On fMRI data, it is almost impossible to distinguish our generated samples with respect to real ones. This result further demonstrate the great potential of our CR-VAE in other medical applications.

% Each column provides the visualization for each method. 

%Fig. \ref{tsne} illustrates that synthetic samples generated by CR-VAE show markedly better overlap with the original data than TimeGAN and achieve competetive results compare to VRAE using t-SNE. On fMRI dataset, the generated samples of our method and real samples are almost perfectly in sync.

Next, we adopt the maximum mean discrepancy (MMD) \cite{gretton2006kernel} and the ``train on synthetic and test on real" (TSTR) strategy to further evaluate the performances of different methods quantitatively. Specifically, MMD is utilized to measure the distance between generated data and real data. Same to \cite{goudet2018learning}, we take account of a bandwidth of kernel size $[0.01, 0.1, 1, 10, 100]$. For TSTR, we use synthetic samples to train a sequential prediction neural network with LSTM-RNN layers to predict next samples. Then we test the trained model on real time series. Prediction performance is measured by root mean square error (RMSE). Intuitively, if a generative model captures well the underlying dynamics of a real time series (i.e., $p(\mathbf{x}_t |\mathbf{x}_{1:t-1})$), it is expected to have low prediction error under TSTR framework. 

As shown in Table~\ref{table.2}, CR-VAE consistently generates higher-quality synthetic data in comparison to baselines. For fMRI, our result is slightly outperformed by VRAE. This is because CR-VAE fails to discover some causal relations. 
%which leads to this causal variable is not used as input for prediction.

% \begin{table}[]
% \begin{tabular}{llll|l}
% Methods & TimeGAN & VRAE            & Ours            & Real (TRTR) \\ \hline
% Henon   &0.0884         & 0.0157          & \textbf{0.0149} & 0.0006      \\
% Lorenz  & 0.0155  & 0.0077          & \textbf{0.0032} & 0.0003      \\
% fMRI    & 0.0400  & \textbf{0.0142} & 0.0159          & 0.0115 \\ 
% EEG    &0.0018   & 0.0009 & \textbf{0.0006}          & 0.0001    
% \end{tabular}
% \caption{Performance comparison of generation using TSTR on the Henon, Lorenz 96, fMRI and EEG. The best performance is in bold.}
% \label{table.2}
% \end{table}

\begin{table}[]
\begin{tabular}{c|c|c|c|c|c}
Metric                        & Methods      & Henon          & Lorenz         & fMRI           & EEG            \\ \hline
                           & TimeGAN      & 0.476          & 0.040          & 0.157          & 0.064          \\
MMD                           & VRNN         & 0.324           & 0.043  & 0.145                         & \underline{0.072}           \\
                        & VRAE         & \underline{0.125}          & \textbf{0.010} & \underline{0.011}          & 0.107          \\
                              & Ours & \textbf{0.118} & \underline{0.015}          & \textbf{0.010} & \textbf{0.050} \\ \hline
                              & TimeGAN      & 0.297          & 0.124          & 0.163          & 0.042          \\
TSTR                          & VRNN         & 0.185           & 0.131         & 0.233   & 0.054          \\
                            & VRAE         & \underline{0.125}          & \underline{0.088}          & \textbf{0.119} & \underline{0.030}          \\
(RMSE)                              & Ours & \textbf{0.122} & \textbf{0.056} & \underline{0.122}          & \textbf{0.024} \\ \cline{2-6} 
                              & Real (TRTR)  & 0.024          & 0.017          & 0.107          & 0.010         
\end{tabular}
\caption{Quantitative comparison with MMD and TSTR. The best performance is in bold. The second-best performance is underlined.}
\label{table.2}
\end{table}
\section{Conclusion}
We develop a unified model, termed \emph{causal recurrent variational autoencoder} (CR-VAE), that integrates the concepts of Granger causality into a recurrent VAE framework. CR-VAE is able to discover Granger causality from past observations to present values between pairwise variables and within a single variable. Such functionality makes the generation process of CR-VAE more transparent. 
We test CR-VAE in two synthetic dynamic systems and two benchmark medical datasets. Our CR-VAE always has smaller maximum mean discrepancy values and prediction mean square errors using the ``train on synthetic and test on real" strategy.  

Future works are twofold. First, same to other Granger causality approaches, our model assumes no unmeasured confounders. Second, an isotropic Gaussian assumption for latent factors limits our generative capability. We will continue the design of time series generative models to account for latent confounders and more flexible latent distributions. 

\onecolumn
\pagebreak

\section{Acknowledgments}
This work was funded in part by the U.S. ONR under grant ONR N00014-21-1-2295, and in part by the Research Council of Norway (RCN) under grant 309439.

\section{Supplementary Material}
This document contains the supplementary material for the \textit{``Causal Recurrent Variational Autoencoder for Medical Time Series Generation"} manuscript. It is organized into the following topics and sections:

\begin{enumerate}
%\item The Dependence between $\mathbf{t}$ and $y$
\item More on Related Works
\item Details of Experimental Setting
\begin{enumerate}
\item More Details of Dataset
\item Details of Traditional Granger Causality
\item Details of Neural Network-based Approaches of Granger Causality
\item Details of Generation
\end{enumerate}
\item Additional Experimental Results
\begin{enumerate}
\item Adjacency Matrices of Granger Causal Summary Graphs
\item Synthetic Time Series by Different Methods
\item Prediction Results of TSTR
\item PCA Visualization
\item Ablation Study of the Two-Stage Training Strategy
\end{enumerate}

\item Minimal Structure of CR-VAE in PyTorch and Code Link

\end{enumerate}

\section{More on Related Works}

\begin{figure}[htbp]
\centerline{\includegraphics[scale=.4]{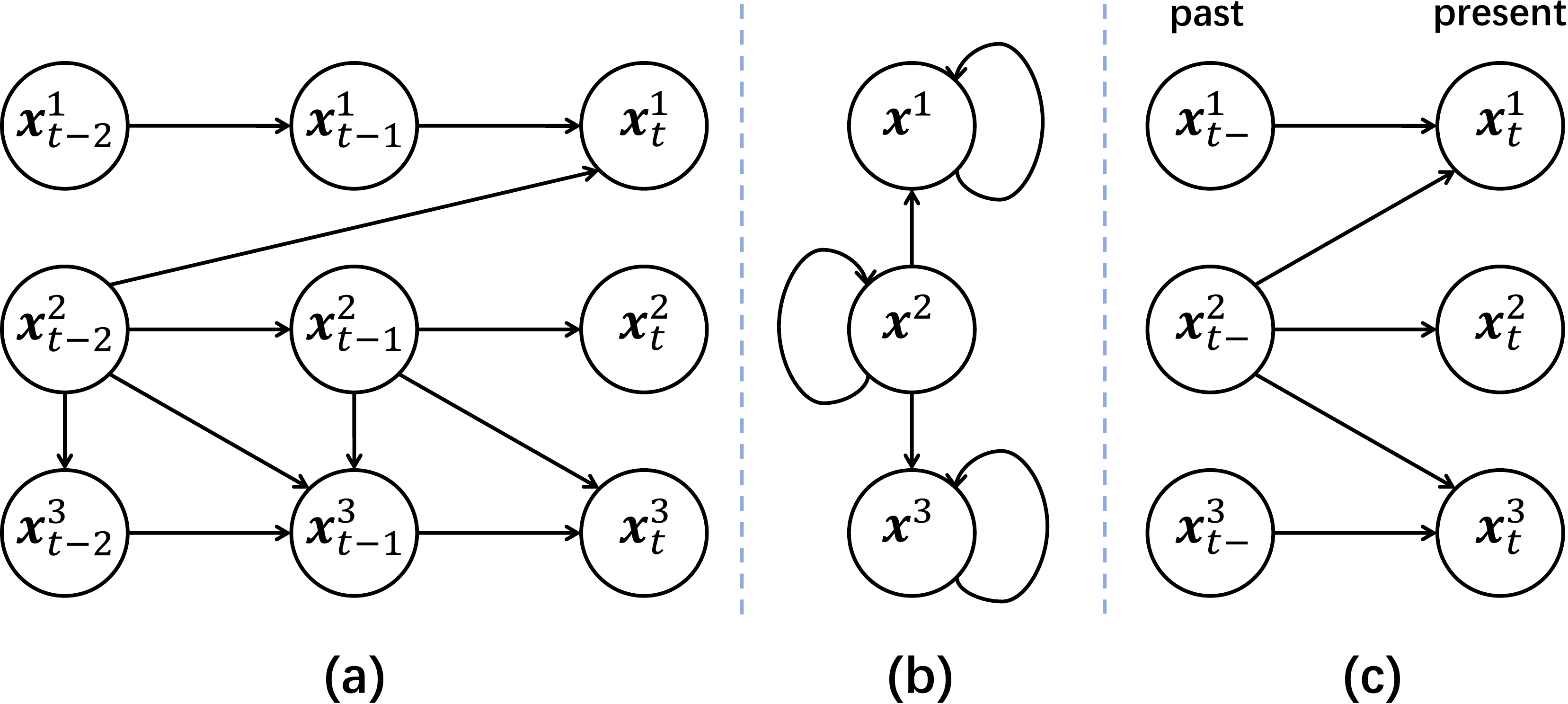}}
\caption{Example of (a) window causal graph; (b) summary
causal graph; and (c) our Granger causal graph. Figure adapted from~\cite{assaad2022causal}.}
\label{fig:causal_graphs}
\end{figure}

Different types of causal graphs can be considered for time series~\cite{assaad2022survey}. The window causal graph (see Fig.~\ref{fig:causal_graphs}(a)) only covers a fixed number of time instants (with a maximum causal influence lag $\tau$) and assumes the causal relations amongst different variables are consistent over time. The summary causal graph (see Fig.~\ref{fig:causal_graphs}(b)) directly relates variables without any indication of time. Usually, it is difficult to estimate window causal graph because it requires to determine which exact time instant is the cause of another. It is of course easier to estimate a summary causal graph~\cite{assaad2022causal}. In practice, it is often sufficient to know the causal relations between time series as a whole, without knowing precisely the relations between time instants~\cite{assaad2022survey}.

In our work, we consider recovery of a Granger causal graph (see Fig.~\ref{fig:causal_graphs}(c)), which separates past observations and present values of each variable and aims to if the past of $\mathbf{x}^p$ (denoted $\mathbf{x}_{t-}^p$) causes the present value of $\mathbf{x}^q$ (denoted $\mathbf{x}_t^q$). Obviously, our Granger causal graph lies between window causal graph and summary causal graph.

Substantial efforts have been made on the causal discovery of time series. The most two popular approaches in this topic are the \underline{Granger causality-based approach} and the \underline{constraint-based approach}. The former typically assumes that the past of a cause is necessary and sufficient for optimally forecasting its effect, whereas the latter firstly exploits conditional independencies to build a skeleton between each dimension and then orients the skeleton according to a set of rules that define constraints on admissible orientations (e.g., the PC algorithm with momentary conditional independence (PCMCI)~\cite{runge2019detecting}). Other well-known approaches include the \underline{noise-based approach} (e.g., the Time Series Models with Independent Noise (TiMINo)~\cite{peters2013causal}) and the \underline{score-based approach} (e.g., the DYNOTEARS~\cite{pamfil2020dynotears}). Our model belongs to the Granger causality-based approach.

\section{Details of Experiment Setting}

\subsection{More Details of Dataset}

The equations of our H\'enon chaotic dataset can be written as:
\begin{equation}\label{eq.henon}
\centering
\begin{array}{lr}
\mathbf{x}^{1}_{t+1} = 1.4 - (\mathbf{x}^{1}_{t})^{2} + 0.3\mathbf{x}^{1}_{t-1},
\\
\mathbf{x}^{p}_{t+1} = 1.4 - (e\mathbf{x}^{p-1}_{t}+(1-e)\mathbf{x}^{p}_{t})^2 + 0.3\mathbf{x}^{p}_{t-1},
\end{array}
\end{equation}
where $p = 2,.....K$. The true direct causal relations are $\mathbf{x}^{p-1} \rightarrow \mathbf{x}^{p}$ and self-causal relations $\mathbf{x}^{p} \rightarrow \mathbf{x}^{p}$, i.e., the index of positive elements should be $(p, p-1)$. The length of lag is $2$ while $e = 0.3$, $K =6$. 

Table~\ref{table.d} summarizes the crucial statistics of four datasets. We conduct our experiments using more than $2,000$ samples on each dataset. 
For Henon and Lorenz $96$, we sample initial values from standard Gaussian distribution, and then infer the trajectories using transition functions. The fMRI is different from other datasets since it is not a long sequence but $50$ subjects of time sequences. Therefore, we crop them in shorter clips according to models' lag and randomly select $2,048$ clips to train models. The true lag is not available on fMRI. The EEG data consists of time sequences from $76$ nodes. The first $64$ nodes record the EEG signals through a $8 \times 8$ grid at the cortical level (CE), while
the other $12$ are placed at deeper brain structures (DE). Therefore, we select the last $12$ contacts as our dataset. We also reassign the values of each node to $[0,1]$ for normalization. Ground truth of causal relations is not available on EEG data. 

\begin{table*}[h]
\centering
\begin{tabular}{c|c|c|c|c}
Dataset & Number of Samples & Number of Variables & True Lag & Number of Causal Relations \\ \hline
H\'enon   & 2048              & 6                   & 2        & 11                         \\
Lorenz  & 2048              & 10                  & 3        & 40                         \\
fMRI    & 2048              & 10                  & NA       & 21                         \\
EEG     & 5000              & 12                  & NA       & NA                        
\end{tabular}
\caption{Statistics of Datasets. NA notes ground truth is not available.}
\label{table.d}
\end{table*}

\subsection{Details of Traditional Granger Causality}
For kernel Granger causality (KGC), we use the official MATLAB code from authors \footnote{\url{https://github.com/danielemarinazzo/KernelGrangerCausality}}.
For transfer entropy (TE), we follow~\cite{de2019data} and estimate information theoretic metrics using the matrix-based R{\'e}nyi's $\alpha$-order entropy functional~\cite{Giraldo2014}. Suppose there are $N$ samples for the $p$-th variable, i.e., $\mathbf{x}^{p} = [\mathbf{x}^{p}(1), \mathbf{x}^{p}(2), \cdots,\mathbf{x}^{p}(N)]$ where the numbers in parentheses denotes sample index. A Gram matrix $K^p \in \mathbb{R}^{N \times N}$ can be obtained by computing $K^p(n,m) = \kappa(\mathbf{x}^{p}(n), \mathbf{x}^{p}(m))$, where $\kappa$ is a Gaussian kernel with kernel width $\sigma$. The entropy of $\mathbf{x}^{p}$ can be expressed by \cite{Giraldo2014}:
\begin{equation}\label{eq:renyi_entropy}
H_{\alpha}(\mathbf{x}^p) = H_{\alpha }(A^p) = \frac{1}{1-\alpha }\log_{2}[\sum_{n=1}^{N}\lambda _{n}(A^p)^{\alpha}],
\end{equation}
where $A^p=K^p/trace(K^p)$ is the normalized Gram matrix and $\lambda _{n}(A_i)$ denotes $n$-th eigenvalue of $A^p$.

Further, the joint entropy for $\{\mathbf{x}^p\}_{p=1}^M$ is defined as \cite{Yu2019}:

\begin{equation}\label{eq:renyi_joint_entropy}
H_{\alpha}(\mathbf{x}^1,\mathbf{x}^2,\cdots,\mathbf{x}^M) = H_{\alpha }(\frac{A_1\circ A_2\cdots \circ A_M}{trace(A_1\circ A_2\cdots \circ A_M)}).
\end{equation} 
where $\circ$ denotes element-wise product. Give a variable $\mathbf{y}$, We also have the: 

\begin{equation}
\begin{aligned}
    & {\rm TE}_{\alpha}(\mathbf{x} \rightarrow \mathbf{y}) = H_{\alpha}(\mathbf{y}|\mathbf{y}-) - H_{\alpha}(\mathbf{y}|\mathbf{x}-,\mathbf{y}-),\\
    & H_{\alpha}(A|B) = H_{\alpha}(A,B)-H_{\alpha}(B),
\end{aligned}
\label{eq.TE}
\end{equation}

Therefore, we can estimated TE from data directly using Eqs.~(\ref{eq:renyi_entropy})-(\ref{eq.TE}).

%  We select Gaussian kernel for both KGC and TE. 

In our experiments, for traditional Granger causal discovery methods (i.e., KGC and TE), the hyperparameters such as the kernel width $\sigma$ are determined by grid search. Results reported in the main manuscript use hyperparameters that achieved the highest AUROC. Details are summarized in Table~\ref{table.1}.

\begin{table*}[h]
\centering
\begin{tabular}{l|l|l|l|l|l|l|l|l|l|l|l}
Dataset                & Method & Lag & Kernel Size & Dataset                 & Method & Lag & Kernel Size & Dataset               & Method & Lag & Kernel Size \\ \hline
Henon & KGC    & 2   & 0.1         & Lorenz & KGC    & 5   & 0.5         & fMRI & KGC    & 5   & 0.2         \\
                       & TE     & 2   & 0.1         &                         & TE     & 5   & 0.1         &                       & TE     & 5   & 0.5        
\end{tabular}
\caption{Hyper-parameters of traditional Granger causal discovery methods}
\label{table.1}
\end{table*}

\subsection{Details of Neural Network-based Approaches of Granger Causality}
For neural network-based methods, we assume a range of sparsity level and then search a grid of hyperparameters that minimize the convex part of their loss functions. With this testing setup, our goal is to fairly compare the performance of neural network-based techniques that rely on sparisty trick. For example, we assume the sparse level of causal relations on H\'enon chaotic data ranges from $20\%$ to $40\%$, and then we tune the $\lambda$ of TCDF, NGC and our CR-VAE to converge to this sparsity level. In the end, we tune other hyper-parameters including early stopping by searching lowest loss without the sparsity penalty, i.e., convex part of loss functions. We fix the lag to be $10$ and batch size to be $256$. Other hyper-parameters are shown in Table~\ref{table.2}.

\begin{table*}[h]
\centering
\begin{tabular}{c|c|c|c|c|c}
Dataset                 & Method & Sparsity Level Range & Sparsity Coefficient $\lambda$ & Hidden Neurons & Hidden Layers \\ \hline
                        & TCDF   & {[}20\% , 40\% {]}   & 1                                            & 64             & 2             \\
Henon                   & NGC    & {[}20\% , 40\% {]}   & 0.5                                          & 64             & 2             \\
                        & Ours   & {[}20\% , 40\% {]}   & 0.2                                          & 64             & 2             \\ \hline
                        & TCDF   & {[}35\% , 50\% {]}   & 2.5                                          & 64             & 2             \\
Lorenz                  & NGC    & {[}35\% , 50\% {]}   & 10                                           & 64             & 2             \\
                        & Ours   & {[}35\% , 50\% {]}   & 8                                            & 64             & 2             \\ \hline
                        & TCDF   & {[}15\% , 28\% {]}   & 1                                            & 128            & 2             \\
fMRI                    & NGC    & {[}15\% , 28\% {]}   & 0.5                                          & 64             & 2             \\
                        & Ours   & {[}15\% , 28\% {]}   & 0.75                                         & 64             & 2            
\end{tabular}
\caption{Hyper-parameters of neural-network-based Granger causal discovery methods}
\label{table.2}
\end{table*}

\subsection{Details of Generation}
We have two baselines that represent two major tracks of generative models. TimeGAN is a recently proposed state-of-the-art (SOTA) GAN-based generative model for time series that takes transition dynamics $p(\mathbf{x}_t|\mathbf{x}_{1:t-1})$ into account and also outperforms other GAN-based generators, such as~\cite{esteban2017real}. We use the official codes in tensorflow\footnote{\url{https://github.com/jsyoon0823/TimeGAN}}, and search a range of hyper-parameters starting from default setting of the codes to minimize the MMD between real and generated data.
VRAE is a classic extension of VAE for time series data. We replicate the official codes of theano using PyTorch\footnote{\url{https://github.com/y0ast/Variational-Recurrent-Autoencoder}}. For both VRAE and our CR-VAE, we searched across
a grid of hyper-parameters to minimize MMD.

\subsubsection{Evaluation Metrics of Generation}

The first metric we apply is the maximum mean discrepancy (MMD)~\cite{gretton2012kernel} which has been widely used to measure the distance between two distributions. In other words, it quantifies if two sets of samples - one synthetic, and the other
for the real data - are generated from the same distribution. More formally, given $n$ samples of real and synthetic, we have:
\begin{equation}\label{eq:mmd}
MMD(\mathbf{x}, \hat{\mathbf{x}}) = \frac{1}{n^2}\kappa(\mathbf{x}(i), \mathbf{x}(j)) + \frac{1}{n^2}\kappa(\hat{\mathbf{x}}(i), \hat{\mathbf{x}}(j)) - \frac{2}{n^2}\kappa(\mathbf{x}(i), \hat{\mathbf{x}}(j)),
\end{equation}
where $\hat{\mathbf{x}}$ and $\mathbf{x}$ denote synthetic and real data, respectively, indexed by $i,j$; kernel $\kappa$ is taken as the Gaussian kernel. We average the results of a bandwidth of kernel size $[0.01, 0.1, 1, 10, 100]$, same to \cite{goudet2018learning}. The MMD is more informative than either generator or discriminator loss in GAN - based models \cite{esteban2017real}, so we early stop to obtain lowest MMD for generative models.

The second metric we apply is ``train on synthetic and test on real" (TSTR) by prediction. In our TSTR experiment, we adopt a two-layer RNN with GRU gates to predict next values using synthetic data. Each layer has 64 neurons. The order (length) of the RNN is fixed to be 10. We minimize the mean square error (MSE) using Adam, and fix the learning rate to be 1e-4. We train the RNN with a batch size of 128.
Besides, we also split 10\% synthetic data as our validation set for early stopping via cross-validation. Then we test the trained model on real time series and quantify the prediction performance using RMSE. More formally:
\begin{equation}\label{eq:mmd}
RMSE(\mathbf{x}, \hat{\mathbf{x}}) = \frac{1}{M}\sum^M_{p = 1}\sqrt{\frac{\sum_t(\mathbf{x}_t - \hat{\mathbf{x}}_t)^2}{T}},
\end{equation}
where $\hat{\mathbf{x}}$ and $\mathbf{x}$ denote predicted and true samples on real datasets; $T$ is the length of the time clips; $M$ is the number of dimensions indexed by $p$.

\begin{figure}[h]

\vspace{-0.3cm}
\begin{minipage}[b]{1\linewidth}
  \centering
  \centerline{\includegraphics[scale=0.85]{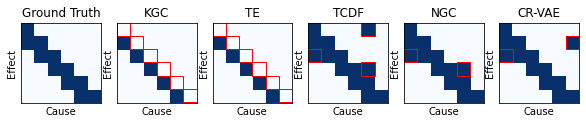}}
  %\hspace{-1.1cm}
  \centerline{}\medskip
   %\hspace{-1.1cm}
\end{minipage}

\begin{minipage}[b]{1\linewidth}
  \centering
  \centerline{\includegraphics[scale=0.85]{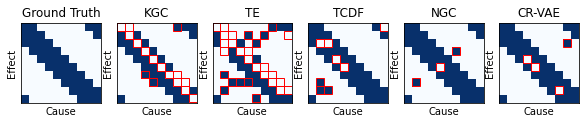}}
   %\hspace{-1.1cm}
  \centerline{}\medskip
\end{minipage}

\begin{minipage}[b]{1\linewidth}
  \centering
  \centerline{\includegraphics[scale=0.85]{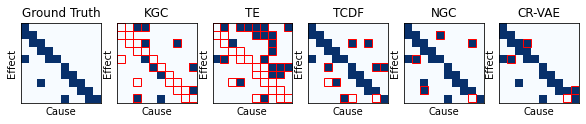}}
  %\hspace{-1.1cm}
  \centerline{}\medskip
\end{minipage}

\caption{Adjacency matrices on H\'enon (1st row), Lorenz 96  (2nd row) and fMRI (3rd row). The $1$-st column provides the visualization for ground truth, while other columns provide matrices estimated by causal discovery techniques. False causal relations are highlighted by red rectangles.}
\label{compare}
\end{figure}

\section{Additional Experimental Results}
\subsection{Adjacency Matrices of Granger Causal
Summary Graphs}

We provide the ground truth of adjacency
matrices of Granger causal graphs and compare them with counterparts estimated by CR-VAE and other competing methods. For traditional Granger causality techniques, we search the thresholds by maximizing AUROC and get rid of all values under the thresholds. For neural-based algorithms, we can directly show the estimated matrices. 

False positive or negative elements are highlighted by red squares. As illustrated in Fig.~\ref{compare}, CR-VAE outperforms most of baselines and achieves competitive results among neural network-based methods. 

%It also shows a big drawback for traditional method that cannot detect self-causal relations.

\subsection{Synthetic Time Series by Different Methods}
In this section, we generate synthetic time series of length $20$ using all methods on different datasets. For TimeGAN, we sample a noise from $\mathbf{z} \sim N(0,I)$
and feed it to the generator which outputs sequences of synthetic samples. For VRAE, we sample a noise $\mathbf{z} \sim N(0,I)$ and utilize it as the initial state of decoder RNN. Meanwhile, the first input of decoder RNN is zero. We use the predicted value as the input of the next time step to obtain synthetic sequences step-wisely. Similar to VRAE, our CR-VAE also infers the synthetic sequences step-wisely with an initial state $\mathbf{z} \sim N(0,I)$ and zero input. The difference is that we sample another noise $\mathbf{z}_{\varepsilon}\sim N(0,I)$ for error-compensation network to generate a sequence of innovation terms. They are added step-wisely to the inferred values of decoder RNNs in CR-VAE. 

\begin{figure}[h]

\vspace{-0.3cm}
\begin{minipage}[b]{0.24\linewidth}
  \centering
  \centerline{\includegraphics[width=4cm]{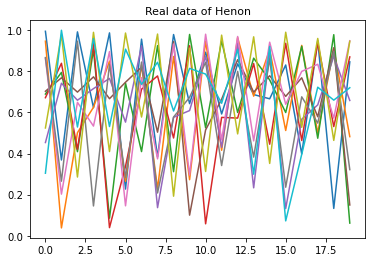}}
  %\hspace{-1.1cm}
  \centerline{}\medskip
   %\hspace{-1.1cm}
\end{minipage}
\hfill
\begin{minipage}[b]{.24\linewidth}
  \centering
  \centerline{\includegraphics[width=4cm]{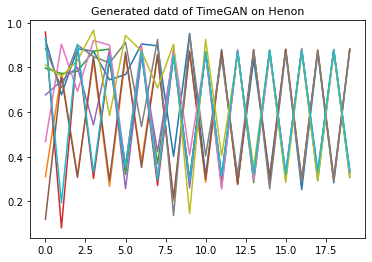}}
   %\hspace{-1.1cm}
  \centerline{}\medskip
\end{minipage}
\hfill
\begin{minipage}[b]{0.24\linewidth}
  \centering
  \centerline{\includegraphics[width=4cm]{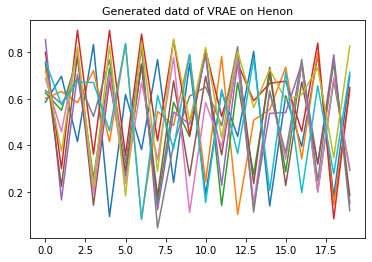}}
  %\hspace{-1.1cm}
  \centerline{}\medskip
\end{minipage}
\hfill
\begin{minipage}[b]{0.24\linewidth}
  \centering
  \centerline{\includegraphics[width=4cm]{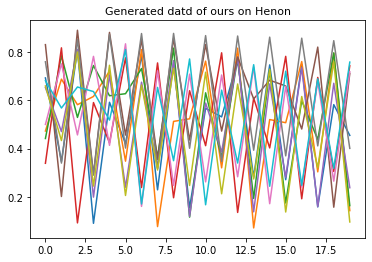}}
  %\hspace{-1.1cm}
  \centerline{}\medskip
\end{minipage}

\vspace{-0.3cm}
\begin{minipage}[b]{0.24\linewidth}
  \centering
  \centerline{\includegraphics[width=4cm]{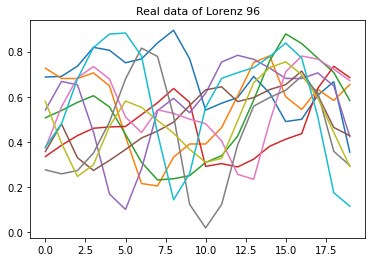}}
  %\hspace{-1.1cm}
  \centerline{}\medskip
   %\hspace{-1.1cm}
\end{minipage}
\hfill
\begin{minipage}[b]{.24\linewidth}
  \centering
  \centerline{\includegraphics[width=4cm]{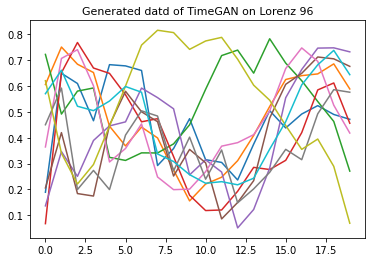}}
   %\hspace{-1.1cm}
  \centerline{}\medskip
\end{minipage}
\hfill
\begin{minipage}[b]{0.24\linewidth}
  \centering
  \centerline{\includegraphics[width=4cm]{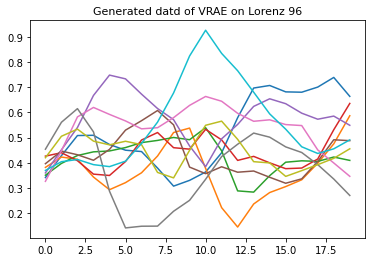}}
  %\hspace{-1.1cm}
  \centerline{}\medskip
\end{minipage}
\hfill
\begin{minipage}[b]{0.24\linewidth}
  \centering
  \centerline{\includegraphics[width=4cm]{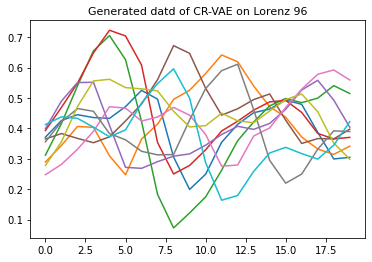}}
  %\hspace{-1.1cm}
  \centerline{}\medskip
\end{minipage}

\vspace{-0.3cm}
\begin{minipage}[b]{0.24\linewidth}
  \centering
  \centerline{\includegraphics[width=4cm]{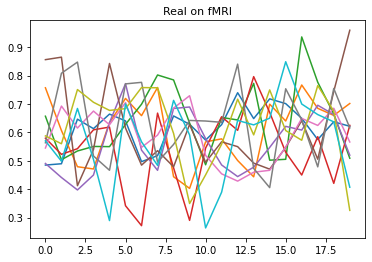}}
  %\hspace{-1.1cm}
  \centerline{}\medskip
   %\hspace{-1.1cm}
\end{minipage}
\hfill
\begin{minipage}[b]{.24\linewidth}
  \centering
  \centerline{\includegraphics[width=4cm]{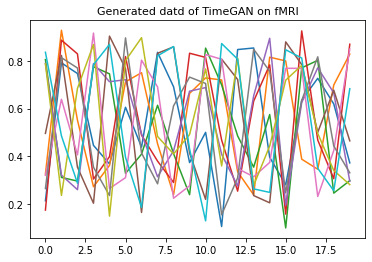}}
   %\hspace{-1.1cm}
  \centerline{}\medskip
\end{minipage}
\hfill
\begin{minipage}[b]{0.24\linewidth}
  \centering
  \centerline{\includegraphics[width=4cm]{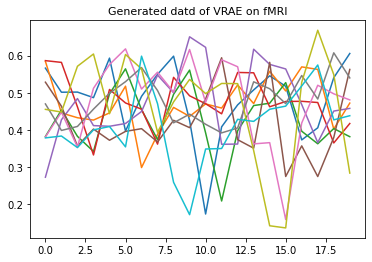}}
  %\hspace{-1.1cm}
  \centerline{}\medskip
\end{minipage}
\hfill
\begin{minipage}[b]{0.24\linewidth}
  \centering
  \centerline{\includegraphics[width=4cm]{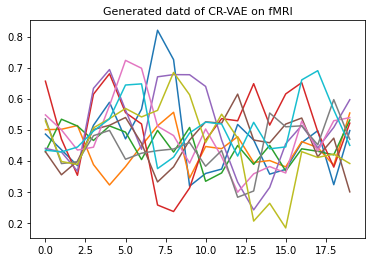}}
  %\hspace{-1.1cm}
  \centerline{}\medskip
\end{minipage}

\vspace{-0.3cm}
\begin{minipage}[b]{0.24\linewidth}
  \centering
  \centerline{\includegraphics[width=4cm]{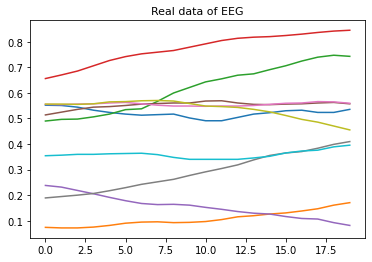}}
  %\hspace{-1.1cm}
  \centerline{Real Data}\medskip
   %\hspace{-1.1cm}
\end{minipage}
\hfill
\begin{minipage}[b]{.24\linewidth}
  \centering
  \centerline{\includegraphics[width=4cm]{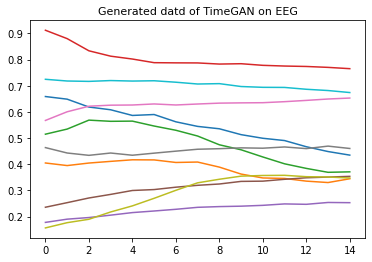}}
   %\hspace{-1.1cm}
  \centerline{TimeGAN}\medskip
\end{minipage}
\hfill
\begin{minipage}[b]{0.24\linewidth}
  \centering
  \centerline{\includegraphics[width=4cm]{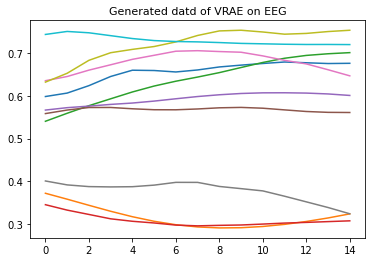}}
  %\hspace{-1.1cm}
  \centerline{VRAE}\medskip
\end{minipage}
\hfill
\begin{minipage}[b]{0.24\linewidth}
  \centering
  \centerline{\includegraphics[width=4cm]{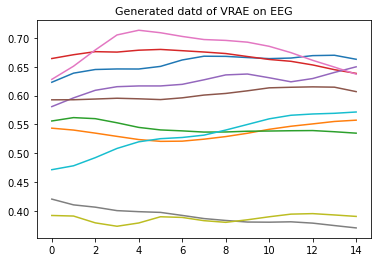}}
  %\hspace{-1.1cm}
  \centerline{Ours}\medskip
\end{minipage}

\caption{Generated data on H\'enon (1st row), Lorenz 96  (2nd row), fMRI (3rd row) and EEG (4th row). We plot 10 generated sequences to show the texture on average.}
\label{generate}
\end{figure}

Here, we show $10$ generated
sequences for each subplot in Fig.~\ref{generate} to show the ``texture" of time series. We can find that our results are similar to VRAE and outperform TimeGAN.

\clearpage
\newpage
\mbox{~}
\subsection{Prediction Results of TSTR}

Up to now, we have quantitatively evaluated our generated samples using TSTR, measured by RMSE. Here, we show the predicted curves to evaluate our generated samples qualitatively. More specific, we train the two-layer RNN on synthetic data generated by CR-VAE, and then use the trained RNN to perform one-step prediction on real data.

\begin{figure}[htp]

\begin{minipage}[b]{0.24\linewidth}
  \centering
  \centerline{\includegraphics[width=4cm]{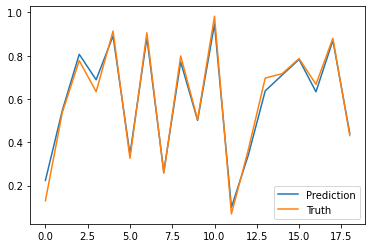}}
  %\hspace{-1.1cm}
  \centerline{H\'enon}\medskip
   %\hspace{-1.1cm}
\end{minipage}
\hfill
\begin{minipage}[b]{.24\linewidth}
  \centering
  \centerline{\includegraphics[width=4cm]{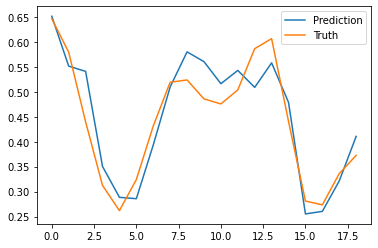}}
   %\hspace{-1.1cm}
  \centerline{Lorenz 96}\medskip
\end{minipage}
\hfill
\begin{minipage}[b]{0.24\linewidth}
  \centering
  \centerline{\includegraphics[width=4cm]{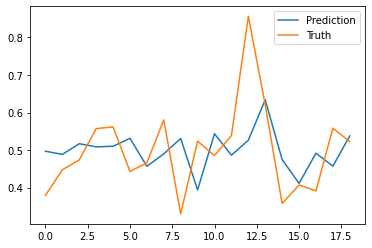}}
  %\hspace{-1.1cm}
  \centerline{fMRI}\medskip
   %\hspace{-1.1cm}
\end{minipage}
\hfill
\begin{minipage}[b]{.24\linewidth}
  \centering
  \centerline{\includegraphics[width=4cm]{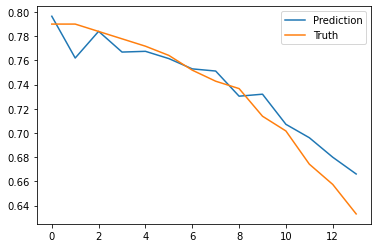}}
   %\hspace{-1.1cm}
  \centerline{EEG}\medskip
\end{minipage}

\caption{Prediction Results of CR-VAE on different datasets.}
\label{TSTR}
\end{figure}

Fig.~\ref{TSTR} illustrates the $20$-length sequences predicted by CR-VAE, step-wisely on real data. CR-VAE performs quite well on H\'enon, Lorenz $96$ and EEG. As can be seen, the predicted curves on H\'enon, Lorenz $96$ and EEG demonstrate markedly better overlap with the original curves.

%compared to fMRI. As can be seen from Fig.~\ref{TSTR}, predicted curves on Henon, Lorenz, EEG demonstrate markedly better overlap with the original curves than on fMRI. Regardless, though, predictions on fMRI can still capture the trend of fMRI sequences.

\subsection{PCA Visualization}

\begin{figure}[tb]

\vspace{-0.3cm}
\begin{minipage}[b]{0.24\linewidth}
  \centering
  \centerline{\includegraphics[width=4.8cm]{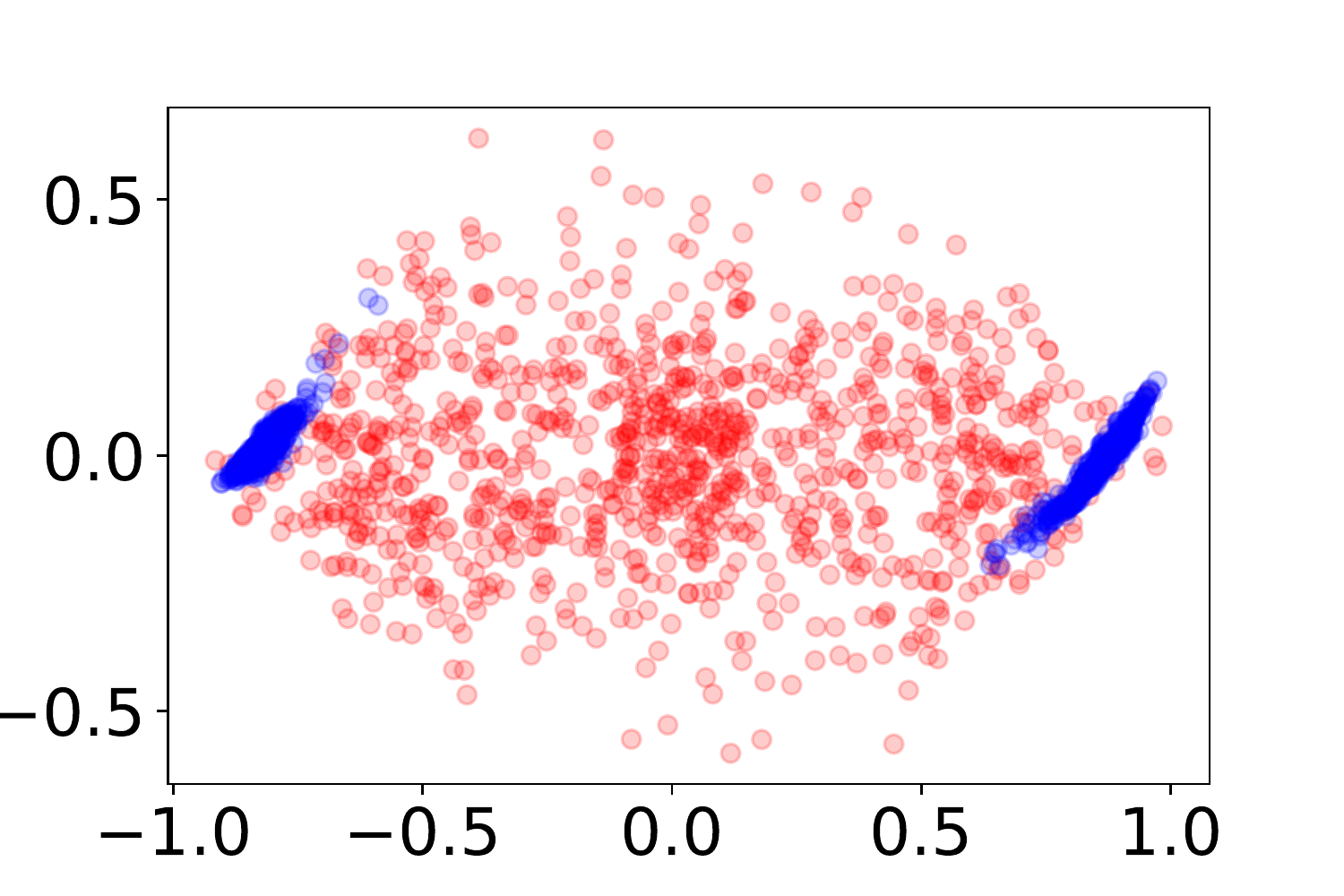}}
  %\hspace{-1.1cm}
  \centerline{}\medskip
   %\hspace{-1.1cm}
\end{minipage}
\hfill
\begin{minipage}[b]{.24\linewidth}
  \centering
  \centerline{\includegraphics[width=4.8cm]{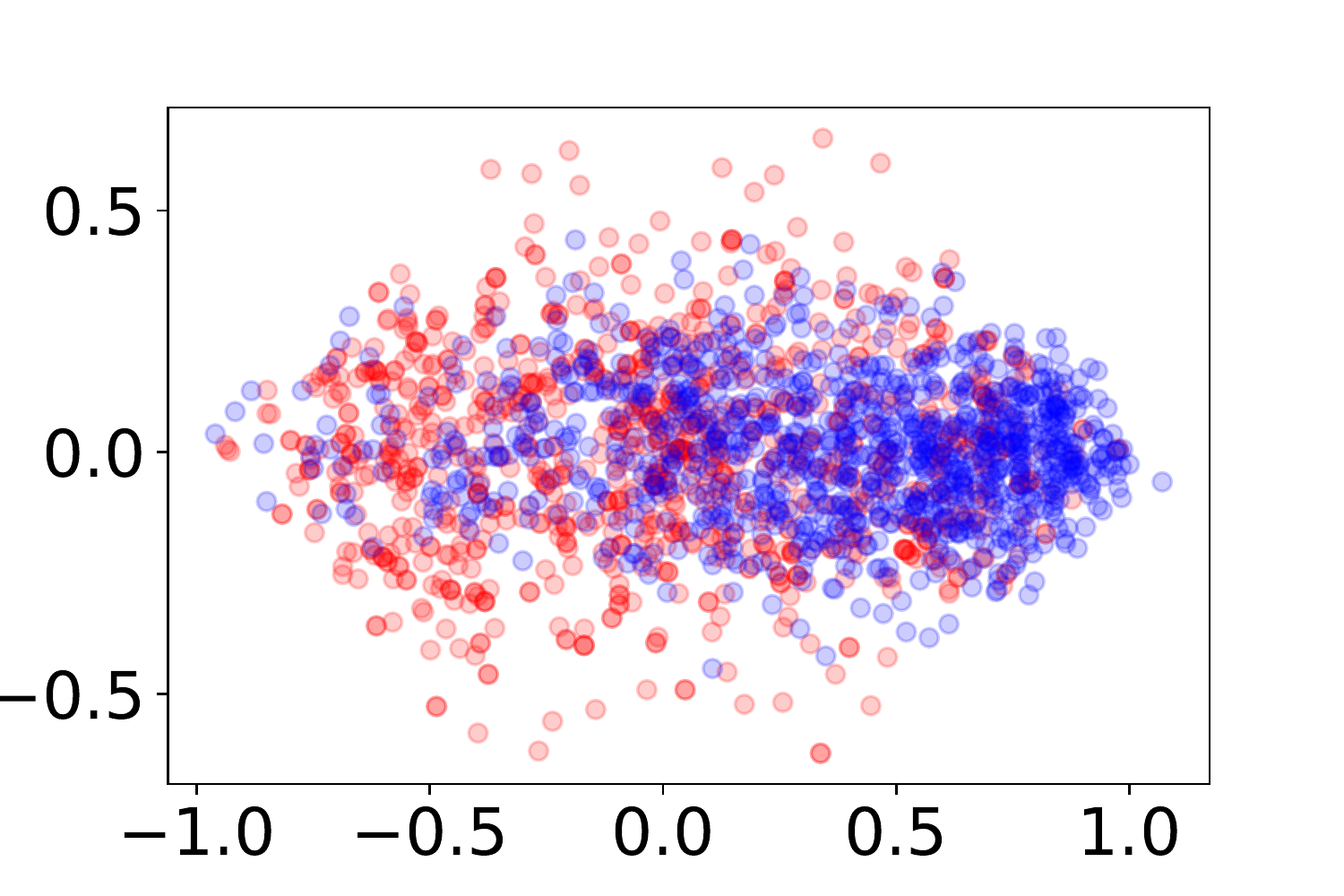}}
   %\hspace{-1.1cm}
  \centerline{}\medskip
\end{minipage}
\hfill
\begin{minipage}[b]{0.24\linewidth}
  \centering
  \centerline{\includegraphics[width=4.8cm]{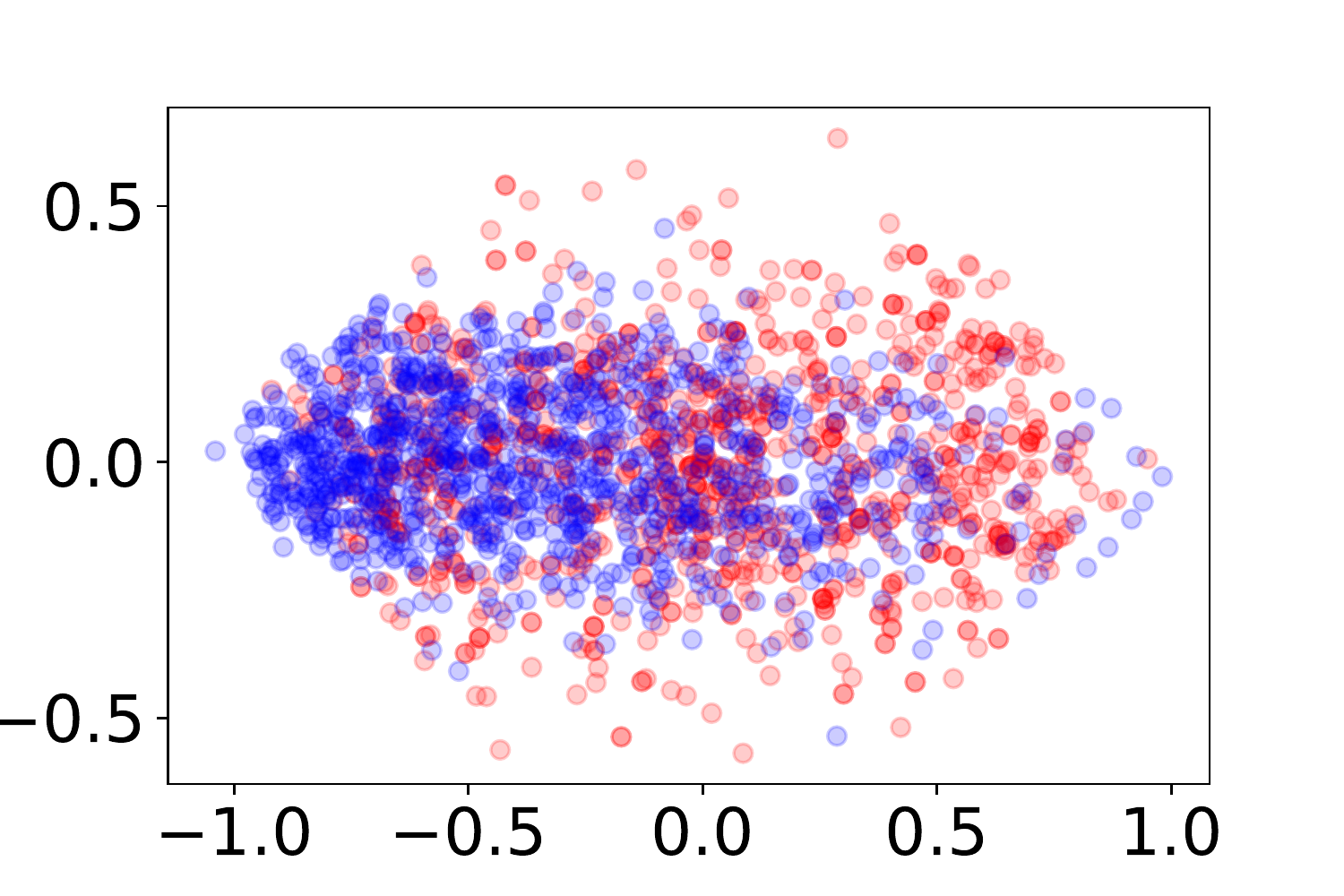}}
  %\hspace{-1.1cm}
  \centerline{}\medskip
\end{minipage}

\vspace{-0.5cm}

\begin{minipage}[b]{0.24\linewidth}
  \centering
  \centerline{\includegraphics[width=4.8cm]{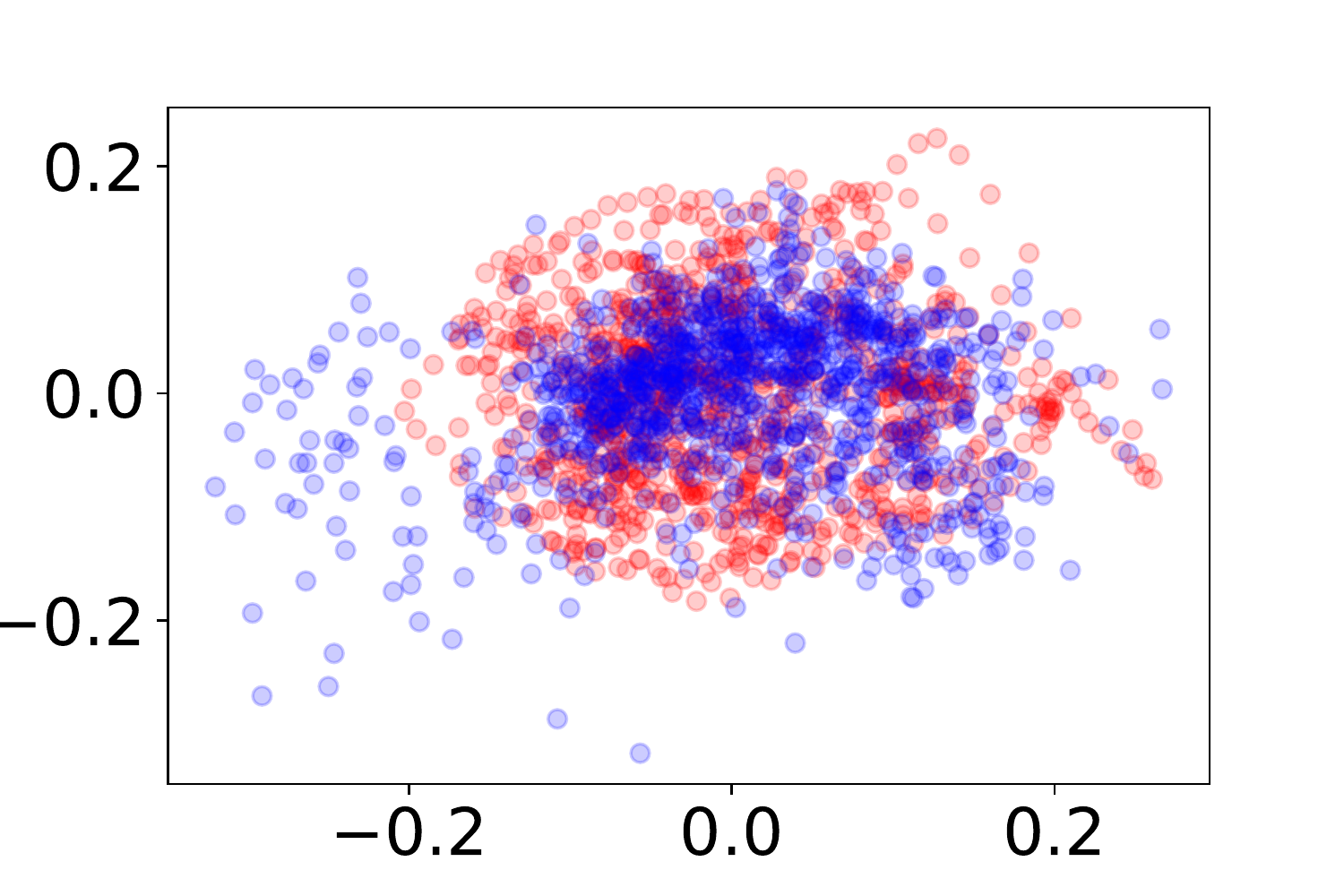}}
  % \hspace{0.cm}
  \centerline{}\medskip
\end{minipage}
\hfill
\begin{minipage}[b]{.24\linewidth}
  \centering
  \centerline{\includegraphics[width=4.8cm]{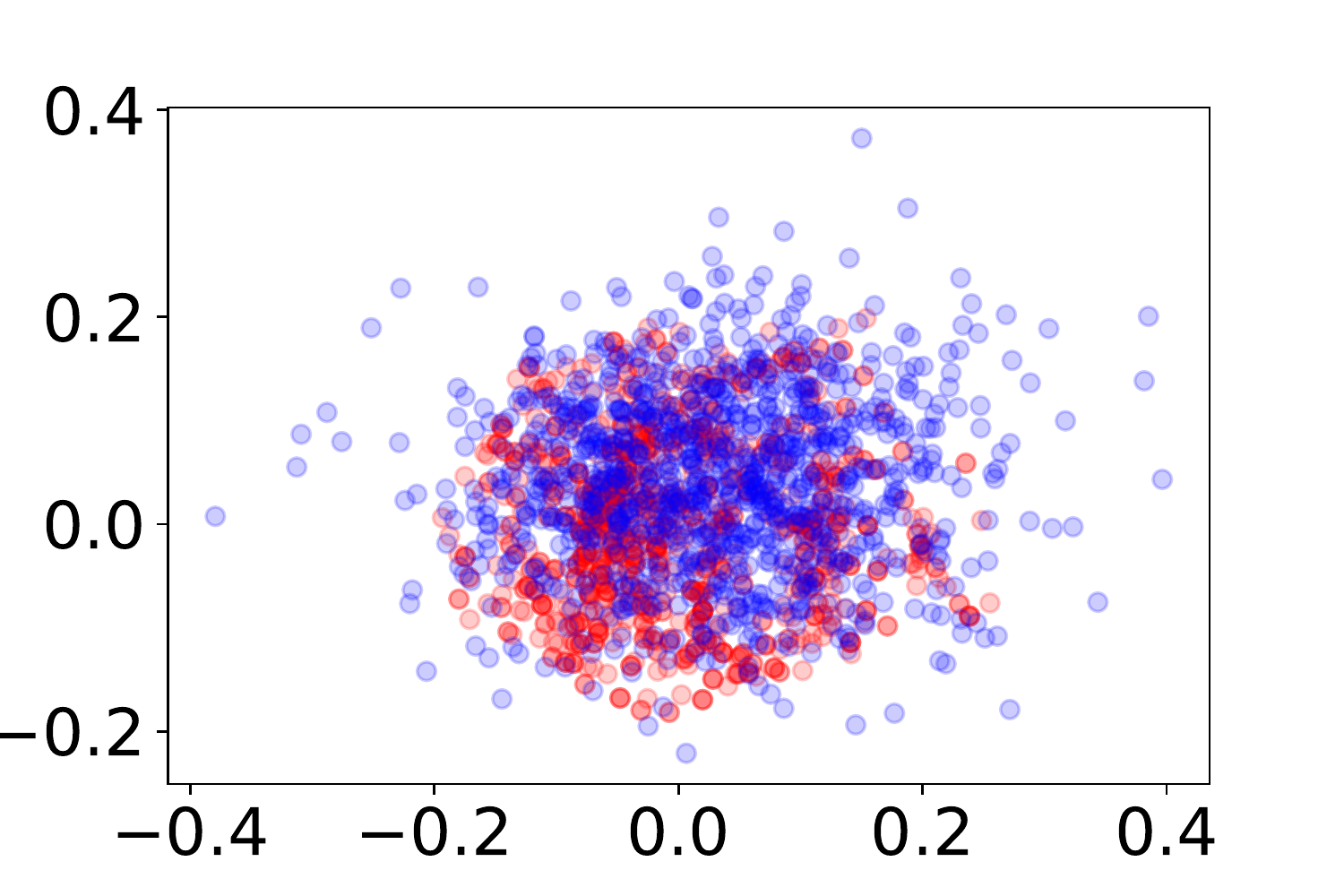}}
  % \hspace{0.cm}
  \centerline{}\medskip
\end{minipage}
\hfill
\begin{minipage}[b]{0.24\linewidth}
  \centering
  \centerline{\includegraphics[width=4.8cm]{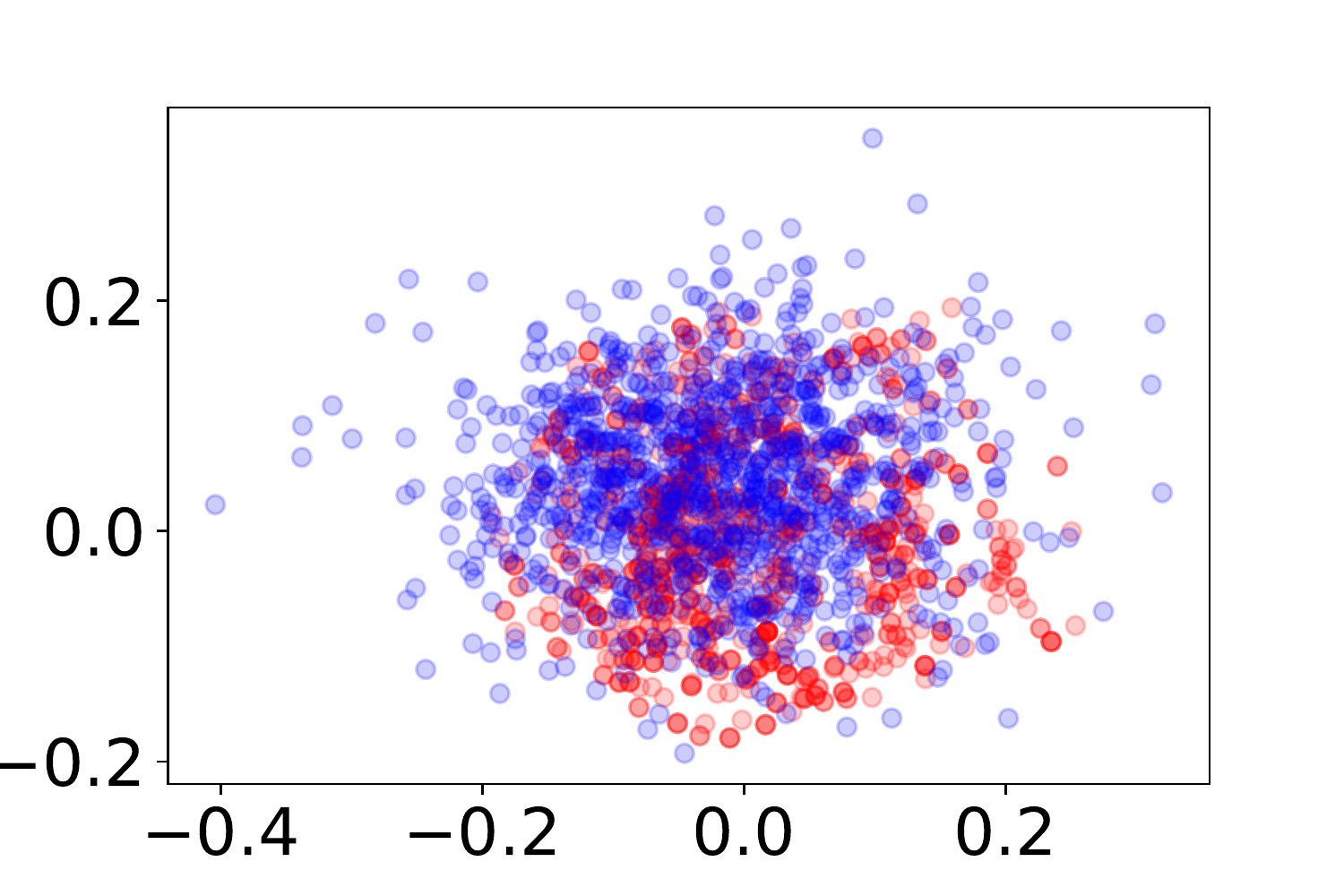}}
  % \hspace{0.cm}
  \centerline{}\medskip
\end{minipage}
\vspace{-0.5cm}

\begin{minipage}[b]{0.24\linewidth}
  \centering
  \centerline{\includegraphics[width=4.8cm]{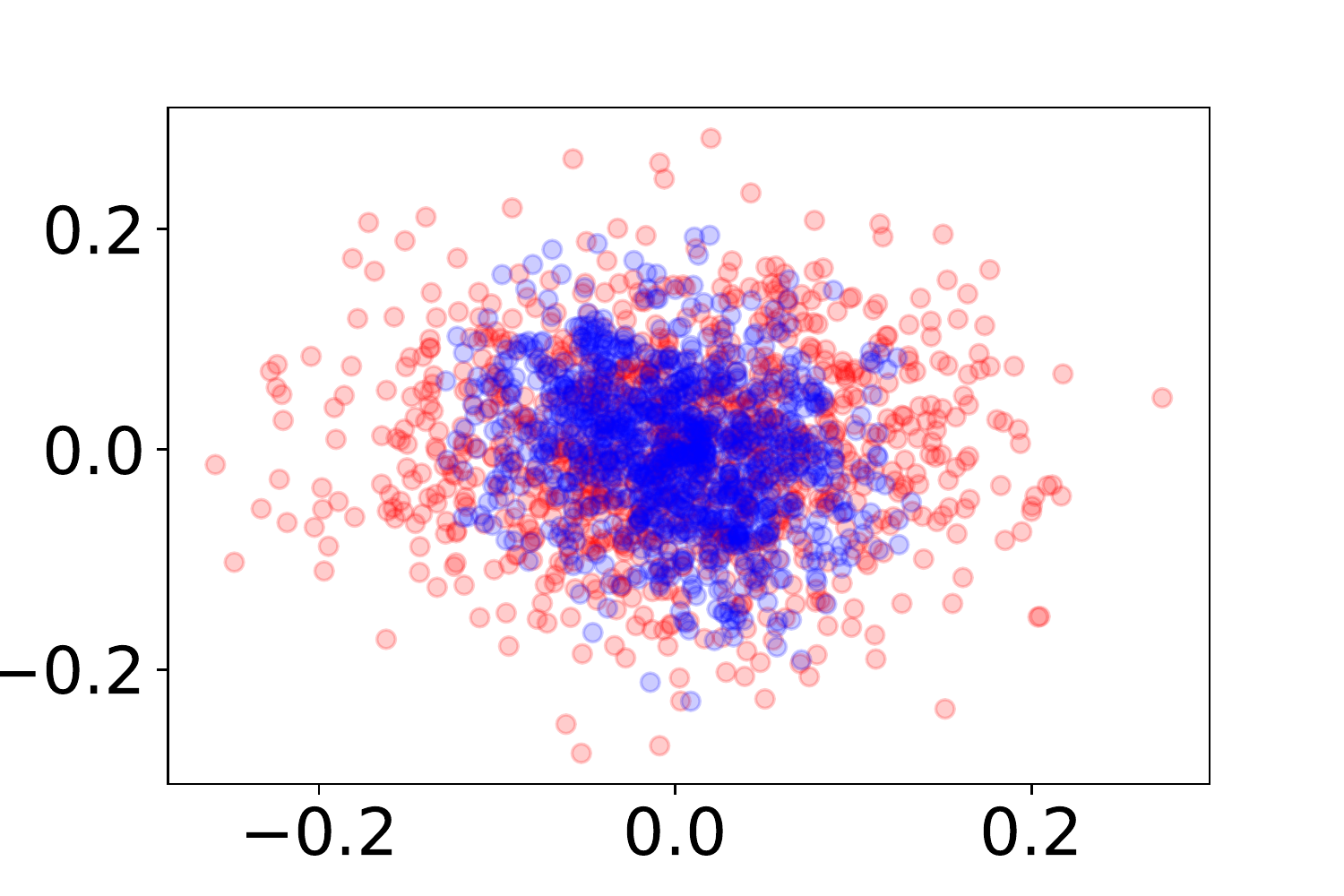}}
%  \vspace{1.5cm}
  \centerline{}\medskip
\end{minipage}
\hfill
\begin{minipage}[b]{.24\linewidth}
  \centering
  \centerline{\includegraphics[width=4.8cm]{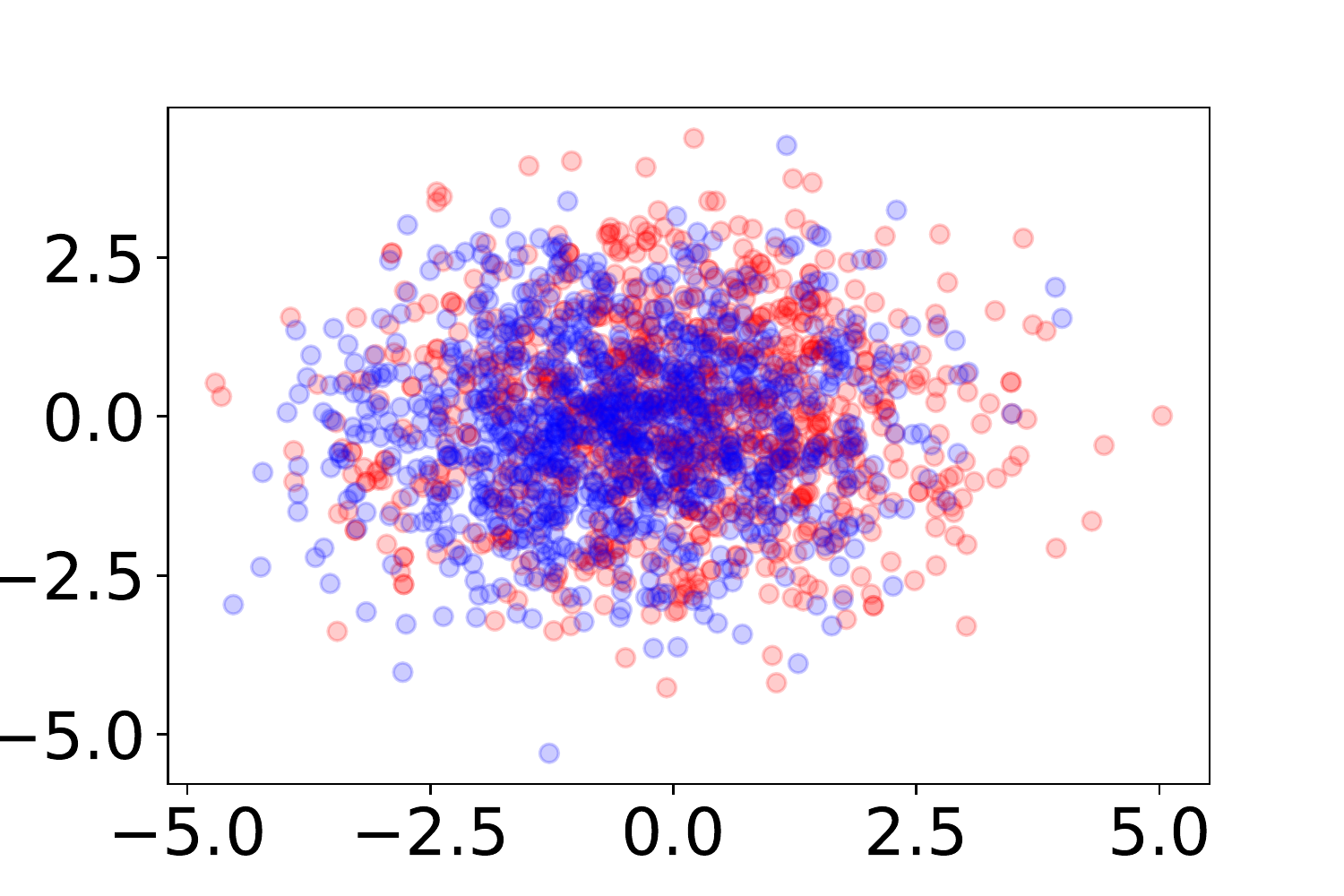}}
%  \vspace{1.5cm}
  \centerline{}\medskip
\end{minipage}
\hfill
\begin{minipage}[b]{0.24\linewidth}
  \centering
  \centerline{\includegraphics[width=4.8cm]{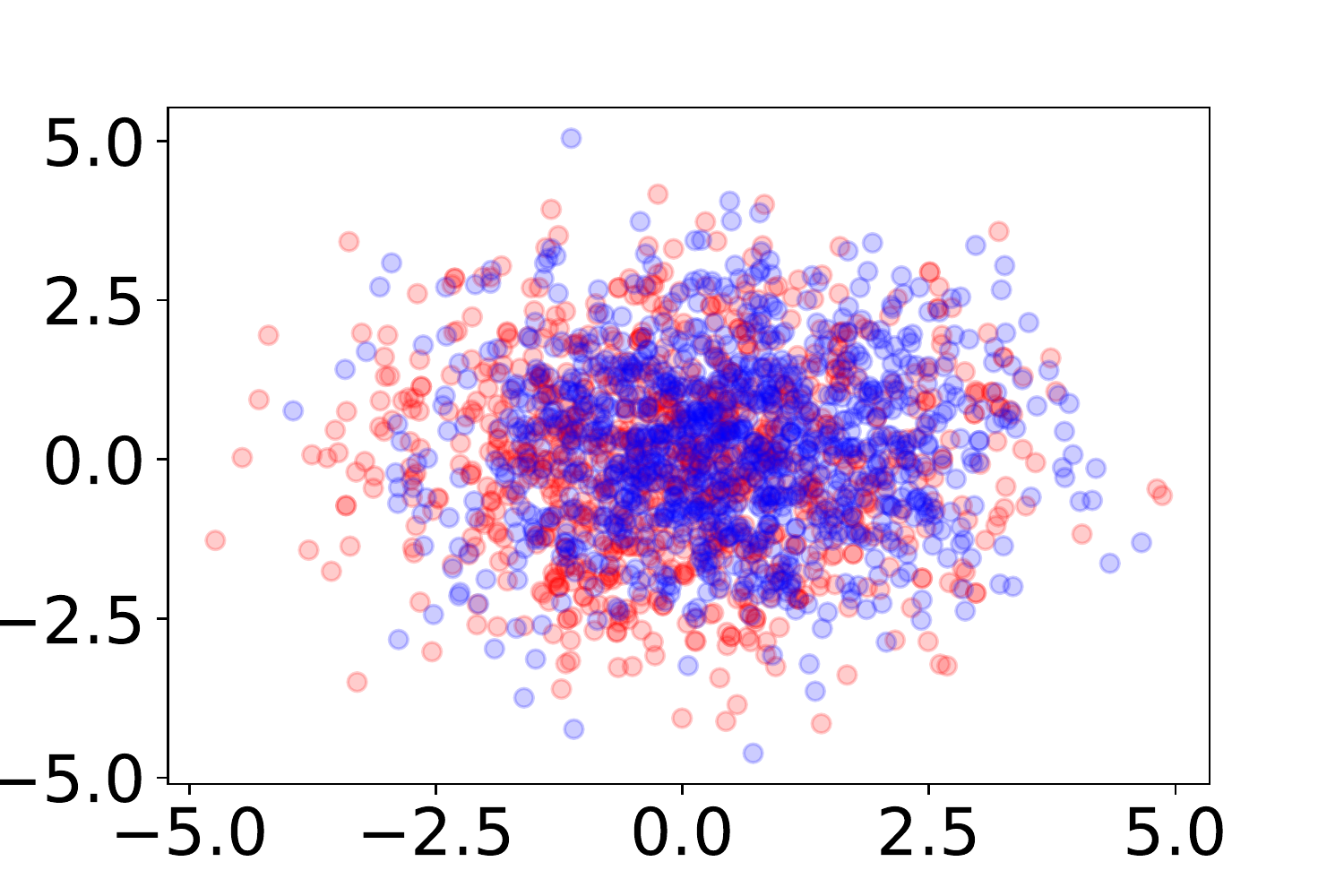}}
%  \vspace{1.5cm}
  \centerline{}\medskip
\end{minipage}

\vspace{-0.5cm}

\begin{minipage}[b]{0.24\linewidth}
  \centering
  \centerline{\includegraphics[width=4.8cm]{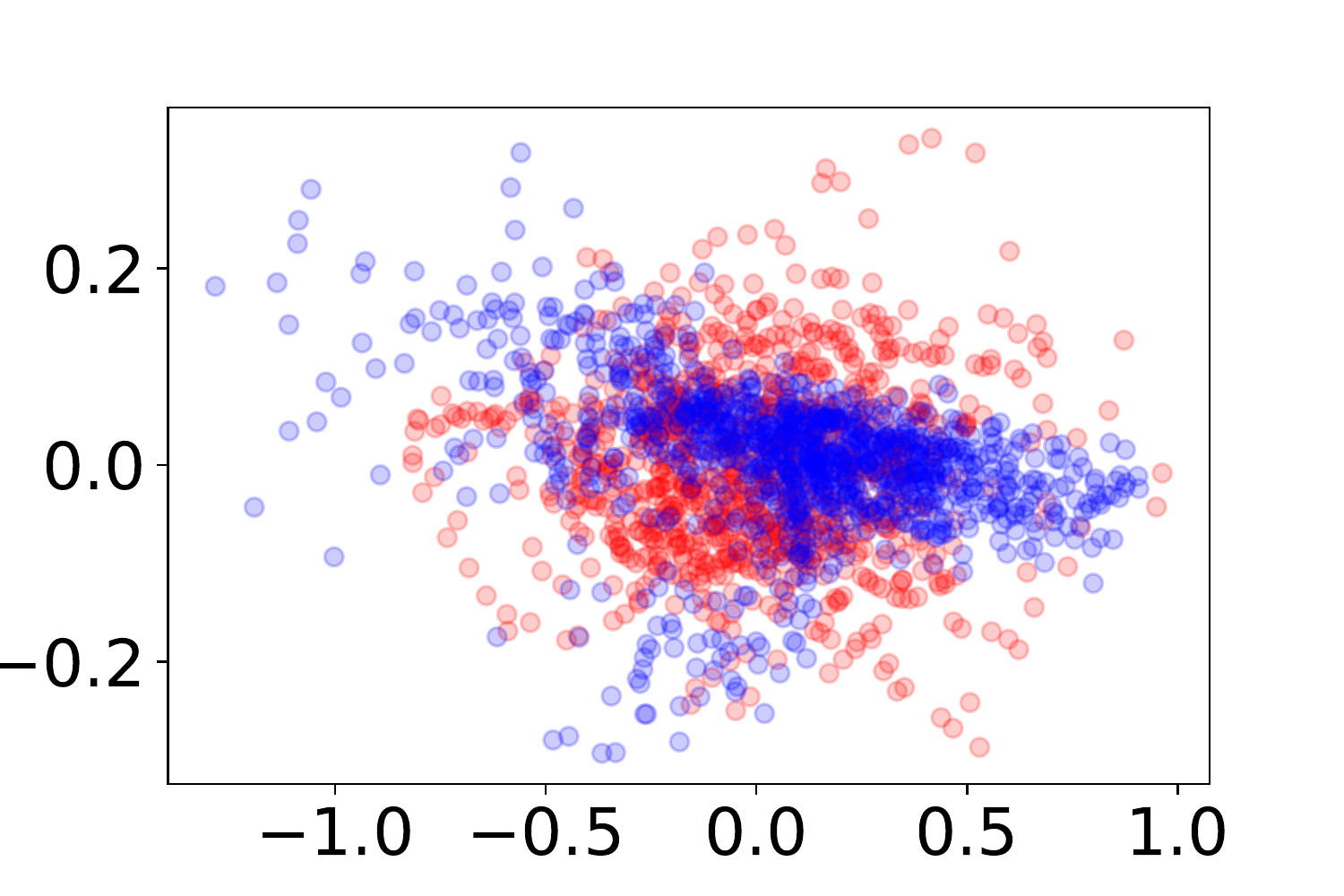}}
%  \vspace{1.5cm}
  \centerline{TimeGAN}\medskip
\end{minipage}
\hfill
\begin{minipage}[b]{.24\linewidth}
  \centering
  \centerline{\includegraphics[width=4.8cm]{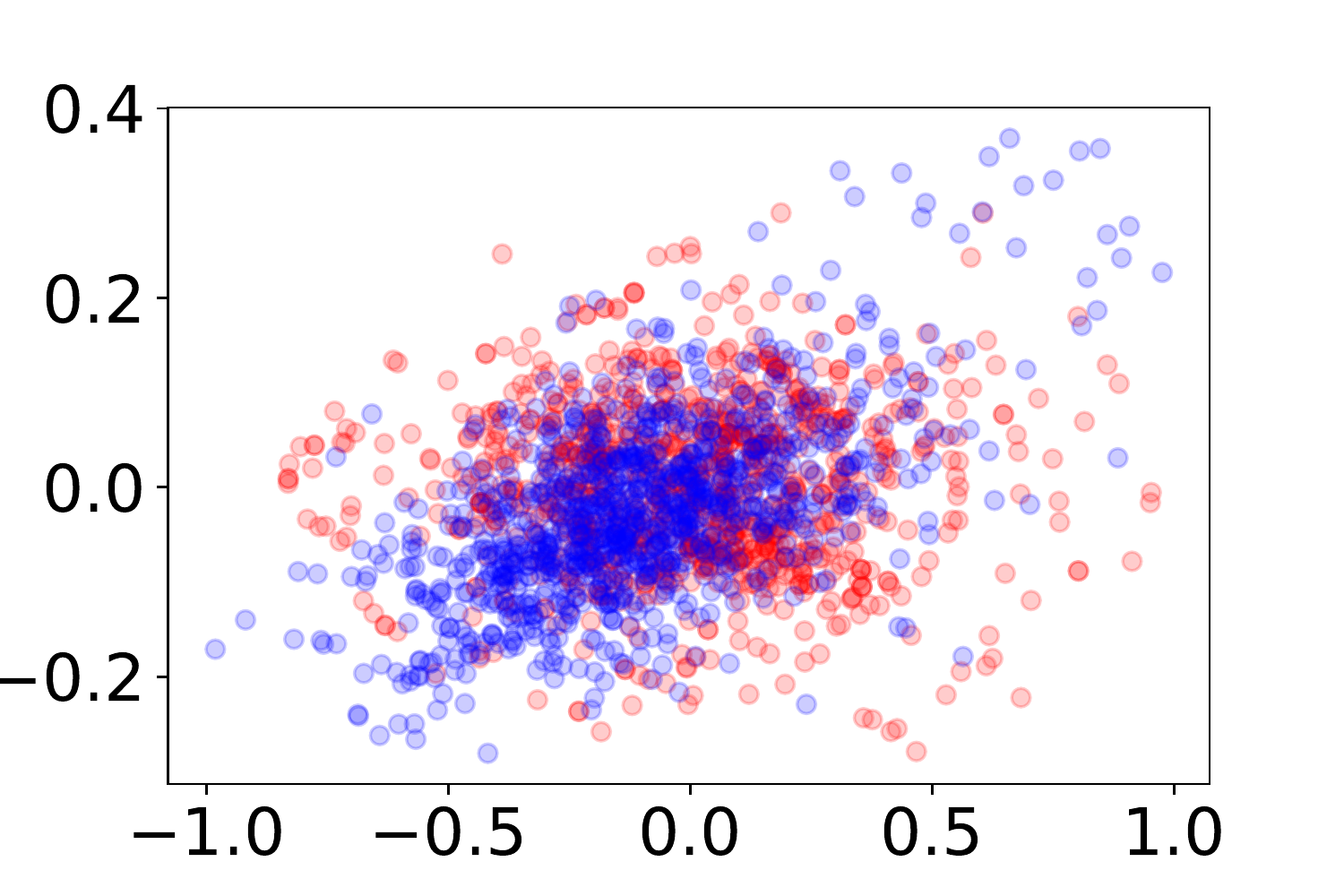}}
%  \vspace{1.5cm}
  \centerline{VRAE}\medskip
\end{minipage}
\hfill
\begin{minipage}[b]{0.24\linewidth}
  \centering
  \centerline{\includegraphics[width=4.8cm]{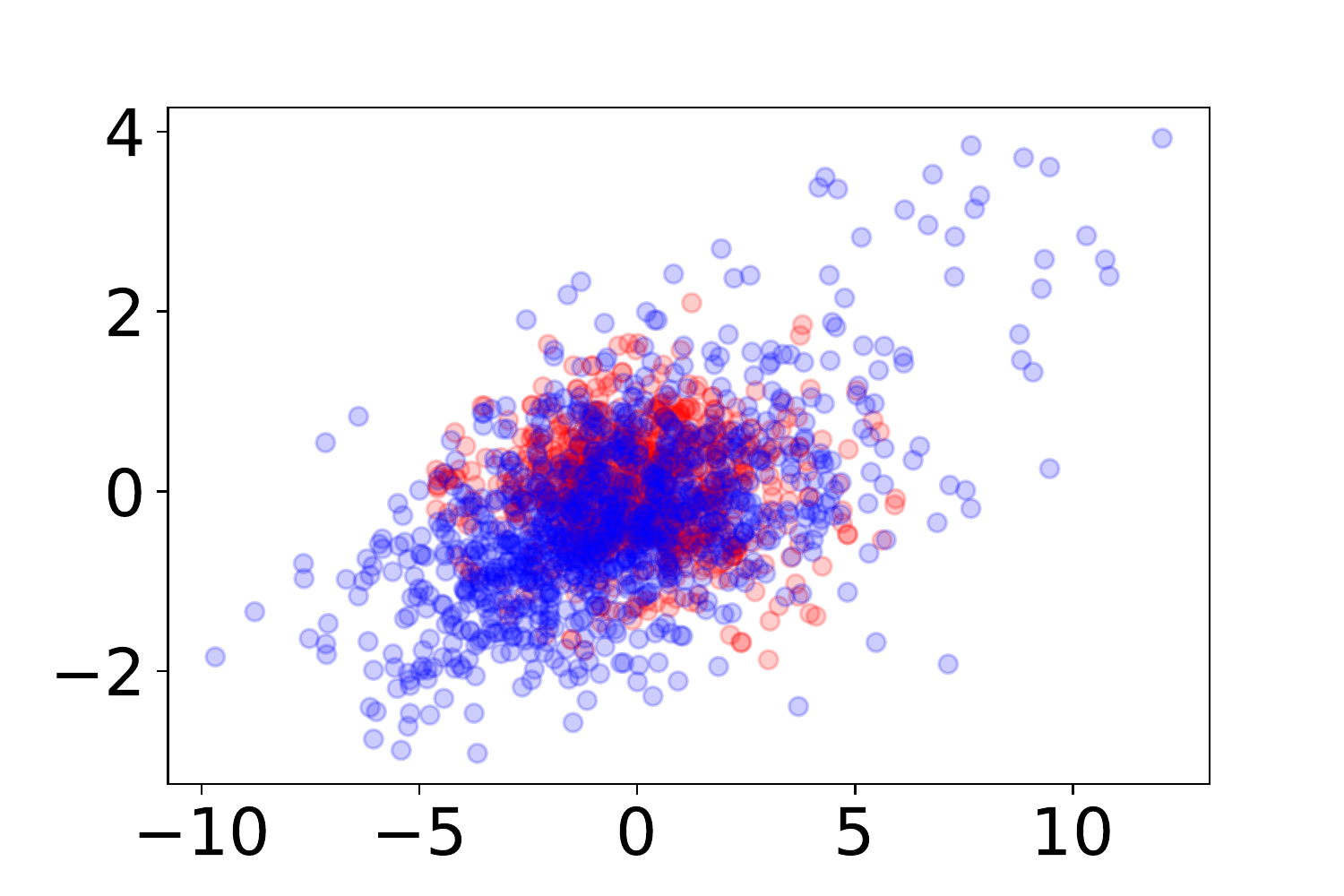}}
%  \vspace{1.5cm}
  \centerline{Ours}\medskip
\end{minipage}
\caption{PCA visualization on H\'enon (1st row), Lorenz 96  (2nd row), fMRI (3rd row) and EEG (4th row). \textcolor{red}{Red} samples correspond to real time series, whereas \textcolor{blue}{blue} samples correspond to synthetic time series.}
\label{pca}
\end{figure}

Due to page limits, we just show t-SNE visualization in the main manuscript. Here, we qualitatively evaluate the quality of generated time series by projecting both real and synthetic ones into a $2$-dimensional space with PCA in Fig.~\ref{pca}. We expect that a good generative model has close distributions for real and synthetic data, but most of synthetic points overlap with real points well. We conclude that PCA is not a good visualization algorithm for time series since it is not consistent with t-SNE, MMD, and TSTR results that are representative to evaluate generation results.  Note, the first PCA plot is quite different from others. This is because synthetic  results of TimeGAN on H\'enon are clustered to a few fixed values.

\newpage
\subsection{Ablation Study of the Two-Stage Learning Strategy}
As we mentioned in the main manuscript, invoking $\ell_1$ norm reduces the generation performance since it restricts the solution space of the network. To solve this problem, we propose a two-stage training strategy that we train our model using $\ell_1$ norm at first and then stop when the criteria is met or convergence. After that, we fix all zero weights, and continue training other weights using SGD. It improves the performance of generation significantly. 

\begin{figure}[h]
\centering
\noindent\begin{subfigure}[b]{0.2\textwidth}
    \centering 
    \includegraphics[width=1.0\textwidth]{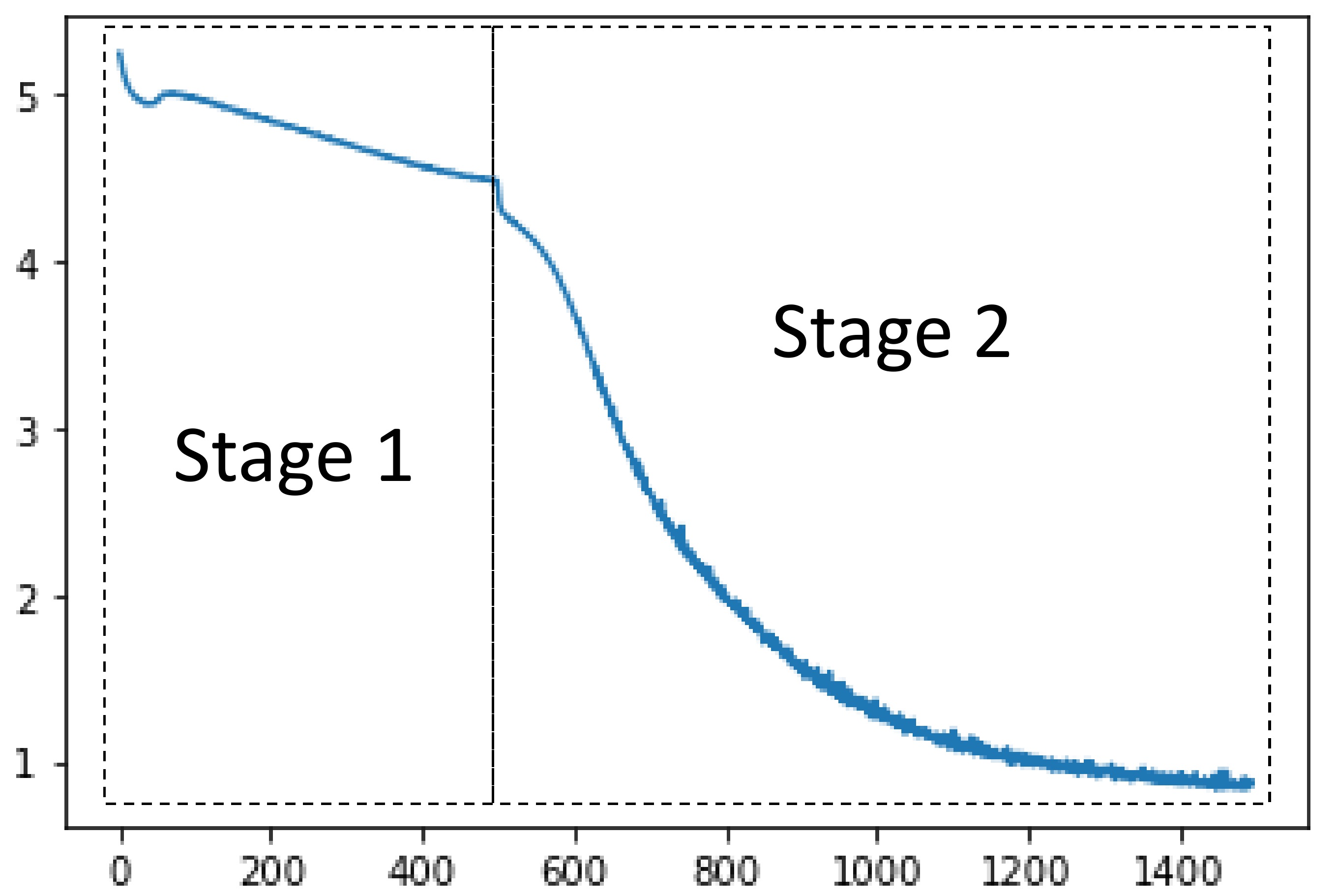}

\caption{The learning curve on fMRI}
\end{subfigure}%
\hspace{1.1cm}
\noindent\begin{subfigure}[b]{0.2\textwidth}
    \centering
    \includegraphics[width=1.05\textwidth]{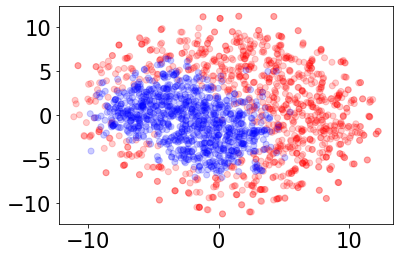}

\caption{t-SNE without stage 2 learning on fMRI}
\end{subfigure}%
\hspace{1.1cm}
\noindent\begin{subfigure}[b]{0.2\textwidth}
    \centering
    \includegraphics[width=1.05\textwidth]{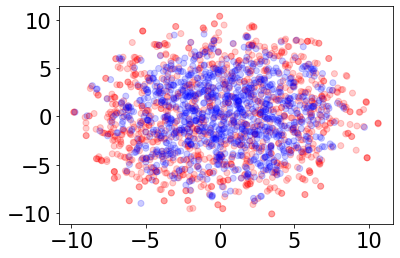}
    \caption{t-SNE with stage 2 learning on fMRI}
\end{subfigure}%
\caption{Learning curve and t-SNE visualization on fMRI: \textcolor{red}{red} samples correspond to real time series, whereas \textcolor{blue}{blue} samples correspond to synthetic time series.}
\label{fig.abl}
\end{figure}

Fig.~\ref{fig.abl}(a) shows the learning curve of CR-VAE on fMRI dataset. We can see that the loss drops significantly after we start our stage II learning. Fig.~\ref{fig.abl}(b) and Fig.~\ref{fig.abl}(c) compare the generation results with and without stage II learning using t-SNE projections. The stage II learning leads to a significant performance gain.

\newpage
\section{Minimal Structure of CR-VAE in PyTorch and Code Link}
PyTorch code of CR-VAE is available at \url{https://github.com/hongmingli1995/CR-VAE}.

We also provide the minimal structure of CR-VAE for better understanding.

\begin{lstlisting}
class CRVAE(nn.Module):
    def __init__(self, num_series, connection, hidden):
        '''
        CRVAE model has num_series decoder GRUs but has only one encoder GRU.

        Args:
          num_series: Dimension of multivariate time series.
          connection: Adjacency matrices of causal graphs. 
          Used to prune zero edges in stage 2 learning
          hidden: Number of units in GRU and 1-D CNN.
        '''
        super(CRVAE, self).__init__()
        
        self.device = torch.device('cuda')
        self.p = num_series
        self.hidden = hidden
        
        self.lstm_left = nn.GRU(num_series, hidden, batch_first=True)
        self.lstm_left.flatten_parameters()
        
        self.fc_mu = nn.Linear(hidden, hidden)
        self.fc_std = nn.Linear(hidden, hidden)
        self.connection = connection

        # num_series GRUs for num_series time series
        self.networks = nn.ModuleList([
            GRU(int(connection[:,i].sum()), hidden) for i in range(num_series)])

    def forward(self, X, hidden=None, mode = 'train'):
        '''
        Args:
          X: torch tensor of shape (batch_size, Time Length, Dimension).
          hidden: initail hidden states of GRU.
        '''
        T = X.shape[1]     
        X = torch.cat((torch.zeros(X.shape,device = self.device)[:,0:1,:],X),1)

        ###train the model using real data###
        if mode == 'train':
            

            hidden_0 = torch.zeros(1, X.shape[0], self.hidden, device=self.device)
            out, h_t = self.lstm_left(X[:,1:(T/2)+1,:], hidden_0.detach())
            
            mu = self.fc_mu(h_t)
            log_var = self.fc_std(h_t)
            
            sigma = torch.exp(0.5*log_var)
            z = torch.randn(size = mu.size())
            z = z.type_as(mu)
            z = mu + sigma*z

            pred = [self.networks[i](X, z, self.connection[:,i])[0]
                  for i in range(self.p)]

            return pred, log_var, mu
        
        ### generate the synthetic date iteratively.###
        if mode == 'generate':
            X_seq = torch.zeros(X[:,:1,:].shape).to(self.device)
            h_0 = torch.randn(size = (1,
            X_seq[:,-2:-1,:].size(0),self.hidden)).to(self.device)
            ht_last =[]
            for i in range(self.p):
                ht_last.append(h_0)
            for i in range(int(T/2)+1):
                
                ht_new = []
                for j in range(self.p):
                    out, h_t = self.networks[j](X_seq[:,-1:,:],
                    ht_last[j], self.connection[:,j])
                    if j == 0:
                        X_t = out
                    else:
                        X_t = torch.cat((X_t,out),-1)
                    ht_new.append(h_t)
                ht_last = ht_new
                if i ==0:
                    X_seq = X_t
                else:
                    X_seq = torch.cat([X_seq,X_t],dim = 1)
                
            return X_seq
\end{lstlisting}
\bibliography{aaai23}

\begin{thebibliography}{59}
\providecommand{\natexlab}[1]{#1}

\bibitem[{Amblard and Michel(2012)}]{amblard2012relation}
Amblard, P.-O.; and Michel, O.~J. 2012.
\newblock The relation between Granger causality and directed information
  theory: A review.
\newblock \emph{Entropy}, 15(1): 113--143.

\bibitem[{Assaad, Devijver, and
  Gaussier(2022{\natexlab{a}})}]{assaad2022causal}
Assaad, C.~K.; Devijver, E.; and Gaussier, E. 2022{\natexlab{a}}.
\newblock Causal Discovery of Extended Summary Graphs in Time Series.
\newblock In \emph{The 38th Conference on Uncertainty in Artificial
  Intelligence}.

\bibitem[{Assaad, Devijver, and
  Gaussier(2022{\natexlab{b}})}]{assaad2022survey}
Assaad, C.~K.; Devijver, E.; and Gaussier, E. 2022{\natexlab{b}}.
\newblock Survey and Evaluation of Causal Discovery Methods for Time Series.
\newblock \emph{Journal of Artificial Intelligence Research}, 73: 767--819.

\bibitem[{Barnett, Barrett, and Seth(2009)}]{barnett2009granger}
Barnett, L.; Barrett, A.~B.; and Seth, A.~K. 2009.
\newblock Granger causality and transfer entropy are equivalent for Gaussian
  variables.
\newblock \emph{Physical review letters}, 103(23): 238701.

\bibitem[{Bengio et~al.(2015)Bengio, Vinyals, Jaitly, and
  Shazeer}]{bengio2015scheduled}
Bengio, S.; Vinyals, O.; Jaitly, N.; and Shazeer, N. 2015.
\newblock Scheduled sampling for sequence prediction with recurrent neural
  networks.
\newblock \emph{Advances in neural information processing systems}, 28.

\bibitem[{Chambolle et~al.(1998)Chambolle, De~Vore, Lee, and
  Lucier}]{chambolle1998nonlinear}
Chambolle, A.; De~Vore, R.~A.; Lee, N.-Y.; and Lucier, B.~J. 1998.
\newblock Nonlinear wavelet image processing: variational problems,
  compression, and noise removal through wavelet shrinkage.
\newblock \emph{IEEE Transactions on Image Processing}, 7(3): 319--335.

\bibitem[{Chen, Feng, and Lu(2021)}]{chen2021wiener}
Chen, J.; Feng, J.; and Lu, W. 2021.
\newblock A Wiener causality defined by divergence.
\newblock \emph{Neural Processing Letters}, 53(3): 1773--1794.

\bibitem[{Chen et~al.(2004)Chen, Rangarajan, Feng, and
  Ding}]{chen2004analyzing}
Chen, Y.; Rangarajan, G.; Feng, J.; and Ding, M. 2004.
\newblock Analyzing multiple nonlinear time series with extended Granger
  causality.
\newblock \emph{Physics letters A}, 324(1): 26--35.

\bibitem[{Cho et~al.(2014)Cho, van Merri{\"e}nboer, Bahdanau, and
  Bengio}]{cho2014properties}
Cho, K.; van Merri{\"e}nboer, B.; Bahdanau, D.; and Bengio, Y. 2014.
\newblock On the properties of neural machine translation: Encoder--decoder
  approaches.
\newblock In \emph{8th Workshop on Syntax, Semantics and Structure in
  Statistical Translation, SSST 2014}, 103--111. Association for Computational
  Linguistics (ACL).

\bibitem[{Chu, Glymour, and Ridgeway(2008)}]{chu2008search}
Chu, T.; Glymour, C.; and Ridgeway, G. 2008.
\newblock Search for Additive Nonlinear Time Series Causal Models.
\newblock \emph{Journal of Machine Learning Research}, 9(5).

\bibitem[{Chung et~al.(2015)Chung, Kastner, Dinh, Goel, Courville, and
  Bengio}]{NIPS2015_b618c321}
Chung, J.; Kastner, K.; Dinh, L.; Goel, K.; Courville, A.~C.; and Bengio, Y.
  2015.
\newblock A Recurrent Latent Variable Model for Sequential Data.
\newblock In Cortes, C.; Lawrence, N.; Lee, D.; Sugiyama, M.; and Garnett, R.,
  eds., \emph{Advances in Neural Information Processing Systems}, volume~28.
  Curran Associates, Inc.

\bibitem[{Daubechies, Defrise, and De~Mol(2004)}]{daubechies2004iterative}
Daubechies, I.; Defrise, M.; and De~Mol, C. 2004.
\newblock An iterative thresholding algorithm for linear inverse problems with
  a sparsity constraint.
\newblock \emph{Communications on Pure and Applied Mathematics: A Journal
  Issued by the Courant Institute of Mathematical Sciences}, 57(11):
  1413--1457.

\bibitem[{De~La Pava~Panche, Alvarez-Meza, and
  Orozco-Gutierrez(2019)}]{de2019data}
De~La Pava~Panche, I.; Alvarez-Meza, A.~M.; and Orozco-Gutierrez, A. 2019.
\newblock A data-driven measure of effective connectivity based on Renyi's
  $\alpha$-entropy.
\newblock \emph{Frontiers in neuroscience}, 13: 1277.

\bibitem[{Desai et~al.(2021)Desai, Freeman, Wang, and
  Beaver}]{desai2021timevae}
Desai, A.; Freeman, C.; Wang, Z.; and Beaver, I. 2021.
\newblock TimeVAE: A Variational Auto-Encoder for Multivariate Time Series
  Generation.
\newblock \emph{arXiv preprint arXiv:2111.08095}.

\bibitem[{Deshpande et~al.(2009)Deshpande, LaConte, James, Peltier, and
  Hu}]{deshpande2009multivariate}
Deshpande, G.; LaConte, S.; James, G.~A.; Peltier, S.; and Hu, X. 2009.
\newblock Multivariate Granger causality analysis of fMRI data.
\newblock \emph{Human brain mapping}, 30(4): 1361--1373.

\bibitem[{Esteban, Hyland, and R{\"a}tsch(2017)}]{esteban2017real}
Esteban, C.; Hyland, S.~L.; and R{\"a}tsch, G. 2017.
\newblock Real-valued (medical) time series generation with recurrent
  conditional gans.
\newblock \emph{arXiv preprint arXiv:1706.02633}.

\bibitem[{Fabius and Van~Amersfoort(2014)}]{fabius2014variational}
Fabius, O.; and Van~Amersfoort, J.~R. 2014.
\newblock Variational recurrent auto-encoders.
\newblock \emph{arXiv preprint arXiv:1412.6581}.

\bibitem[{Fraccaro et~al.(2016)Fraccaro, S{\o}nderby, Paquet, and
  Winther}]{fraccaro2016sequential}
Fraccaro, M.; S{\o}nderby, S.~K.; Paquet, U.; and Winther, O. 2016.
\newblock Sequential neural models with stochastic layers.
\newblock \emph{Advances in neural information processing systems}, 29.

\bibitem[{Giraldo, Rao, and Principe(2014)}]{giraldo2014measures}
Giraldo, L. G.~S.; Rao, M.; and Principe, J.~C. 2014.
\newblock Measures of entropy from data using infinitely divisible kernels.
\newblock \emph{IEEE Transactions on Information Theory}, 61(1): 535--548.

\bibitem[{Goodfellow et~al.(2014)Goodfellow, Pouget-Abadie, Mirza, Xu,
  Warde-Farley, Ozair, Courville, and Bengio}]{goodfellow2014generative}
Goodfellow, I.; Pouget-Abadie, J.; Mirza, M.; Xu, B.; Warde-Farley, D.; Ozair,
  S.; Courville, A.; and Bengio, Y. 2014.
\newblock Generative adversarial nets.
\newblock \emph{Advances in neural information processing systems}, 27.

\bibitem[{Goudet et~al.(2018)Goudet, Kalainathan, Caillou, Guyon, Lopez-Paz,
  and Sebag}]{goudet2018learning}
Goudet, O.; Kalainathan, D.; Caillou, P.; Guyon, I.; Lopez-Paz, D.; and Sebag,
  M. 2018.
\newblock Learning functional causal models with generative neural networks.
\newblock In \emph{Explainable and interpretable models in computer vision and
  machine learning}, 39--80. Springer.

\bibitem[{Goyal et~al.(2017)Goyal, Sordoni, C{\^o}t{\'e}, Ke, and
  Bengio}]{alias2017z}
Goyal, A.; Sordoni, A.; C{\^o}t{\'e}, M.-A.; Ke, N.~R.; and Bengio, Y. 2017.
\newblock Z-forcing: Training stochastic recurrent networks.
\newblock \emph{Advances in neural information processing systems}, 30.

\bibitem[{Granger(1969)}]{Granger1969}
Granger, C. W.~J. 1969.
\newblock {Investigating Causal Relations by Econometric Models and
  Cross-spectral Methods}.
\newblock \emph{Econometrica}, 37(3): 424.

\bibitem[{Gretton et~al.(2006)Gretton, Borgwardt, Rasch, Sch{\"o}lkopf, and
  Smola}]{gretton2006kernel}
Gretton, A.; Borgwardt, K.; Rasch, M.; Sch{\"o}lkopf, B.; and Smola, A. 2006.
\newblock A kernel method for the two-sample-problem.
\newblock \emph{Advances in neural information processing systems}, 19.

\bibitem[{Gretton et~al.(2012)Gretton, Borgwardt, Rasch, Sch{\"o}lkopf, and
  Smola}]{gretton2012kernel}
Gretton, A.; Borgwardt, K.~M.; Rasch, M.~J.; Sch{\"o}lkopf, B.; and Smola, A.
  2012.
\newblock A kernel two-sample test.
\newblock \emph{The Journal of Machine Learning Research}, 13(1): 723--773.

\bibitem[{Hoyer et~al.(2008)Hoyer, Janzing, Mooij, Peters, and
  Sch{\"o}lkopf}]{hoyer2008nonlinear}
Hoyer, P.; Janzing, D.; Mooij, J.~M.; Peters, J.; and Sch{\"o}lkopf, B. 2008.
\newblock Nonlinear causal discovery with additive noise models.
\newblock \emph{Advances in neural information processing systems}, 21.

\bibitem[{Huijse et~al.(2012)Huijse, Estevez, Protopapas, Zegers, and
  Principe}]{huijse2012information}
Huijse, P.; Estevez, P.~A.; Protopapas, P.; Zegers, P.; and Principe, J.~C.
  2012.
\newblock An information theoretic algorithm for finding periodicities in
  stellar light curves.
\newblock \emph{IEEE Transactions on Signal Processing}, 60(10): 5135--5145.

\bibitem[{Isaksson, Wennberg, and Zetterberg(1981)}]{isaksson1981computer}
Isaksson, A.; Wennberg, A.; and Zetterberg, L.~H. 1981.
\newblock Computer analysis of EEG signals with parametric models.
\newblock \emph{Proceedings of the IEEE}, 69(4): 451--461.

\bibitem[{Kingma and Welling(2013)}]{kingma2013auto}
Kingma, D.~P.; and Welling, M. 2013.
\newblock Auto-encoding variational bayes.
\newblock \emph{arXiv preprint arXiv:1312.6114}.

\bibitem[{Kramer(1998)}]{kramer1998causal}
Kramer, G. 1998.
\newblock Causal conditioning, directed information and the multiple-access
  channel with feedback.
\newblock In \emph{Proceedings. 1998 IEEE International Symposium on
  Information Theory (Cat. No. 98CH36252)}, 189. IEEE.

\bibitem[{Kramer, Kolaczyk, and Kirsch(2008)}]{kramer2008emergent}
Kramer, M.~A.; Kolaczyk, E.~D.; and Kirsch, H.~E. 2008.
\newblock Emergent network topology at seizure onset in humans.
\newblock \emph{Epilepsy research}, 79(2-3): 173--186.

\bibitem[{Kugiumtzis(2013)}]{kugiumtzis2013direct}
Kugiumtzis, D. 2013.
\newblock Direct-coupling information measure from nonuniform embedding.
\newblock \emph{Physical Review E}, 87(6): 062918.

\bibitem[{Liang et~al.(2021)Liang, Glossner, Wang, Shi, and
  Zhang}]{liang2021pruning}
Liang, T.; Glossner, J.; Wang, L.; Shi, S.; and Zhang, X. 2021.
\newblock Pruning and quantization for deep neural network acceleration: A
  survey.
\newblock \emph{Neurocomputing}, 461: 370--403.

\bibitem[{Litterman(1986)}]{litterman1986forecasting}
Litterman, R.~B. 1986.
\newblock Forecasting with Bayesian vector autoregressions—five years of
  experience.
\newblock \emph{Journal of Business \& Economic Statistics}, 4(1): 25--38.

\bibitem[{Liu et~al.(2020)Liu, Ji, Xun, Yao, Huai, and Zhang}]{liu2020ec}
Liu, J.; Ji, J.; Xun, G.; Yao, L.; Huai, M.; and Zhang, A. 2020.
\newblock EC-GAN: inferring brain effective connectivity via generative
  adversarial networks.
\newblock In \emph{Proceedings of the AAAI Conference on Artificial
  Intelligence}, volume~34, 4852--4859.

\bibitem[{Liu, Pokharel, and Principe(2008)}]{liu2008kernel}
Liu, W.; Pokharel, P.~P.; and Principe, J.~C. 2008.
\newblock The kernel least-mean-square algorithm.
\newblock \emph{IEEE Transactions on Signal Processing}, 56(2): 543--554.

\bibitem[{Lorenz(1996)}]{lorenz1996predictability}
Lorenz, E.~N. 1996.
\newblock Predictability: A problem partly solved.
\newblock In \emph{Proc. Seminar on predictability}, volume~1.

\bibitem[{Marcinkevi{\v{c}}s and Vogt(2021)}]{marcinkevivcs2021interpretable}
Marcinkevi{\v{c}}s, R.; and Vogt, J.~E. 2021.
\newblock Interpretable models for granger causality using self-explaining
  neural networks.
\newblock \emph{arXiv preprint arXiv:2101.07600}.

\bibitem[{Marinazzo, Pellicoro, and Stramaglia(2008)}]{marinazzo2008kernel}
Marinazzo, D.; Pellicoro, M.; and Stramaglia, S. 2008.
\newblock Kernel method for nonlinear Granger causality.
\newblock \emph{Physical review letters}, 100(14): 144103.

\bibitem[{Massey(1990)}]{massey1990causality}
Massey, J. 1990.
\newblock Causality, feedback and directed information.
\newblock In \emph{Proc. Int. Symp. Inf. Theory Applic.(ISITA-90)}, 303--305.

\bibitem[{Mogren(2016)}]{mogren2016c}
Mogren, O. 2016.
\newblock C-RNN-GAN: Continuous recurrent neural networks with adversarial
  training.
\newblock \emph{Advances in Neural Information Processing Systems (NeurIPS)}.

\bibitem[{Nauta, Bucur, and Seifert(2019)}]{nauta2019causal}
Nauta, M.; Bucur, D.; and Seifert, C. 2019.
\newblock Causal discovery with attention-based convolutional neural networks.
\newblock \emph{Machine Learning and Knowledge Extraction}, 1(1): 19.

\bibitem[{Pamfil et~al.(2020)Pamfil, Sriwattanaworachai, Desai, Pilgerstorfer,
  Georgatzis, Beaumont, and Aragam}]{pamfil2020dynotears}
Pamfil, R.; Sriwattanaworachai, N.; Desai, S.; Pilgerstorfer, P.; Georgatzis,
  K.; Beaumont, P.; and Aragam, B. 2020.
\newblock Dynotears: Structure learning from time-series data.
\newblock In \emph{International Conference on Artificial Intelligence and
  Statistics}, 1595--1605.

\bibitem[{Peters, Janzing, and Sch{\"o}lkopf(2013)}]{peters2013causal}
Peters, J.; Janzing, D.; and Sch{\"o}lkopf, B. 2013.
\newblock Causal inference on time series using restricted structural equation
  models.
\newblock In \emph{Advances in Neural Information Processing Systems},
  volume~26.

\bibitem[{Rangapuram et~al.(2018)Rangapuram, Seeger, Gasthaus, Stella, Wang,
  and Januschowski}]{rangapuram2018deep}
Rangapuram, S.~S.; Seeger, M.~W.; Gasthaus, J.; Stella, L.; Wang, Y.; and
  Januschowski, T. 2018.
\newblock Deep state space models for time series forecasting.
\newblock \emph{Advances in neural information processing systems}, 31.

\bibitem[{Runge et~al.(2019)Runge, Nowack, Kretschmer, Flaxman, and
  Sejdinovic}]{runge2019detecting}
Runge, J.; Nowack, P.; Kretschmer, M.; Flaxman, S.; and Sejdinovic, D. 2019.
\newblock Detecting and quantifying causal associations in large nonlinear time
  series datasets.
\newblock \emph{Science advances}, 5(11): eaau4996.

\bibitem[{{Sanchez Giraldo}, Rao, and Principe(2015)}]{Giraldo2014}
{Sanchez Giraldo}, L.~G.; Rao, M.; and Principe, J.~C. 2015.
\newblock {Measures of entropy from data using infinitely divisible Kernels}.
\newblock \emph{IEEE Trans. Inf. Theory}, 61(1): 535--548.

\bibitem[{Schreiber(2000)}]{schreiber2000measuring}
Schreiber, T. 2000.
\newblock Measuring information transfer.
\newblock \emph{Physical review letters}, 85(2): 461.

\bibitem[{Smith et~al.(2011)Smith, Miller, Salimi-Khorshidi, Webster, Beckmann,
  Nichols, Ramsey, and Woolrich}]{smith2011network}
Smith, S.~M.; Miller, K.~L.; Salimi-Khorshidi, G.; Webster, M.; Beckmann,
  C.~F.; Nichols, T.~E.; Ramsey, J.~D.; and Woolrich, M.~W. 2011.
\newblock Network modelling methods for FMRI.
\newblock \emph{Neuroimage}, 54(2): 875--891.

\bibitem[{Stramaglia, Cortes, and Marinazzo(2014)}]{stramaglia2014synergy}
Stramaglia, S.; Cortes, J.~M.; and Marinazzo, D. 2014.
\newblock Synergy and redundancy in the Granger causal analysis of dynamical
  networks.
\newblock \emph{New Journal of Physics}, 16(10): 105003.

\bibitem[{Takahashi, Chen, and Tanaka-Ishii(2019)}]{takahashi2019modeling}
Takahashi, S.; Chen, Y.; and Tanaka-Ishii, K. 2019.
\newblock Modeling financial time-series with generative adversarial networks.
\newblock \emph{Physica A: Statistical Mechanics and its Applications}, 527:
  121261.

\bibitem[{Tank et~al.(2021)Tank, Covert, Foti, Shojaie, and
  Fox}]{tank2021neural}
Tank, A.; Covert, I.; Foti, N.; Shojaie, A.; and Fox, E.~B. 2021.
\newblock Neural granger causality.
\newblock \emph{IEEE Transactions on Pattern Analysis and Machine
  Intelligence}.

\bibitem[{Van~der Maaten and Hinton(2008)}]{van2008visualizing}
Van~der Maaten, L.; and Hinton, G. 2008.
\newblock Visualizing data using t-SNE.
\newblock \emph{Journal of machine learning research}, 9(11).

\bibitem[{Wang et~al.(2020)Wang, Wang, Li, Lin, Zhao, and Hu}]{wang2020large}
Wang, X.; Wang, R.; Li, F.; Lin, Q.; Zhao, X.; and Hu, Z. 2020.
\newblock Large-scale granger causal brain network based on resting-state fMRI
  data.
\newblock \emph{Neuroscience}, 425: 169--180.

\bibitem[{West and Harrison(2006)}]{west2006bayesian}
West, M.; and Harrison, J. 2006.
\newblock \emph{Bayesian forecasting and dynamic models}.
\newblock Springer Science \& Business Media.

\bibitem[{Wiener(1956)}]{Wiener.N1956}
Wiener, N. 1956.
\newblock {The Theory of Prediction}.
\newblock \emph{Modern Mathematics for Engineers}, 58: 323--329.

\bibitem[{Williams and Zipser(1989)}]{williams1989learning}
Williams, R.~J.; and Zipser, D. 1989.
\newblock A learning algorithm for continually running fully recurrent neural
  networks.
\newblock \emph{Neural computation}, 1(2): 270--280.

\bibitem[{Yoon, Jarrett, and Van~der Schaar(2019)}]{yoon2019time}
Yoon, J.; Jarrett, D.; and Van~der Schaar, M. 2019.
\newblock Time-series generative adversarial networks.
\newblock \emph{Advances in neural information processing systems}, 32.

\bibitem[{Yu et~al.(2020)Yu, Giraldo, Jenssen, and Principe}]{Yu2019}
Yu, S.; Giraldo, L. G.~S.; Jenssen, R.; and Principe, J.~C. 2020.
\newblock {Multivariate extension of matrix-based r{\'{e}}nyi's $\alpha$-order
  entropy functional}.
\newblock \emph{IEEE PAMI}, 42(11): 2960--2966.

\end{thebibliography}
%\pagebreak
% \onecolumn

% This document contains the supplementary material for the \textit{``Causal Recurrent Variational Autoencoder for Medical Time Series Generation"} manuscript. It is organized into the following topics and sections:

\end{document}